%% file: paper.tex
\documentclass{fairmeta}

\usepackage{amsmath}
\usepackage{enumerate} 
\usepackage{algorithm}
\usepackage{algpseudocode}
\usepackage{amsfonts}
\usepackage{amsthm}
\usepackage{newtxtt}
\usepackage{algorithm}
\usepackage{cleveref}
\usepackage{diagbox} 
\usepackage{colortbl}
\usepackage{nicematrix}
\usepackage{subcaption}

\usepackage[frozencache,cachedir=.]{minted}

\usepackage{tcolorbox}
\usepackage[framemethod=tikz]{mdframed}
\usepackage{amssymb}
\usepackage{xspace}
\usepackage{wrapfig}
\usepackage{adjustbox}
\usepackage{tabularx}
\usepackage{booktabs}
\usepackage{mathtools}

\usepackage{tikz}
\usetikzlibrary{positioning, calc}
\usetikzlibrary{decorations.pathmorphing}

\usepackage{silence}
\makeatletter
\patchcmd{\wrong@fontshape}{\@gobbletwo}{}{}{}
\makeatother
\input{math_commands}

\newcommand{\highlight}[1]{{\color{metablue} \textbf{#1}}}
\newmdenv[backgroundcolor=metabg, roundcorner=5pt, skipabove=7pt, linewidth=0pt, innertopmargin=4pt]{myframe}

\renewcommand{\eqref}[1]{\labelcref{#1}}

\numberwithin{equation}{section} 
\tcbuselibrary{minted}

\setminted[python]{
  linenos,
  breaklines,
  fontsize=\footnotesize,
  xleftmargin=2em
}

\definecolor{light}{RGB}{125, 125, 125}

\crefname{tcb@cnt@pbox}{code}{code}
\Crefname{tcb@cnt@pbox}{Code}{Code}
\crefname{assumption}{assumption}{assumption}
\Crefname{assumption}{Assumption}{Assumptions}

\newtcolorbox[auto counter]{pbox}[2][]{
  colback=white,
  title=Code~\thetcbcounter: #2,
  #1,fonttitle=\sffamily,
  fontupper=\sffamily,
  arc=10pt,
  colframe=metabg,
  coltitle=metafg,
  colbacktitle=metabg,
  toptitle=0.25cm,
  bottomtitle=0.125cm
}

\title{Flow Matching Guide and Code}

\author[1]{Yaron Lipman}
\author[1]{Marton Havasi}
\author[2]{Peter Holderrieth}
\author[3]{Neta Shaul}
\author[1]{Matt Le}
\author[1]{Brian Karrer} 
\author[1]{\qquad \qquad \qquad Ricky T. Q. Chen}
\author[1]{David Lopez-Paz}
\author[3]{Heli Ben-Hamu}
\author[1]{Itai Gat}

\affiliation[1]{FAIR at Meta}
\affiliation[2]{MIT CSAIL}
\affiliation[3]{Weizmann Institute of Science}

\abstract{%
Flow Matching (FM) is a recent framework for generative modeling that has achieved state-of-the-art performance across various domains, including image, video, audio, speech, and biological structures.
This guide offers a comprehensive and self-contained review of FM, covering its mathematical foundations, design choices, and extensions.
By also providing a PyTorch package featuring relevant examples (\eg, image and text generation), this work aims to serve as a  resource for both novice and experienced researchers interested in understanding, applying and further developing FM.
}

\date{\today}
\metadata[Code]{\fmlibrary{} at  \url{https://github.com/facebookresearch/flow_matching} }

\begin{document}

\maketitle

\tableofcontents
\clearpage
\newpage

\section{Introduction}\label{section:intro}

Flow matching (FM) \citep{lipman2022flow,albergo2022building,liu2022flow} is a simple framework for generative modeling framework that has pushed the state-of-the-art in various fields and large-scale applications including generation of images \citep{esser2024scaling}, videos \citep{polyak2024moviegencastmedia}, speech \citep{le2024voicebox}, audio \citep{vyas2023audiobox}, proteins \citep{huguet2024sequence}, and robotics \citep{black2024robotics}.
This manuscript and its accompanying codebase have two primary objectives.
First, to serve as a comprehensive and self-contained reference to Flow Matching, detailing its design choices and numerous extensions developed by the research community.
Second, to enable newcomers to quickly adopt and build upon Flow Matching for their own applications.

\begin{figure}
\centering
\begin{subfigure}[b]{0.24\textwidth}
\centering
\includegraphics[width=\textwidth]{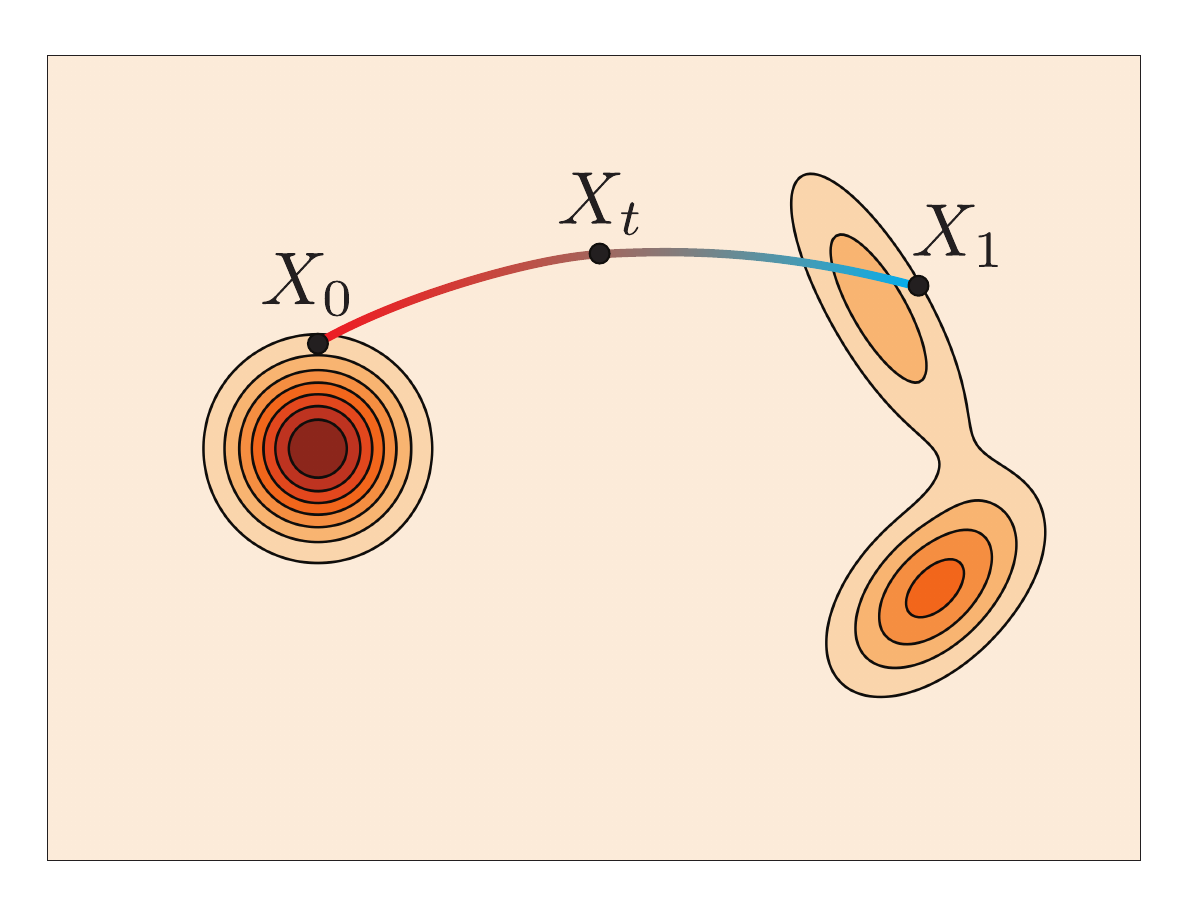}
\caption{Flow}
\label{fig:types:flow}
\end{subfigure}
\hfill
\begin{subfigure}[b]{0.24\textwidth}
\centering
\includegraphics[width=\textwidth]{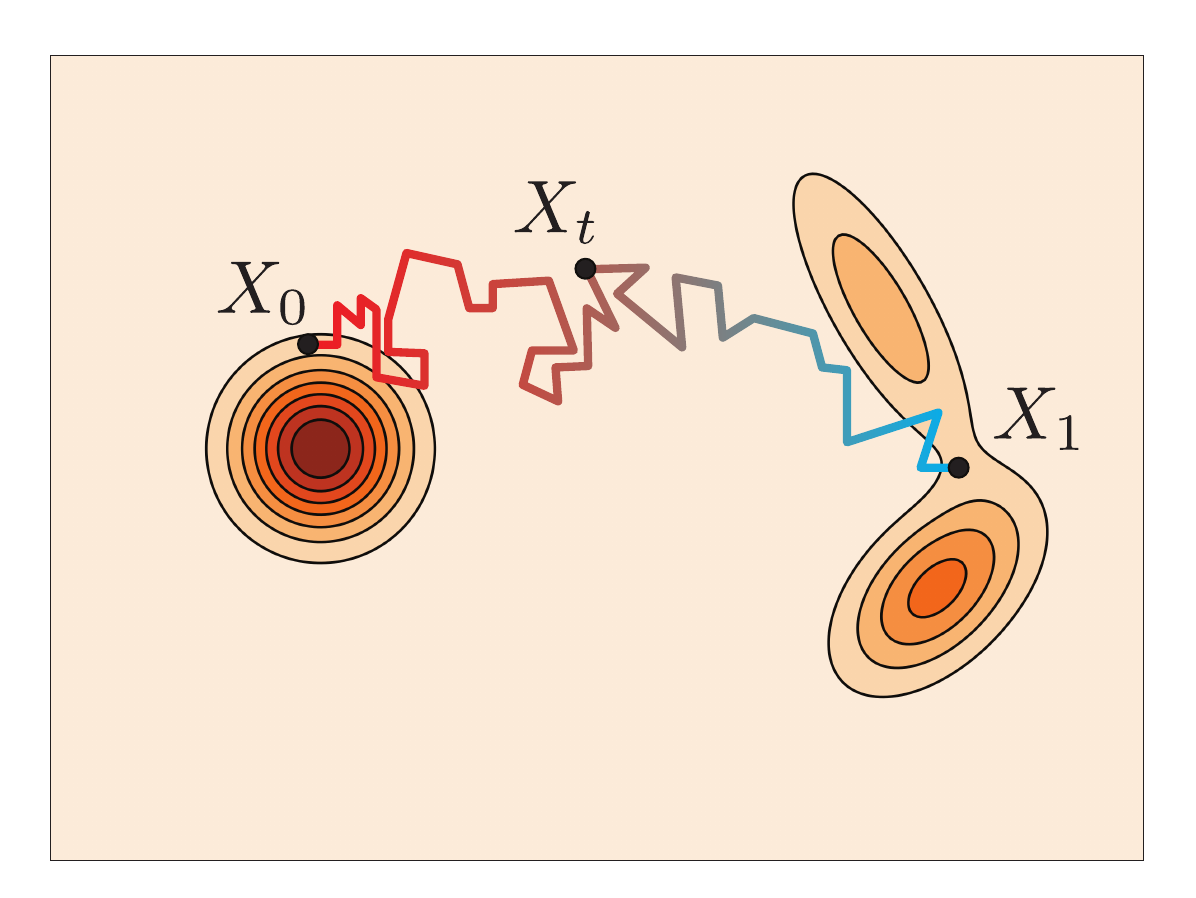}
\caption{Diffusion}
\label{fig:types:diffusion}
\end{subfigure}
\hfill
\begin{subfigure}[b]{0.24\textwidth}
\centering
\includegraphics[width=\textwidth]{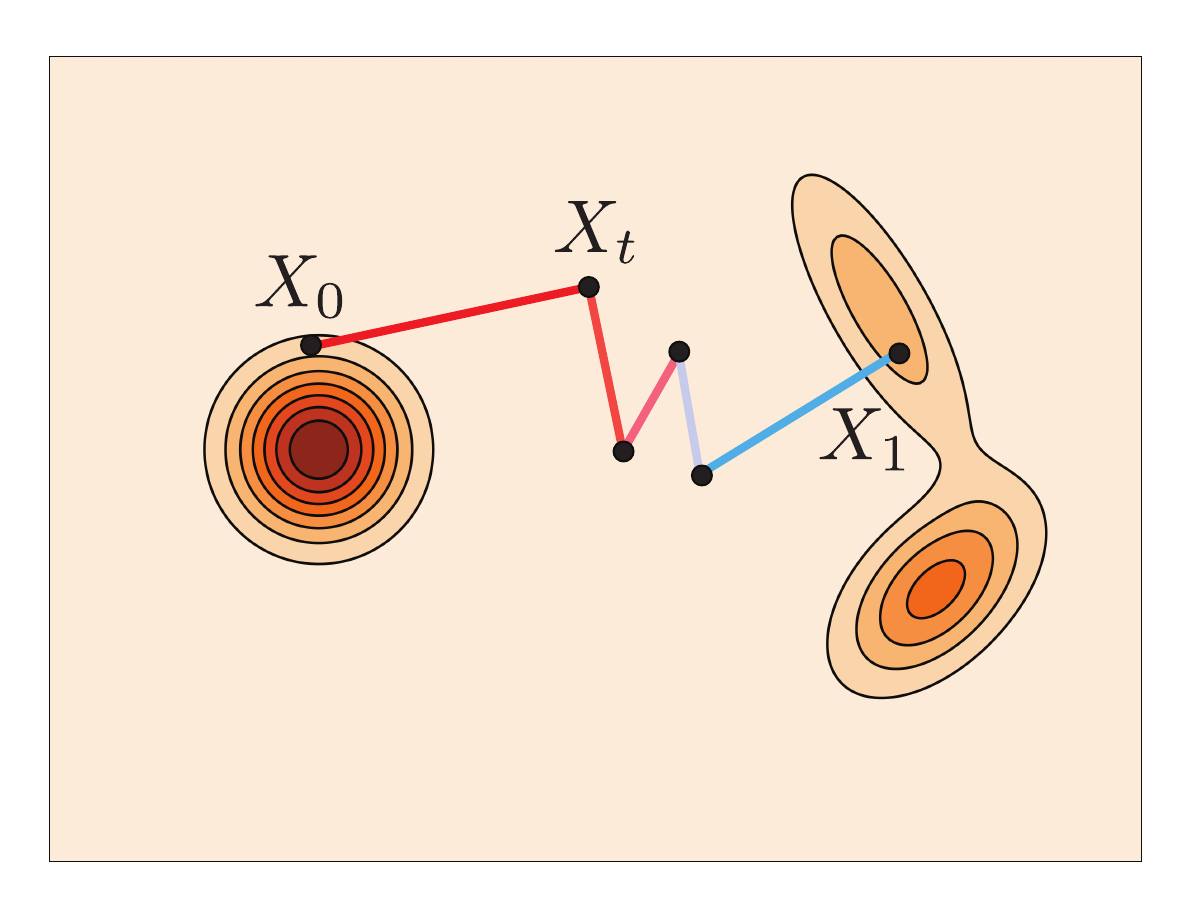}
\caption{Jump}
\label{fig:types:jump}
\end{subfigure}
\hfill
\begin{subfigure}[b]{0.24\textwidth}
\centering
\includegraphics[width=\textwidth]{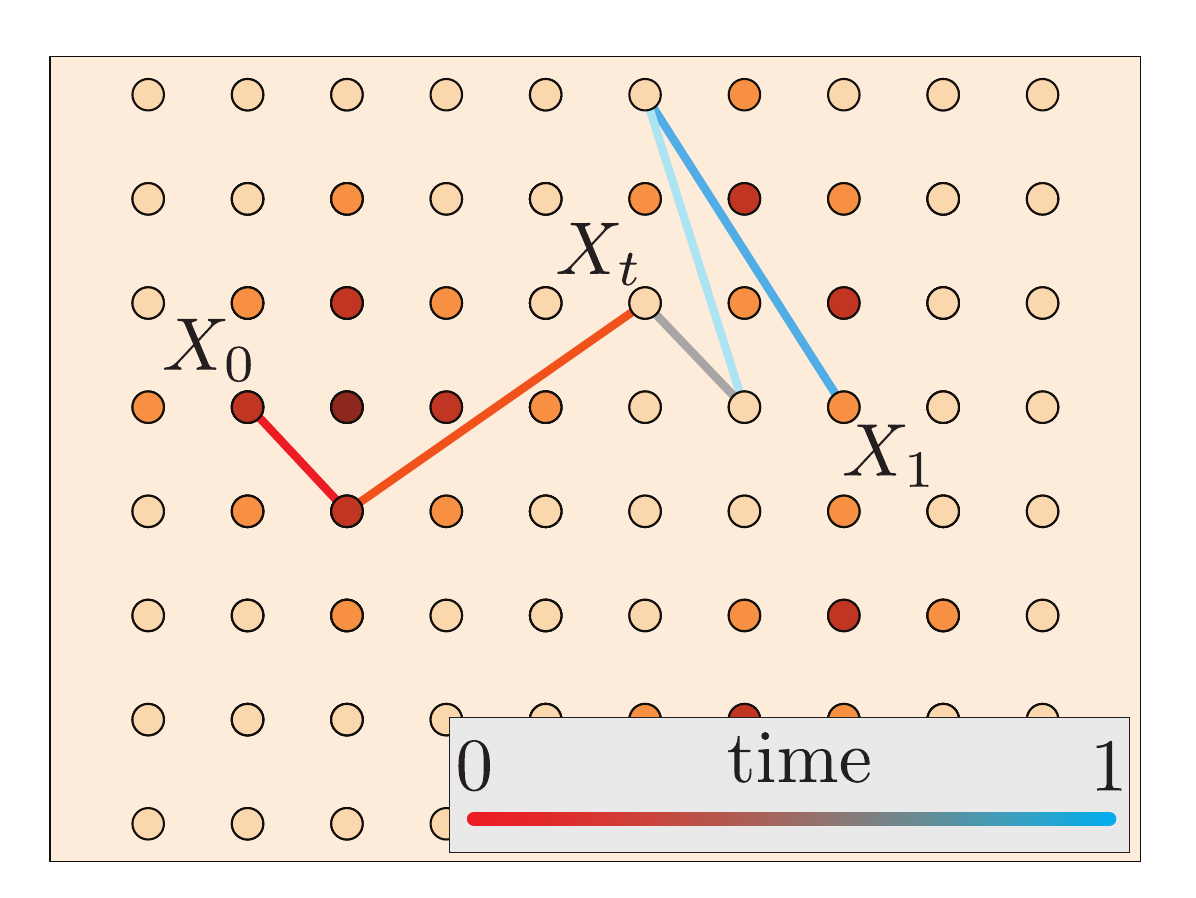}
\caption{CTMC}
\label{fig:types:ctmc}
\end{subfigure}
\caption{Four time-continuous processes $(X_t)_{0\leq t \leq 1}$ taking source sample $X_0$ to a target sample $X_1$.
These are a flow in a continuous state space, a diffusion in continuous state space, a jump process in continuous state space (densities visualized with contours), and a jump process in discrete state space (states as disks, probabilities visualized with colors). }
\label{fig:types}
\end{figure}

The framework of Flow Matching is based on learning a velocity field (also called vector field). Each velocity field defines a \highlight{flow} $\psi_t$ by solving an ordinary differential equation (ODE) in a process called simulation. A flow is a determinstic, time-continuous bijective transformation of the $d$-dimensional Euclidean space, $\Real^d$. The goal of Flow Matching is to build a flow that transforms a sample $X_0 \sim p$ drawn from a source distribution $p$ into a target sample $X_1\defe \psi_1(X_0)$ such that $X_1\sim q$ has a desired distribution $q$, see \cref{fig:types:flow}. Flow models were introduced to the machine learning community by \citep{chen2018neural,grathwohl2018ffjord} as Continuous Normalizing Flows (CNFs). Originally, flows were trained by maximizing the likelihood $p(X_1)$ of training examples $X_1$, resulting in the need of simulation and its differentiation during training.
Due to the resulting computational burdens, later works attempted to learn CNFs without simulation \citep{rozen2021moser,benhamu2022cnfm}, evolving into modern-day Flow Matching algorithms \citep{lipman2022flow,liu2022flow,albergo2022building,neklyudov2023action,heitz2023iterative,tong2023improving}.
The resulting framework is a recipe comprising two steps \citep{lipman2022flow}, see \cref{fig:blueprint}: 
First, choose a probability path $p_t$ interpolating between the source $p$ and target $q$ distributions.
Second, train a velocity field (neural network) that defines the flow transformation $\psi_t$ implementing $p_t$.

The principles of FM can be extended to state spaces $\gS$ other than $\mathbb{R}^d$ and even evolution processes that are not flows. Recently, \highlight{Discrete Flow Matching} \citep{campbell2024generative,gat2024discrete} develops a Flow Matching algorithm for time-continuous Markov processes on discrete state spaces, also known as Continuous Time Markov Chains (CTMC), see \cref{fig:types:jump}. This advancement opens up the exciting possibility of using Flow Matching in discrete generative tasks such as language modeling. \highlight{Riemannian Flow Matching} \citep{chen2024flow} extends Flow Matching to flows on Riemannian manifolds $\gS=\gM$ that now became the state-of-the-art models for a wide variety of applications of machine learning in chemistry such as protein folding \citep{yim2023fast,bose2023se}.
Even more generally, \highlight{Generator Matching} \citep{holderrieth2024gm} shows that the Flow Matching framework works for any modality and for general Continuous Time Markov Processes (\highlight{CTMPs}) including, as illustrated in~\cref{fig:types}, \highlight{flows}, \highlight{diffusions}, and \highlight{jump processes} in continuous spaces, in addition to CTMC in discrete spaces.
Remarkably, for any such CTMP, the Flow Matching recipe remains the same, namely: First, choose a path $p_t$ interpolating source $p$ and target $q$ on the relevant state space $\gS$. Second, train a \emph{generator}, which plays a similar role to velocities for flows, and defines a CTMP process implementing $p_t$. This generalization of Flow Matching allows us to see many existing generative models in a unified light and \hbox{develop new generative models for any modality with a generative Markov process of choice.}

Chronologically, \highlight{Diffusion Models} were the first to develop simulation-free training of a CTMP process, namely a diffusion process, \cref{fig:types:diffusion}. Diffusion Models were originally introduced as discrete time Gaussian processes \citep{sohl2015deep,ho2020denoising} and later formulated in terms of continuous time Stochastic Differential Equations (SDEs) \citep{song2021sde}. In the lens of Flow Matching, Diffusion Models build the probability path $p_t$ interpolating source and target distributions in a particular way via \emph{forward noising processes} modeled by particular SDEs. These SDEs are chosen to have closed form marginal probabilities that are in turn used to parametrize the generator of the diffusion process (\ie, drift and diffusion coefficient) via the \emph{score function} \citep{song2019generative}. This parameterization is based on a reversal process to the forward noising process \citep{anderson1982reverse}. Consequently, Diffusion Models learn the score function of the marginal probabilities. Diffusion Models' literature suggested also other parametrizations of the generator besides the score, including \emph{noise prediction}, \emph{denoisers} \citep{kingma2021variational}, or \emph{$v$-prediction} \citep{salimans2022progressive}---where the latter coincides with velocity prediction for a particular choice of probability path $p_t$. \highlight{Diffusion bridges} \citep{peluchetti2023non} offers another approach to design $p_t$ and generators for diffusion process that extends diffusion models to arbitrary source-target couplings. In particular these constructions are build again on SDEs with marginals known in closed form, and again use the score to formulate the generator (using Doob's $h$-transform). \citet{shi2023diffusion,liu2023i2sb} show that the linear version of Flow Matching can be seen as a certain limiting case of bridge matching.

The rest of this manuscript is organized as follows.
\Cref{s:quick_tour} offers a self-contained ``cheat-sheet'' to understand and implement vanilla Flow Matching in PyTorch.
\Cref{s:flow_models} offers a rigorous treatment of flow models, arguably the simplest of all CTMPs, for continuous state spaces. 
In \cref{s:fm_continuous} we introduce the Flow Matching framework in $\Real^d$ and its various design choices and extensions. We show that flows can be constructed by considering the significantly simpler conditional setting, offering great deal of flexibility in their design, \eg, by readily extending to Riemannian geometries, described in \cref{s:fm_non_euclidean}.
\Cref{sec:ctmcm} provides an introduction to Continuous Time Markov Chains (CTMCs) and the usage as generative models on discrete state spaces. 
Then, \cref{sec:discreteflow} discusses the extension of Flow Matching to CTMC processes. 
In \cref{sec:fellerflow}, we provide an introduction to using Continuous Time Markov Process (CTMPs) as generative models for arbitrary state spaces.
In \cref{sec:generatormatching}, we describe Generator Matching (GM) - a generative modeling framework for arbitrary modalities that describes a scalable way of training CTMPs. GM also unifies all models in previous sections into a common framework.
Finally, due to their wide-spread use, we discuss in \cref{s:relation_to_diffusion_models} denoising diffusion models as a specific instance of the FM family of models.

\section{Quick tour and key concepts}\label{s:quick_tour}

Given access to a training dataset of samples from some target distribution $q$ over $\R^d$, our goal is to build a model capable of generating new samples from $q$.
To address this task, \highlight{Flow Matching (FM)} builds a \highlight{probability path} $(p_t)_{0\leq t\leq 1}$, from a known source distribution $p_0=p$ to the data target distribution $p_1=q$, where each $p_t$ is a distribution over $\R^d$.
Specifically, FM is a simple regression objective to train the \highlight{velocity field} neural network describing the instantaneous velocities of samples---later used to convert the source distribution $p_0$ into the target distribution $p_1$, along the probability path $p_t$. 
After training, we generate a novel sample from the target distribution $X_1 \sim q$ by (i) drawing a novel sample from the source distribution $X_0 \sim p$, and (ii) solving the Ordinary Differential Equation (ODE) determined by the velocity field.

\begin{figure}
\centering
\begin{subfigure}[b]{0.24\textwidth}
\includegraphics[width=\textwidth]{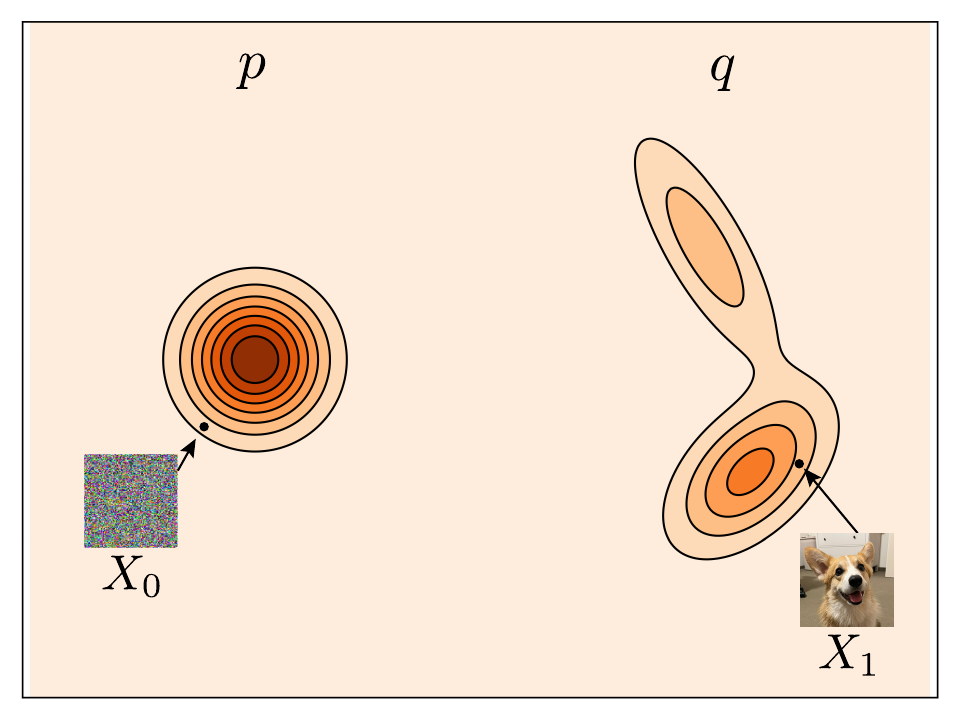}
\caption{Data.}
\label{fig:blueprint:data}
\end{subfigure}
\hfill
\begin{subfigure}[b]{0.24\textwidth}
\includegraphics[width=\textwidth]{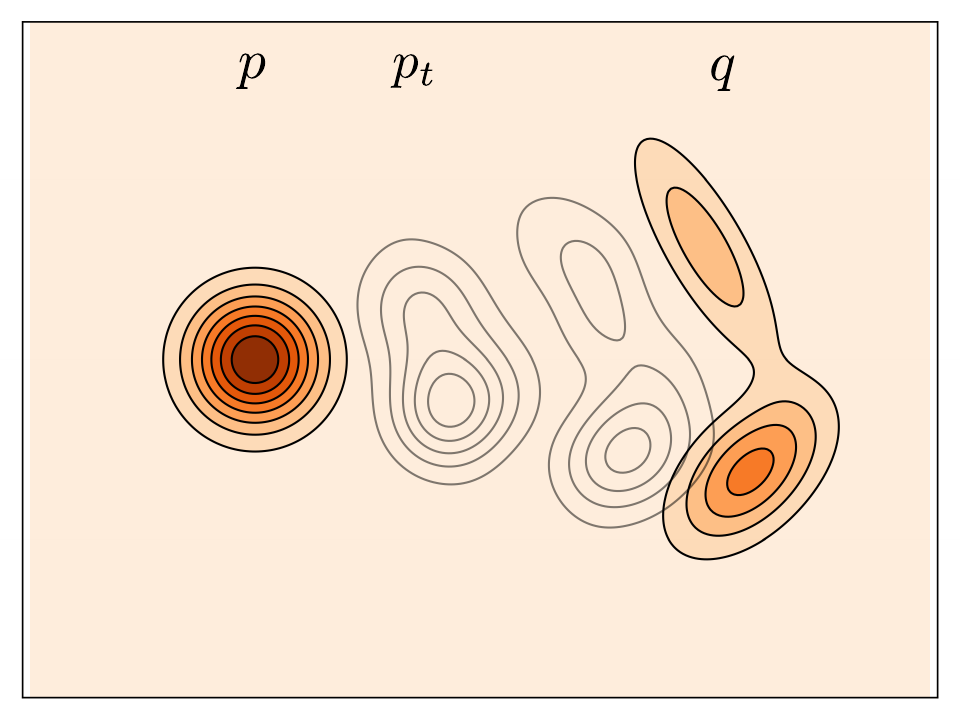}
\caption{Path design.}
\label{fig:blueprint:path}
\end{subfigure}
\hfill
\begin{subfigure}[b]{0.24\textwidth}
\includegraphics[width=\textwidth]{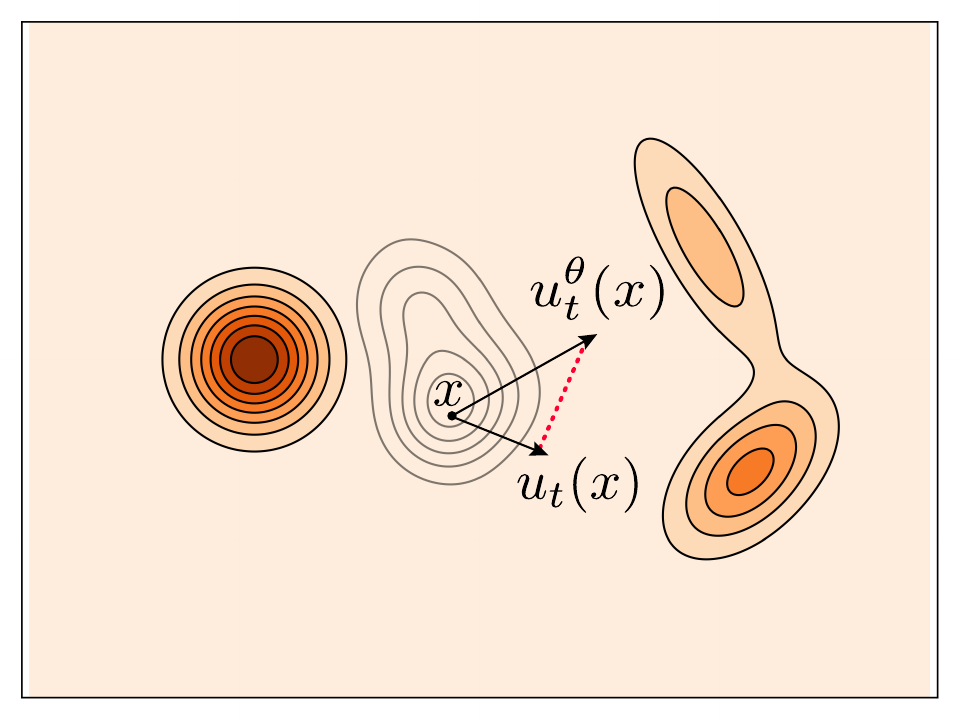}
\caption{Training.}
\label{fig:blueprint:training}
\end{subfigure}
\hfill
\begin{subfigure}[b]{0.24\textwidth}
\includegraphics[width=\textwidth]{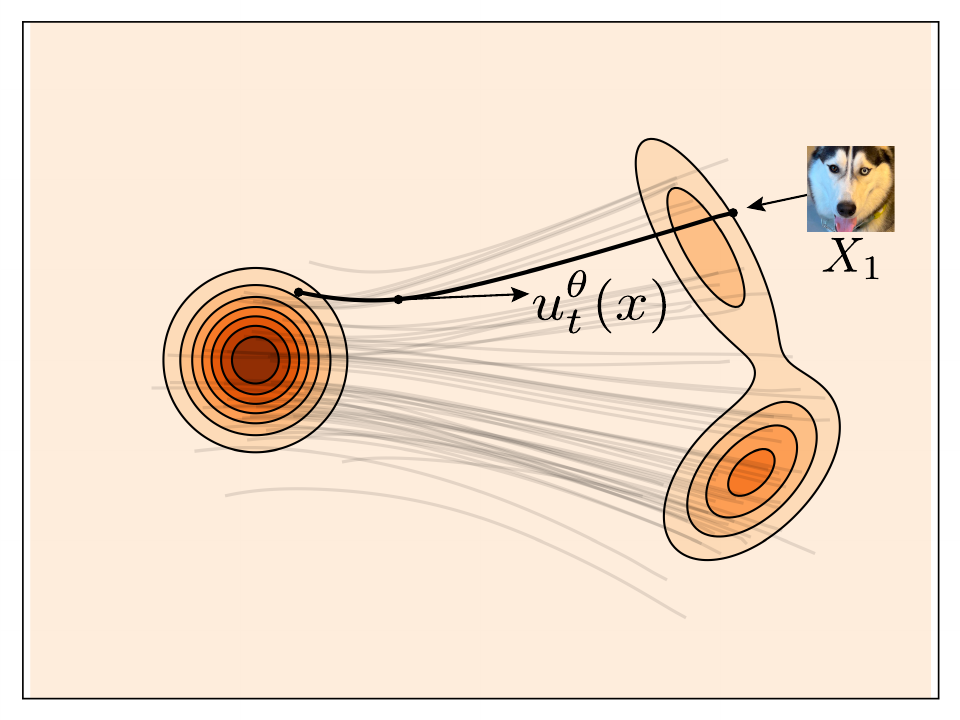}
\caption{Sampling.}
\label{fig:blueprint:sampling}
\end{subfigure}
\caption{\emph{The Flow Matching blueprint}.
(a) The goal is to find a flow mapping samples $X_0$ from a known source or noise distribution $q$ into samples $X_1$ from an unknown target or data distribution $q$.
(b) To do so, design a time-continuous probability path $(p_t)_{0\leq t \leq 1}$ interpolating between $p := p_0$ and $q := p_1$.
(c) During training, use regression to estimate the velocity field $u_t$ known to generate $p_t$.
(d) To draw a novel target sample $X_1 \sim q$, integrate the estimated velocity field $u^\theta_t(X_t)$ from $t=0$ to $t=1$, where $X_0 \sim p$ is a novel source sample.
  }
\label{fig:blueprint}
\end{figure}

More formally, an ODE is defined via a time-dependent vector field $u: [0,1] \times \mathbb{R}^d \to \mathbb{R}^d$ which, in our case, is the velocity field modeled in terms of a neural network. 
This velocity field determines a time-dependent \highlight{flow} $\psi : [0,1] \times \mathbb{R}^d \to \mathbb{R}^d$, defined as 
\begin{equation*}
  \frac{\dd}{\dd t} \psi_t(x) = u_t(\psi_t(x)), 
\end{equation*}
where $\psi_t := \psi(t, x)$ and $\psi_0(x) = x$.
The velocity field $u_t$ \emph{generates} the probability path $p_t$ if its flow $\psi_t$ satisfies 
\begin{equation}
 X_t \defe \psi_t(X_0) \sim p_t \text{ for } X_0 \sim p_0.
\end{equation}
According to the equation above, the velocity field $u_t$ is the only tool necessary to sample from $p_t$ by solving the ODE above. 
As illustrated in~\cref{fig:blueprint:sampling}, solving the ODE until $t=1$ provides us with samples $X_1=\psi_1(X_0)$, resembling the target distribution $q$.
Therefore, and in sum, the goal of Flow Matching is to learn a vector field $u^\theta_t$ such that its flow $\psi_t$ generates a probability path $p_t$ with $p_0=p$ and $p_1=q$.

Using the notations above, the goal of Flow Matching is to learn the parameters $\theta$ of a velocity field $u^\theta_t$ implemented in terms of a neural network.
As anticipated in the introduction, we do this in two steps: design a probability path $p_t$ interpolating between $p$ and $q$ (see~\cref{fig:blueprint:path}), and train a velocity field $u^\theta_t$ generating $p_t$ by means of regression (see~\cref{fig:blueprint:training}).

\begin{figure}
\centering
\begin{subfigure}[b]{0.24\textwidth}
\centering
\includegraphics[width=\textwidth]{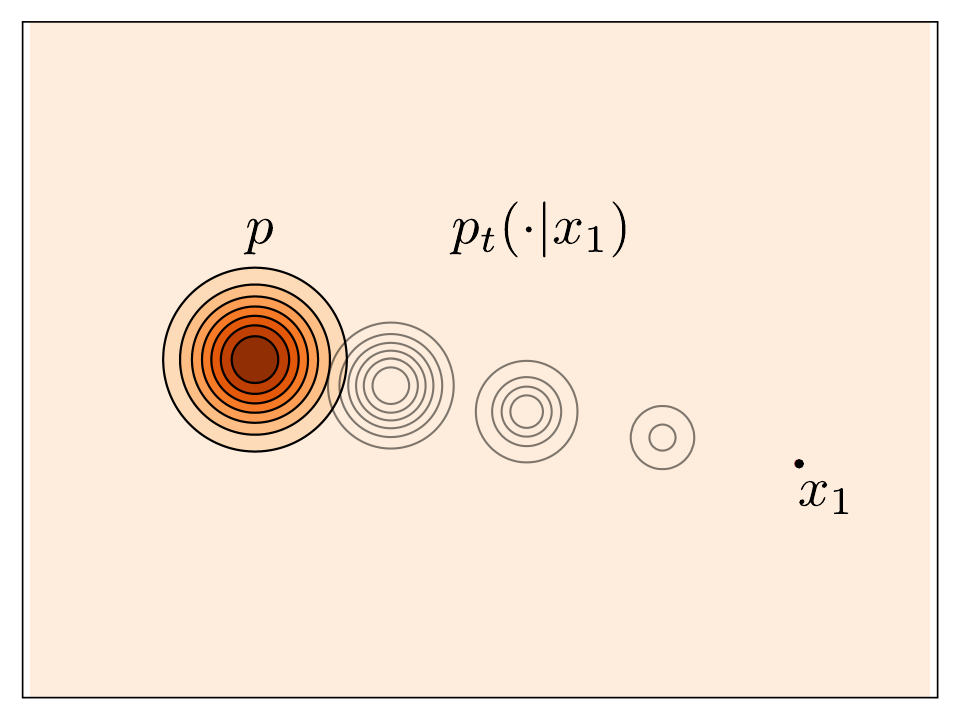}
\caption{Conditional probability path $p_t(x | x_1)$.}
\label{fig:p_t}
\end{subfigure}
\hfill
\begin{subfigure}[b]{0.24\textwidth}
\includegraphics[width=\textwidth]{assets/recepie/p_t.pdf}
\caption{(Marginal) Probability path $p_t(x)$.}
\label{fig:marginal_p_t}
\end{subfigure}
\hfill
\begin{subfigure}[b]{0.24\textwidth}
\centering
\includegraphics[width=\textwidth]{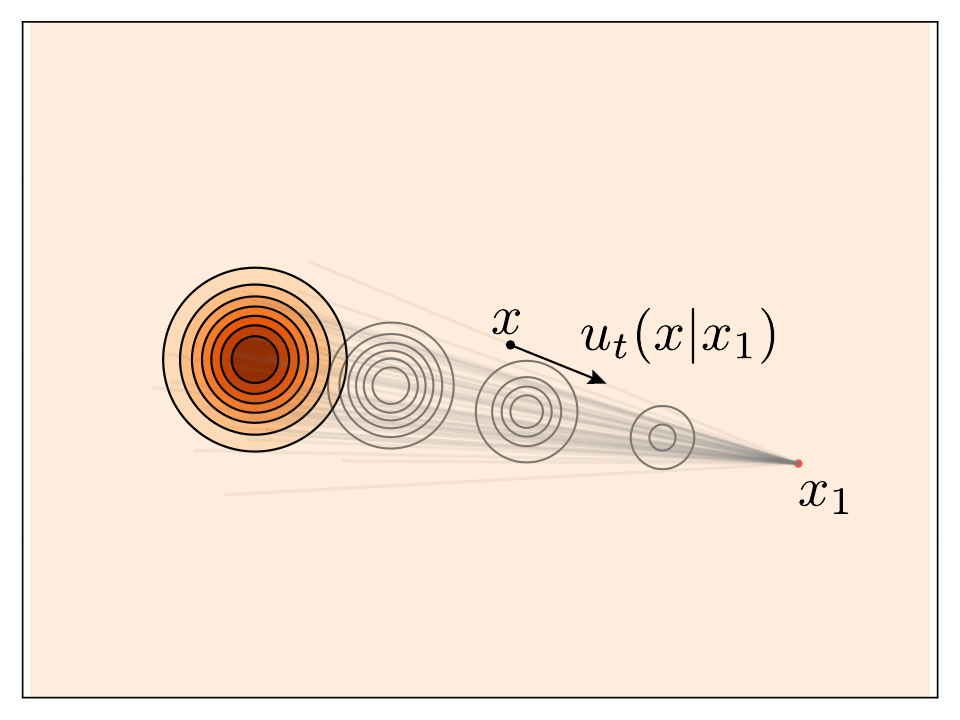}
\caption{Conditional velocity field $u_t(x | x_1)$.}
\label{fig:u_t}
\end{subfigure}
\hfill
\begin{subfigure}[b]{0.24\textwidth}
\centering
\includegraphics[width=\textwidth]{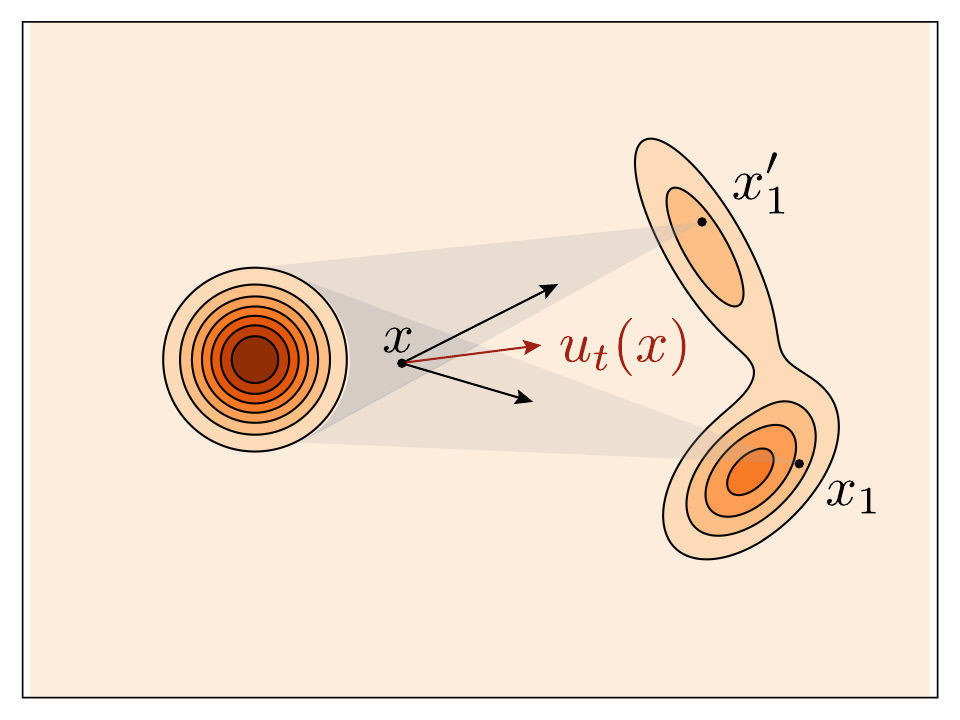}
\caption{(Marginal) Velocity field $u_t(x)$.}
\label{fig:marginal_u_t}
\end{subfigure}
\caption{\emph{Path design in Flow Matching}. Given a fixed target sample $X=x_1$, its conditional velocity field $u_t(x|x_1)$ generates the conditional probability path $p_t(x|x_1)$.
  The (marginal) velocity field $u_t(x)$ results from the aggregation of all conditional velocity fields---and similarly for the probability path $p_t(x)$. }
\end{figure}

Therefore, let us proceed with the first step of the recipe: designing the probability path $p_t$.
In this example, let the source distribution $p := p_0 = \gN(x | 0, I)$, and construct the probability path $p_t$ as the aggregation of the conditional probability paths $p_{t|1}(x | x_1)$, each conditioned on one of the data examples $X_1=x_1$ comprising the training dataset.
(One of such conditional paths is illustrated in~\cref{fig:p_t}.)
The probability path $p_t$ therefore follows the expression:
\begin{equation}\label{e:condot_path}
    p_t(x) = \int p_{t|1}(x|x_1)q(x_1)\dd x_1, \text{ where } p_{t|1}(x|x_1)=\gN(x | t x_1, (1-t)^2I).
\end{equation}
This path, also known as the \emph{conditional optimal-transport} or \emph{linear} path, enjoys some desirable properties that we will study later in this manuscript.
Using this probability path, we may define the random variable $X_t \sim p_t$ by drawing $X_0 \sim p$, drawing $X_1 \sim q$, and taking their linear combination:
\begin{equation}\label{e:quick_tour_X_t}
    X_t = t X_1 + (1-t) X_0 \sim p_t.
\end{equation}

We now continue with the second step in the Flow Matching recipe: regressing our velocity field $u^\theta_t$ (usually implemented in terms of a neural network) to a target velocity field $u_t$ known to generate the desired probability path $p_t$.
To this end, the \highlight{Flow Matching loss} reads:
\begin{equation}\label{eq:fm}
  \gL_{\FM}(\theta) = \mathbb{E}_{t, X_t} \norm{u^\theta_t(X_t) - u_t(X_t) }^2,
  \text{ where } t \sim \gU[0, 1] \text{ and } X_t \sim p_t.
\end{equation}
In practice, one can rarely implement the objective above, because $u_t$ is a complicated object governing the \emph{joint} transformation between two high-dimensional distributions.
Fortunately, the objective simplifies drastically by conditioning the loss on a single target example $X_1=x_1$ picked at random from the training set. 
To see how, borrowing \cref{e:quick_tour_X_t} to realize the conditional random variables
\begin{equation}
    X_{t|1} = tx_1 + (1-t)X_0 \quad \sim \quad  p_{t|1}(\cdot|x_1)=\gN(\cdot\ |\ tx_1,(1-t)^2I).
\end{equation}
Using these variables, solving $\frac{\dd}{\dd t}X_{t|1} = u_t(X_{t|1}|x_1)$ leads to the \highlight{conditional velocity field}
\begin{equation}\label{e:quick_u_t_cond}
   u_t(x|x_1) = \frac{x_1-x}{1-t},
\end{equation}
which generates the conditional probability path $p_{t|1}(\cdot|x_1)$.
(For an illustration on these two conditional objects, see~\cref{fig:u_t}.)
Equipped with the simple \cref{e:quick_u_t_cond} for the conditional velocity fields generating the designed conditional probability paths, we can formulate a tractable version of the Flow Matching loss in~\eqref{eq:fm}.
This is the \highlight{conditional Flow Matching loss}:
\begin{equation}\label{eq:cfm}
    \gL_{\CFM}(\theta) = \mathbb{E}_{t, X_t, X_1} \| u^\theta_t(X_t) - u_t(X_t|X_1) \|^2,
    \text{ where } t \sim U[0,1],\, X_0 \sim p ,\,  X_1 \sim q,    
\end{equation}
and $X_t=(1-t)X_0 + tX_1$.
Remarkably, the objectives in \eqref{eq:fm} and \eqref{eq:cfm} provide the same gradients to learn $u^\theta_t$, \ie, 
\begin{equation}
    \nabla_\theta \gL_{\FM}(\theta) = \nabla_\theta \gL_{\CFM}(\theta).
\end{equation}
Finally, by plugging $u_t(x|x_1)$ from \eqref{e:quick_u_t_cond} into \cref{eq:cfm}, we get the simplest implementation of Flow Matching:
\begin{equation}\label{eq:gauss}
    \gL^{\text{{OT,Gauss}}}_{\CFM}(\theta) = \mathbb{E}_{t, X_0, X_1} \| u^\theta_t(X_t) -\parr{X_1-X_0} \|^2,
    \text{ where } t \sim U[0,1],\, X_0 \sim \gN(0,I),\, X_1 \sim q.     
\end{equation}
A standalone implementation of this quick tour in pure PyTorch is provided in~\cref{ex:fm_standalone}.
Later in the manuscript, we will cover more sophisticated variants and design choices, all of them implemented in the accompanying \fmlibrary.

\begin{pbox}[label={ex:fm_standalone}]{Standalone Flow Matching code \\ \url{flow_matching/examples/standalone_flow_matching.ipynb}}
\begin{minted}{python}
import torch 
from torch import nn, Tensor
import matplotlib.pyplot as plt
from sklearn.datasets import make_moons

class Flow(nn.Module):
  def __init__(self, dim: int = 2, h: int = 64):
    super().__init__()
    self.net = nn.Sequential(
      nn.Linear(dim + 1, h), nn.ELU(),
      nn.Linear(h, h), nn.ELU(),
      nn.Linear(h, h), nn.ELU(),
      nn.Linear(h, dim))
  
  def forward(self, x_t: Tensor, t: Tensor) -> Tensor:
    return self.net(torch.cat((t, x_t), -1))
  
  def step(self, x_t: Tensor, t_start: Tensor, t_end: Tensor) -> Tensor:
    t_start = t_start.view(1, 1).expand(x_t.shape[0], 1)
    # For simplicity, using midpoint ODE solver in this example 
    return x_t + (t_end - t_start) * self(x_t + self(x_t, t_start) * (t_end - t_start) / 2,
                                         t_start + (t_end - t_start) / 2)

# training
flow = Flow()
optimizer = torch.optim.Adam(flow.parameters(), 1e-2)
loss_fn = nn.MSELoss()

for _ in range(10000):
  x_1  = Tensor(make_moons(256, noise=0.05)[0])
  x_0  = torch.randn_like(x_1)
  t    = torch.rand(len(x_1), 1)
  x_t  = (1 - t) * x_0 + t * x_1
  dx_t = x_1 - x_0
  optimizer.zero_grad()
  loss_fn(flow(x_t, t), dx_t).backward()
  optimizer.step()

# sampling
x = torch.randn(300, 2)
n_steps = 8
fig, axes = plt.subplots(1, n_steps + 1, figsize=(30, 4), sharex=True, sharey=True)
time_steps = torch.linspace(0, 1.0, n_steps + 1)

axes[0].scatter(x.detach()[:, 0], x.detach()[:, 1], s=10)
axes[0].set_title(f't = {time_steps[0]:.2f}')
axes[0].set_xlim(-3.0, 3.0)
axes[0].set_ylim(-3.0, 3.0)

for i in range(n_steps):
  x = flow.step(x, time_steps[i], time_steps[i + 1])
  axes[i + 1].scatter(x.detach()[:, 0], x.detach()[:, 1], s=10)
  axes[i + 1].set_title(f't = {time_steps[i + 1]:.2f}')

plt.tight_layout()
plt.show()
\end{minted}
\vspace{0.5cm}
\includegraphics[width=\textwidth]{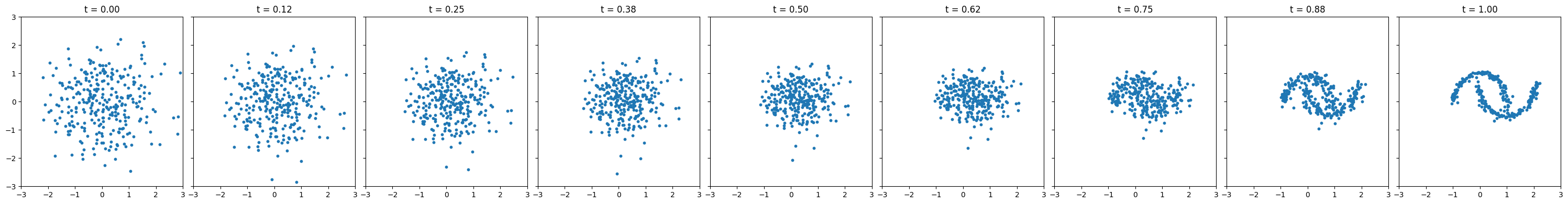}
\end{pbox}

\section{Flow models}\label{s:flow_models}

This section introduces \highlight{flows}, the mathematical object powering the simplest forms of Flow Matching.
Later parts in the manuscript will discuss Markov processes more general than flows, leading to more sophisticated generative learning paradigms introducing many more design choices to the Flow Matching framework. 
The reason we start with flows is three-fold: First, flows are arguably the simplest of all CTMPs---being deterministic and having a compact parametrization via velocities---these models can transform any source distribution $p$ into any target distribution $q$, as long as these two have densities. Second, flows can be sampled rather efficiently by approximating the solution of ODEs, compared, \eg, to the harder-to-simulate SDEs for diffusion processes. Third, the deterministic nature of flows allows an  unbiased model likelihood estimation, while more general stochastic processes require working with lower bounds.
To understand flows, we must first review some background notions in probability and differential equations theory, which we do next. 

\subsection{Random vectors}

Consider data in the $d$-dimensional Euclidean space $x=(x^1,\ldots,x^d)\in \Real^d$ with the standard Euclidean inner product $\ip{x,y}=\sum_{i=1}^d x^i y^i$ and norm $\norm{x}=\sqrt{\ip{x,x}}$.
We will consider random variables (RVs) $X\in\R^d$ with continuous probability density function (PDF), defined as a \emph{continuous} function $p_X:\Real^d\too \Real_{\geq 0}$ providing event $A$ with probability
\begin{equation}
    \sP(X\in A) = \int_A p_X(x) \dd x,
\end{equation}
where $\int p_X(x)\dd x = 1$.
By convention, we omit the integration interval when integrating over the whole space ($\int \equiv \int_{\Real^d}$).
To keep notation concise, we will refer to the PDF $p_{X_t}$ of RV $X_t$ as simply $p_t$.
We will use the notation $X \sim p$ or $X \sim p(X)$ to indicate that $X$ is distributed according to $p$.
One common PDF in generative modeling is the $d$-dimensional isotropic Gaussian:
\begin{equation}\label{e:gaussian}
  \gN(x|\mu,\sigma^2 I) = (2\pi\sigma^2)^{-\frac{d}{2}}\exp\left(-\frac{\norm{x-\mu}_2^2}{2\sigma^2}\right),
\end{equation}
where $\mu\in \Real^d$ and $\sigma \in \Real_{>0}$ stand for the mean and the standard deviation of the distribution, respectively.

The expectation of a RV is the constant vector closest to $X$ in the least-squares sense:
\begin{equation}\label{e:E}
    \E\brac{X}=\argmin_{z\in\Real^d} \int \norm{x-z}^2 p_X(x)\dd x = \int x p_X(x)\dd x. %
\end{equation}
One useful tool to compute the expectation of \emph{functions of RVs} is the \emph{Law of the Unconscious Statistician}:
\begin{equation}\label{e:law_of_uncon_stat}
    \E \brac{f(X)} = \int f(x) p_X(x) \dd x. %
\end{equation}
When necessary, we will indicate the random variables under expectation as $\E_{X} f(X)$.

\subsection{Conditional densities and expectations} \label{sec:conditional_densities_and_expectations}
\begin{wrapfigure}{r}{0.25\textwidth}
  \vspace{-40pt}
  \begin{center}
    \includegraphics[width=0.25\textwidth]{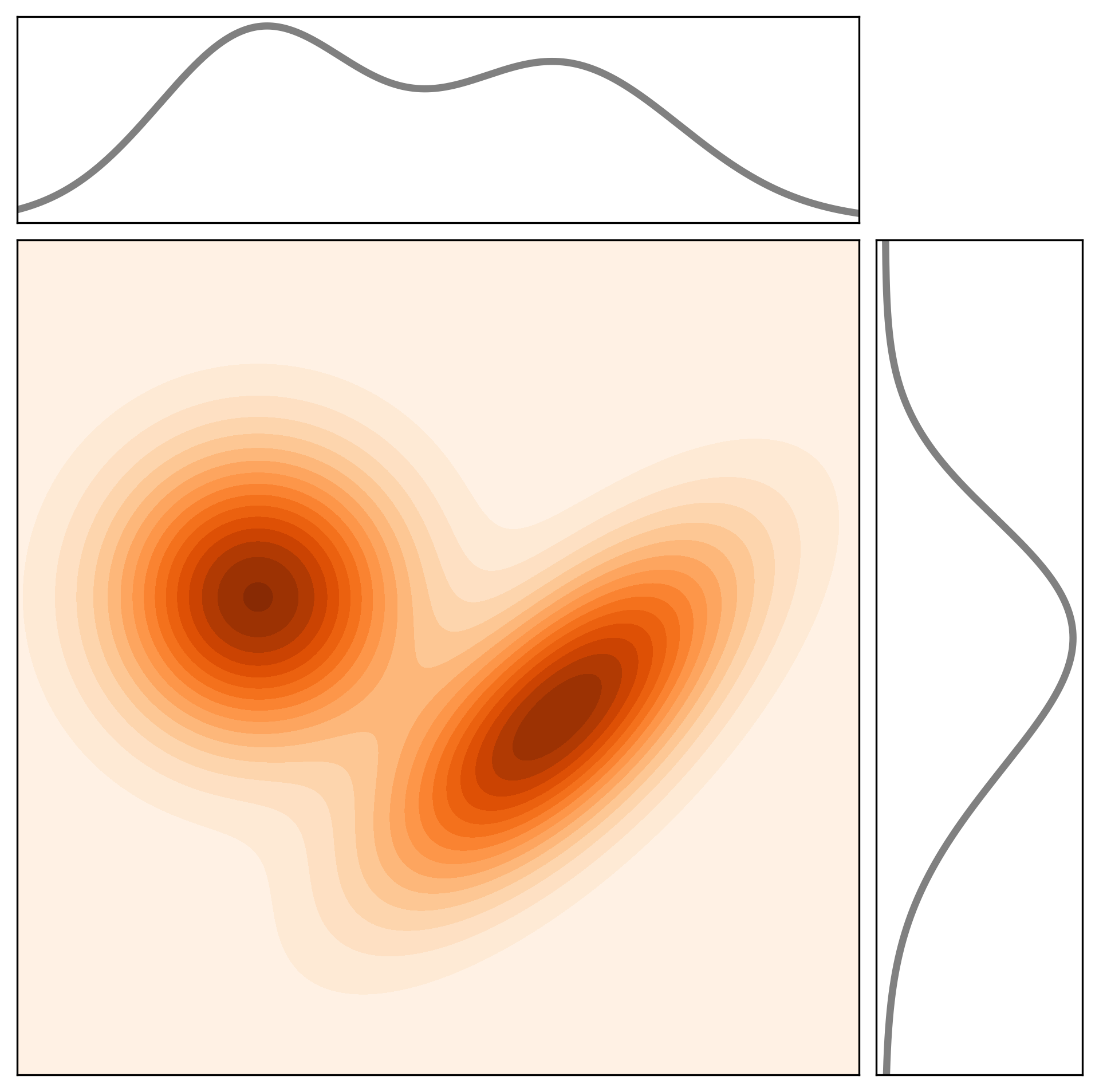}
  \end{center}
  \caption{Joint PDF $p_{X,Y}$ (in shades) and its marginals $p_X$ and $p_Y$ (in black lines).}
  \label{fig:joint}
\end{wrapfigure}

Given two random variables $X,Y\in \Real^d$, their joint PDF $p_{X,Y}(x,y)$ has marginals 
\begin{equation}\label{e:marginals}
 \int p_{X,Y}(x,y)\dd y = p_X(x) \text{ and } \int p_{X,Y}(x,y)\dd x = p_Y(y). 
\end{equation}
See \cref{fig:joint} for an illustration of the joint PDF of two RVs in $\Real$ ($d=1$). 
The \emph{conditional} PDF $p_{X|Y}$ describes the PDF of the random variable $X$ when conditioned on an event $Y=y$ with density $p_Y(y)>0$:
\begin{equation}\label{e:conditional}
    p_{X|Y}(x|y)\defe\frac{p_{X,Y}(x,y)}{p_Y(y)},
\end{equation} 
and similarly for the conditional PDF $p_{Y|X}$. Bayes' rule expresses the conditional PDF $p_{Y|X}$ with $p_{X|Y}$ by
\begin{equation}
    p_{Y|X}(y|x) = \frac{p_{X|Y}(x|y)p_Y(y)}{p_X(x)},
\end{equation}
for $p_X(x)>0$. 

The \emph{conditional expectation} $\E\brac{X | Y}$ is the best approximating \emph{function} $g_\star(Y)$ to $X$ in the least-squares sense: 
\begin{align}\nonumber
    g_\star &\defe  \argmin_{g:\Real^d\too\Real^d}\E\brac{\norm{X-g(Y)}^2} =  \argmin_{g:\Real^d\too\Real^d}\int \norm{x-g(y)}^2 p_{X,Y}(x,y)\dd x \dd y \\ \label{e:cond_E_g_star}
    &= \argmin_{g:\Real^d\too\Real^d} \int\brac{\textcolor{black}{\int \norm{x-g(y)}^2p_{X|Y}(x|y)\dd x}} p_Y(y)\dd y.
\end{align}
For $y\in \Real^d$ such that $p_Y(y)>0$ the conditional expectation function is therefore
\begin{equation}\label{e:cond_E_func}
    \E\brac{X|Y=y} \defe g_\star(y) = \int x p_{X|Y}(x|y) \dd x, %
\end{equation}
where the second equality follows from taking the minimizer of the inner brackets in \cref{e:cond_E_g_star} for $Y=y$, similarly to \cref{e:E}. 
Composing $g_\star$ with the random variable $Y$, we get 
\begin{equation}
    \E\brac{X|Y} \defe g_\star(Y), %
\end{equation}
which is a random variable in $\Real^d$.
Rather confusingly, both $\E\brac{X|Y=y}$ and $\E\brac{X|Y}$ are often called \emph{conditional expectation}, but these are different objects.
In particular, $\E\brac{X|Y=y}$ is a function $\Real^d\too\Real^d$, while $\E\brac{X|Y}$ is a random variable assuming values in $\Real^d$.
To disambiguate these two terms, our discussions will employ the notations introduced here.

The \emph{tower property} is an useful property that helps simplify derivations involving conditional expectations of two RVs $X$ and $Y$:
\begin{equation}\label{e:tower}
    \E\brac{\E\brac{X|Y}} = \E\brac{X} %
\end{equation}
Because $\E\brac{X|Y}$ is a RV, itself a function of the RV $Y$, the outer expectation computes the expectation of $\E\brac{X|Y}$.
The tower property can be verified by using some of the definitions above: 
\begin{align*}
    \E\brac{\E\brac{X|Y}} &= \int \parr{\int x p_{X|Y}(x|y) \dd x} p_Y(y) \dd y \\
    &\overset{\eqref{e:conditional}}{=} \int \int x p_{X,Y}(x,y) \dd x\dd y \\
    &\overset{\eqref{e:marginals}}{=} \int x p_X(x)\dd x \\
    &= \E \brac{X}.
\end{align*}

Finally, consider a helpful property involving two RVs $f(X, Y)$ and $Y$, where $X$ and $Y$ are two arbitrary RVs.
Then, by using the Law of the Unconscious Statistician with \eqref{e:cond_E_func}, we obtain the identity
\begin{equation}\label{e:f_x_y_cond_y}
    \E\brac{f(X,Y)|Y=y} = \int f(x,y) p_{X|Y}(x|y) \dd x.
\end{equation}

\subsection{Diffeomorphisms and push-forward maps}\label{ss:push-forward} 
We denote by $C^r(\Real^m,\Real^n)$ the collection of functions $f:\Real^m\too\Real^n$ with continuous partial derivatives of order $r$:
\begin{equation}
    \frac{\partial^r f^k}{\partial x^{i_1}\cdots \partial x^{i_r}}, \qquad k\in [n], i_j\in [m],    
\end{equation}
where $[n]\defe\set{1,2,\dots,n}$.
To keep notation concise, define also $C^r(\Real^n)\defe C^r(\Real^m,\Real)$ so, for example, $C^1(\Real^m)$ denotes the continuously differentiable scalar functions.
An important class of functions are the \highlight{$C^r$ diffeomorphism}; these are invertible functions $\psi\in C^r(\R^n,\R^n)$ with $\psi^{-1}\in C^r(\R^n,\R^n)$.

Then, given a RV $X\sim p_X$ with density $p_X$, let us consider a RV $Y=\psi(X)$, where $\psi:\Real^d\too\Real^d$ is a $C^1$ diffeomorphism.
The PDF of $Y$, denoted $p_Y$, is also called the \emph{push-forward} of $p_X$.
Then, the PDF $p_Y$ can be computed via a change of variables:
\begin{align*}
\E\brac{f(Y)} = \E\brac{f(\psi(X))} &= \int f(\psi(x)) p_X(x) \dd x = \int f(y) p_X(\psi^{-1}(y)) \abs{\det \partial_y \psi^{-1} (y)} \dd y,
\end{align*} 
where the third equality is due the change of variables $x=\psi^{-1}(y)$, $\partial_y \phi(y)$ denotes the Jacobian matrix (of first order partial derivatives), \ie, 
\begin{equation*}
 \brac{\partial_y \phi(y)}_{i,j} = \frac{\partial \phi^i}{\partial x^j}, \ i,j \in [d],
\end{equation*}
and $\det A$ denotes the determinant of a square matrix $A\in\Real^{d\times d}$. 
Thus, we conclude that the PDF $p_Y$ is
\begin{equation}
    p_Y(y) = p_X(\psi^{-1}(y))\abs{\det \partial_y \psi^{-1} (y)}.
\end{equation}
We will denote the push-forward operator with the symbol $\sharp$, that is 
\begin{equation}\label{e:push-forward_p}
     \brac{\psi_\sharp  p_X}(y) \defe p_X(\psi^{-1}(y))\abs{\det \partial_y \psi^{-1} (y)}.
\end{equation}

\begin{figure}
    \centering
    \begin{tabular}{ccc}
         \includegraphics[width=0.3\textwidth]{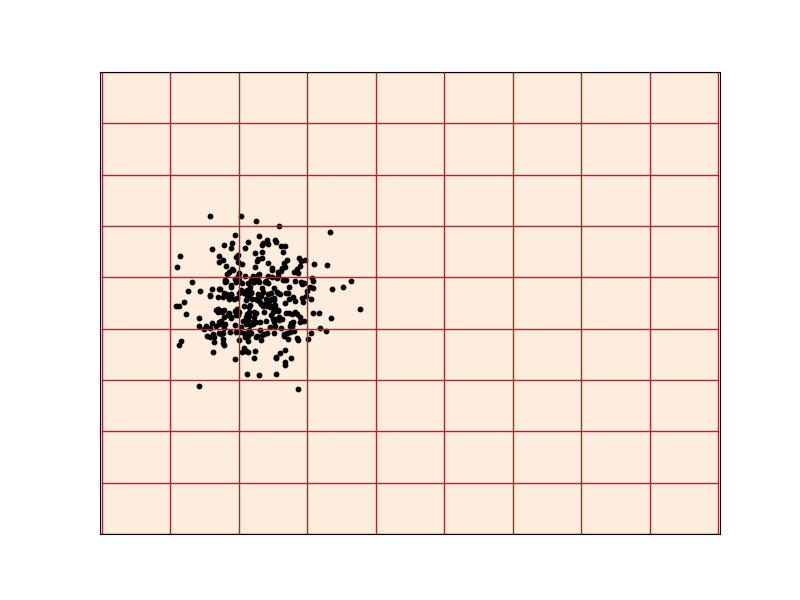} &
         \includegraphics[width=0.3\textwidth]{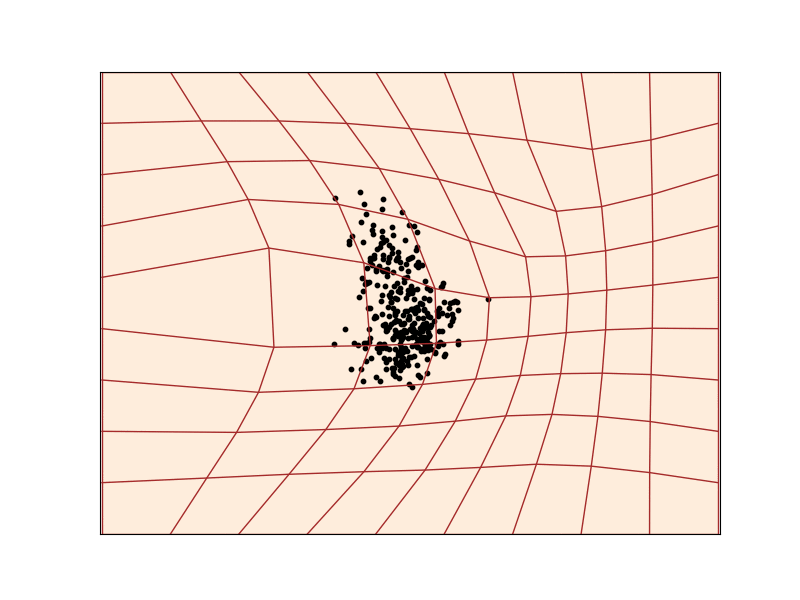} &
         \includegraphics[width=0.3\textwidth]{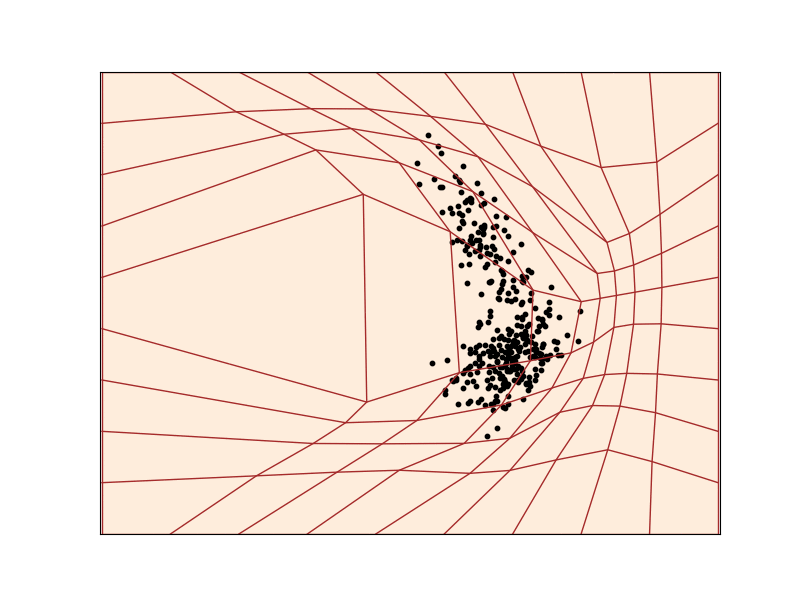} 
    \end{tabular}
    \caption{A flow model $X_t=\psi_t(X_0)$ is defined by a diffeomorphism $\psi_t:\Real^d\too \Real^d$ (visualized with a brown square grid) pushing samples from a source RV $X_0$ (left, black points) toward some target distribution $q$ (right). We show three different times $t$.}
    \label{fig:flow_model}
\end{figure}

\subsection{Flows as generative models}\label{ss:flows_and_velocities}

As mentioned in~\cref{s:quick_tour}, the goal of generative modeling is to transform samples $X_0 = x_0$ from a \highlight{source distribution} $p$ into samples $X_1=x_1$ from a \highlight{target distribution} $q$.
In this section, we start building the tools necessary to address this problem by means of a flow mapping $\psi_t$.
More formally, a $C^r$ \highlight{flow} is a time-dependent mapping $\psi:[0,1]\times \Real^d\too\Real^d$ implementing $\psi : (t,x) \mapsto \psi_t(x)$. 
Such flow is also a $C^r([0,1]\times\Real^{d},\Real^d)$ function, such that the function $\psi_t(x)$ is a $C^r$ diffeomorphism in $x$ for all $t \in [0, 1]$.
A \highlight{flow model} is a \emph{continuous-time Markov process} $(X_t)_{0 \leq t \leq 1}$ defined by applying a flow $\psi_t$ to the RV $X_0$:
\begin{myframe}
\begin{equation}\label{e:flow_model}
    X_t = \psi_t(X_0), \quad t\in [0,1], \text{ where } X_0\sim p. %
\end{equation}    
\end{myframe}

See \Cref{fig:flow_model} for an illustration of a flow model. To see why $X_t$ is Markov, note that, for any choice of $0\leq t < s \leq 1$, we have
\begin{equation}\label{e:flow_is_markov}
    X_s = \psi_s(X_0) = \psi_s(\psi_t^{-1}( \psi_t(X_0) ) ) = \psi_{s|t}(X_t),
\end{equation}
where the last equality follows from using \cref{e:flow_model} to set $X_t = \psi_t(X_0)$, and defining $\psi_{s|t}\defe \psi_s\circ \psi_t^{-1}$, which is also a diffeomorphism.
$X_s=\psi_{s|t}(X_t)$ implies that states later than $X_t$ depend only on $X_t$, so $X_t$ is Markov.
In fact, for flow models, this dependence is \emph{deterministic}.

In summary, the goal \highlight{generative flow modeling} is to find a flow $\psi_t$ such that 
\begin{equation}
    X_1 = \psi_1(X_0) \sim q. %
\end{equation}

\begin{figure}
    \centering
    \begin{tabular}{ccc}
         \includegraphics[width=0.3\textwidth]{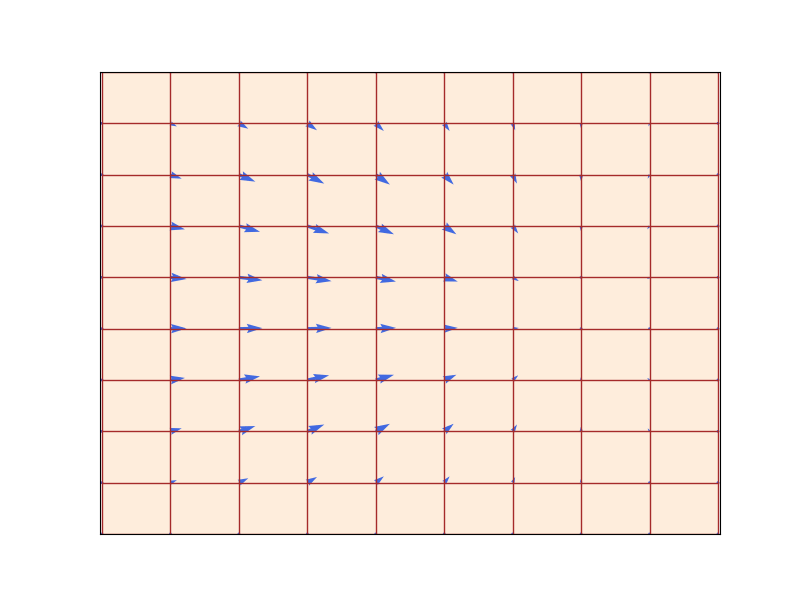} &
         \includegraphics[width=0.3\textwidth]{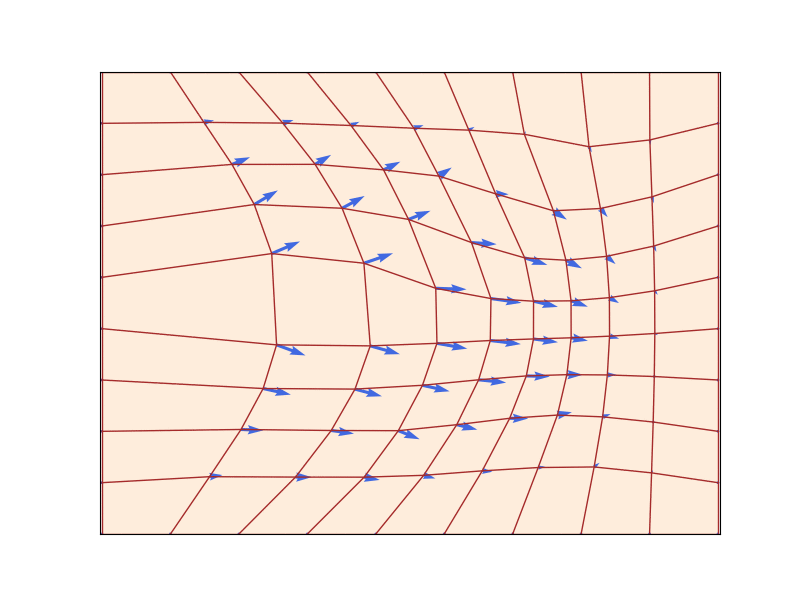} &
         \includegraphics[width=0.3\textwidth]{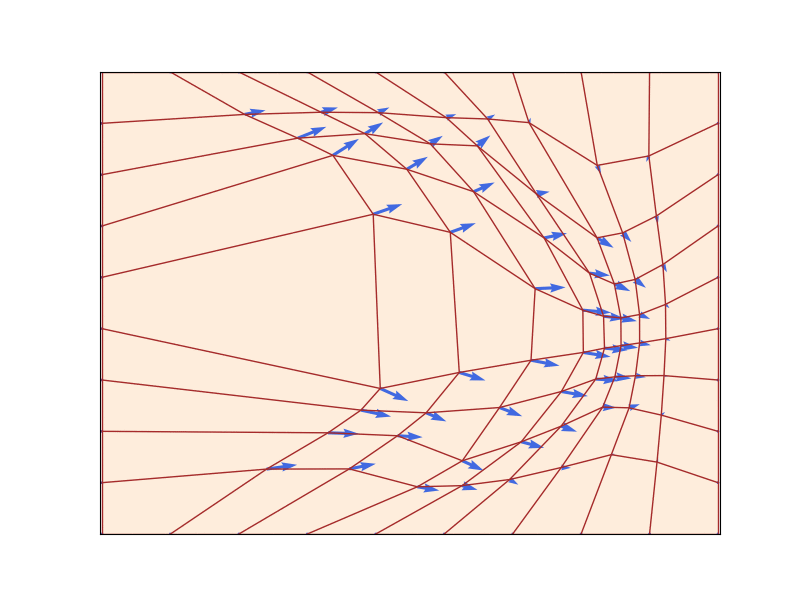} 
    \end{tabular}
    \caption{A flow $\psi_t:\Real^d\too \Real^d$ (square grid) is defined by a velocity field $u_t :\Real^d\too\Real^d$ (visualized with blue arrows) that prescribes its instantaneous movements at all locations. We show three different times $t$.}
    \label{fig:flow}
\end{figure}
\subsubsection{Equivalence between flows and velocity fields}\label{sec:equivalence_flows_velocities}
A $C^r$ flow $\psi$ can be defined in terms of a $C^r([0,1]\times \Real^d,\Real^d)$ \emph{velocity field} $u:[0,1]\times \Real^d \too \Real^d$ implementing $u : (t, x) \mapsto u_t(x)$ via the following ODE:
\begin{subequations}\label{e:flow}\vspace{-10pt}
    \begin{align} 
      \frac{\dd}{\dd t}\psi_{t}(x) &= u_t(\psi_{t}(x)) & \text{(flow ODE)}\label{e:flow_flow}\\
      \psi_{0}(x)             &= x                & \text{(flow initial conditions)}\label{e:flow_boundary} 
    \end{align}
\end{subequations}    
See \cref{fig:flow} for an illustration of a flow together with its velocity field.

A standard result regarding the existence and uniqueness of solutions $\psi_t(x)$ to \cref{e:flow} is (see \eg, \cite{perko2013differential,coddington1956theory}): 
\begin{myframe}
    \begin{theorem}[Flow local existence and uniqueness]\label{thm:ode_existence_and_uniqueness}
    If $u$ is $C^r([0,1]\times\Real^{d},\Real^d)$, $r\geq 1$ (in particular, locally Lipschitz), then the ODE in \eqref{e:flow} has a unique solution which is a  $C^r(\Omega,\Real^d)$ diffeomorphism $\psi_t(x)$ defined over an open set $\Omega$ which is super-set of $\set{0}\times \Real^d$.
    \end{theorem}
\end{myframe}
This theorem guarantees only the \emph{local} existence and uniqueness of a $C^r$ flow moving each point $x\in\Real^d$ by $\psi_t(x)$ during a potentially limited amount of time $t\in[0,t_x)$.
To guarantee a solution until $t=1$ for all $x\in\Real^d$, one must place additional assumptions beyond local Lipschitzness.
For instance, one could consider global Lipschitness, guaranteed by bounded first derivatives in the $C^1$ case.
However, we will later rely on a different condition---namely, integrability---to guarantee the existence of the flow almost everywhere, and until time $t=1$.

So far, we have shown that a velocity field uniquely defines a flow.
Conversely, given a $C^1$ flow $\psi_t$, one can extract its defining velocity field $u_t(x)$ for arbitrary $x \in \Real^d$ by considering the equation $\frac{\dd}{\dd t} \psi_t(x') = u_t(\psi_t(x'))$, and using the fact that $\psi_t$ is an invertible diffeomorphism for every $t \in [0, 1]$ to let $x'=\psi^{-1}_t(x)$.
Therefore, the unique velocity field $u_t$ determining the flow $\psi_t$ is
\begin{equation}\label{e:u_from_psi}
    u_t(x) = \dot{\psi}_t( \psi^{-1}_t(x)),
\end{equation}
where $\dot{\psi}_t\defe \frac{\dd}{\dd t} \psi_t$.
In conclusion, we have shown the equivalence between $C^r$ flows $\psi_t$ and $C^r$ velocity fields $u_t$.

\subsubsection{Computing target samples from source samples}
Computing a target sample $X_1$---or, in general, any sample $X_t$---entails approximating the solution of the ODE in~\cref{e:flow} starting from some initial condition $X_0=x_0$.
Numerical methods for ODEs is a classical and well researched topic in numerical analysis, and a myriad of powerful methods exist \citep{iserles2009first}.
One of the simplest methods is the \emph{Euler method}, implementing the update rule
\begin{equation}\label{e:euler_method}
    X_{t+h} = X_t + h u_t(X_t) %
\end{equation}
where $h=n^{-1}>0$ is a step size hyper-parameter with $n \in \Nat$.
To draw a sample $X_1$ from the target distribution, apply the Euler method starting at some $X_0 \sim p$ to produce the sequence $X_h,X_{2h},\ldots,X_1$.
The Euler method coincides with first-order Taylor expansion of $X_t$:
$$X_{t+h}=X_t + h \dot{X}_t + o(h)=X_t + h u_t(X_t) + o(h),$$
where $o(h)$ stands for a function growing slower than $h$, that is, $o(h)/h \too 0$ as $h\too 0$.
Therefore, the Euler method accumulates $o(h)$ error per step, and can be shown to accumulate $o(1)$ error after $n=1/h$ steps.
Therefore, the error of the Euler method vanishes as we consider smaller step sizes $h\too 0$.
The Euler method is just one example among many ODE solvers.
\Cref{ex:euler_method} exemplifies another alternative, the second-order \emph{midpoint method}, which often outperforms the Euler method in practice.

\begin{pbox}[label={ex:euler_method}]{Computing $X_1$ with Midpoint solver}
\begin{minted}[linenos, breaklines, mathescape, fontsize=\footnotesize, xleftmargin=2em]{python}
from flow_matching.solver import ODESolver
from flow_matching.utils import ModelWrapper

class Flow(ModelWrapper):
    def __init__(self, dim=2, h=64):
        super().__init__()
        self.net = torch.nn.Sequential(
            torch.nn.Linear(dim + 1, h), torch.nn.ELU(),
            torch.nn.Linear(h, dim))

    def forward(self, x, t):
        t = t.view(-1, 1).expand(*x.shape[:-1], -1)
        return self.net(torch.cat((t, x), -1))

velocity_model = Flow()

... # Optimize the model parameters s.t. model(x_t, t) = $u_t(X_t)$

x_0 = torch.randn(batch_size, *data_dim)  # Specify the initial condition

solver = ODESolver(velocity_model=velocity_model)
num_steps = 100
x_1 = solver.sample(x_init=x_0, method='midpoint', step_size=1.0 / num_steps)
\end{minted}
\end{pbox}

\begin{figure}
    \centering
    \begin{tabular}{ccc}
         \includegraphics[width=0.3\textwidth]{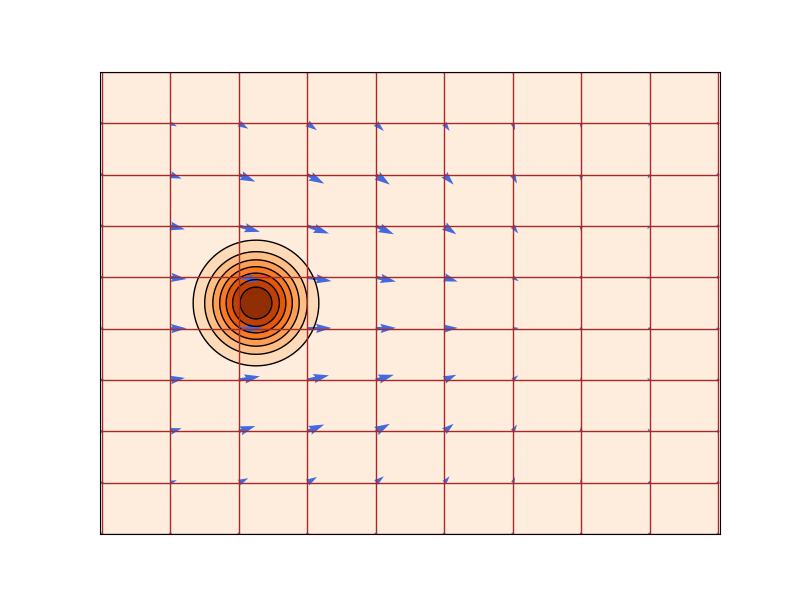} &
         \includegraphics[width=0.3\textwidth]{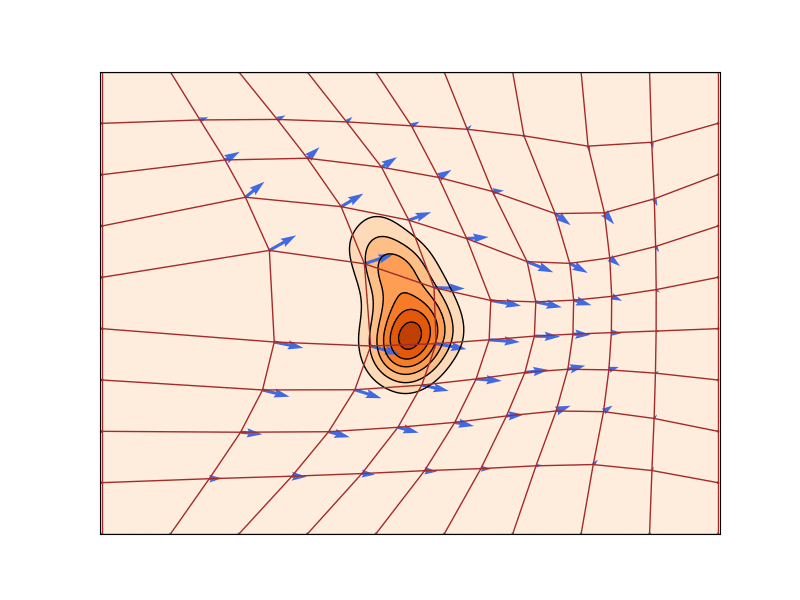} &
         \includegraphics[width=0.3\textwidth]{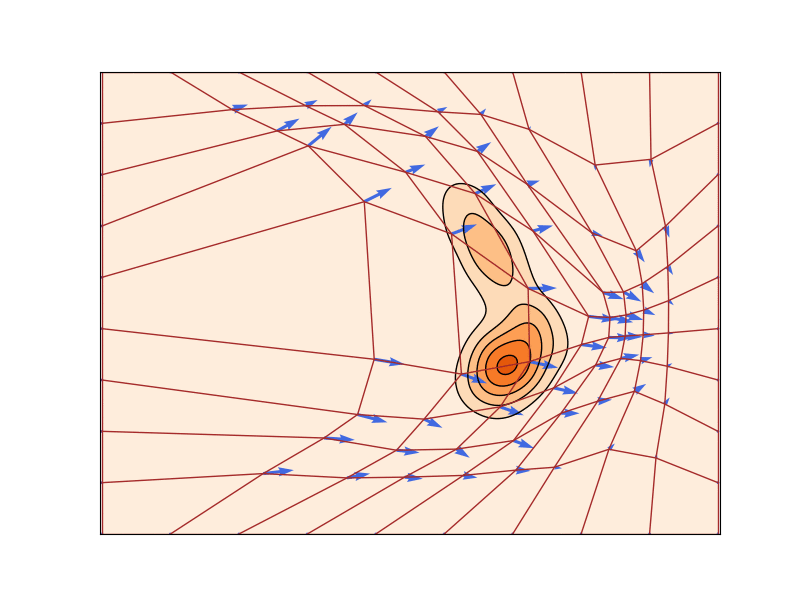} 
    \end{tabular}
    \caption{ A velocity field $u_t$ (in blue) \emph{generates} a probability path $p_t$ (PDFs shown as contours) if the flow defined by $u_t$ (square grid) reshapes $p$ (left) to $p_t$ at all times $t\in [0,1)$.}
    \label{fig:ut_generates_pt}
\end{figure}

\pagebreak
\subsection{Probability paths and the Continuity Equation} \label{ss:continuity_equation}
We call a time-dependent probability $(p_t)_{0\leq t \leq 1}$ a \highlight{probability path}.
For our purposes, one important probability path is the marginal PDF of a flow model $X_t = \psi_t(X_0)$ at time $t$:
\begin{equation}
    X_t\sim p_t. %
\end{equation}
For each time $t \in [0,1]$, these marginal PDFs are obtained via the push-forward formula in \cref{e:push-forward_p}, that is,
\begin{equation}\label{e:p_t_flow}
    p_t(x) = \brac{\psi_{t\sharp}p}(x). %
\end{equation}
Given some arbitrary probability path $p_t$ we define 

\begin{myframe}
\begin{equation}\label{def:generates}
  u_t \text{ \highlight{generates} } p_t \text{ if } X_t = \psi_t(X_0) \sim p_t \text{ for all } t \in[0,1).
\end{equation}
\end{myframe}
In this way, we establish a close relationship between velocity fields, their flows, and the generated probability paths, see Figure \ref{fig:ut_generates_pt} for an illustration. Note that we use the time interval $[0,1)$, open from the right, to allow dealing with target distributions $q$ with compact support where the velocity is not defined precisely at $t=1$.

To verify that a velocity field $u_t$ generates a probability path $p_t$, one can verify if the pair $(u_t, p_t)$ satisfies a partial differential equation (PDE) known as the \emph{Continuity Equation}:
\begin{equation}\label{e:continuity}
    \frac{\dd}{\dd t} p_t(x)+ \divv(p_t u_t)(x) = 0, %
\end{equation}
where $\divv(v)(x) = \sum_{i=1}^d \partial_{x^i} v^i (x)$, and $v(x)=(v^1(x),\ldots,v^d(x))$.
The following theorem, a rephrased version of the \emph{Mass Conservation Formula} \citep{villani2009optimal}, states that a solution $u_t$ to the Continuity Equation generates the probability path $p_t$:
\begin{myframe}
\begin{theorem}[Mass Conservation]\label{thm:continuity}
    Let $p_t$ be a probability path and $u_t$ a locally Lipchitz integrable vector field.
    Then, the following two statements are equivalent:
    \begin{enumerate}
        \item The Continuity Equation \eqref{e:continuity} holds for $t\in [0,1)$.
        \item $u_t$ generates $p_t$, in the sense of \eqref{def:generates}.
    \end{enumerate}
      
\end{theorem}
\end{myframe}
In the previous theorem, local Lipschitzness assumes that there exists a local neighbourhood over which $u_t(x)$ is Lipschitz, for all $(t, x)$. 
Assuming that $u$ is integrable means that:
\begin{equation}\label{e:integrable}
 \int_0^1\int \norm{u_t(x)}p_t(x)\dd x \dd t < \infty. %
\end{equation}
Specifically, integrating a solution to the flow ODE \eqref{e:flow_flow} across times $[0,t]$ leads to the integral equation
\begin{equation}
    \psi_t(x) = x + \int_0^t u_s(\psi_s(x))\dd s.
\end{equation}
Therefore, integrability implies 
\begin{align*}
    \E \norm{X_t} &\overset{\eqref{e:flow_model}}{=} \int \norm{\psi_t(x)}p(x)\dd x \\
    &\,\,\,\,= \int \norm{x + \int_0^t u_s(\psi_s(x))ds}  p(x)\dd x\\
    &\,\,\,\, \overset{(i)}{\leq} \E \norm{X_0} + \int_0^1\int \norm{u_s(x)}p_t(x)\dd t \\
    &\,\,\,\,  \overset{(ii)}{<} \infty,
\end{align*}
where (i) follows from the triangle inequality, and (ii) assumes the integrability condition \eqref{e:integrable} and $\E \norm{X_0} < \infty$.
In sum, integrability allows assuming that $X_t$ has bounded expected norm, if $X_0$ also does.

To gain further insights about the meaning of the Continuity Equation, we may write it in \emph{integral form} by means of the Divergence Theorem---see \citet{matthews2012vector} for an intuitive exposition, and \citet{loomis1968advanced} for a rigorous treatment.
This result states that, considering some domain $\gD$ and some smooth vector field $u:\Real^d\too\Real^d$, accumulating the divergences of $u$ inside $\gD$ equals the \emph{flux} leaving $\gD$ by orthogonally crossing its boundary $\partial \gD$:
\begin{equation}
    \int_\gD \divv(u)(x)\dd x = \int_{\partial \gD} \ip{u(y),n(y)}\dd s_y,
\end{equation}
\begin{wrapfigure}[16]{r}{0.27\textwidth}
  \begin{center}    \includegraphics[width=0.25\textwidth]{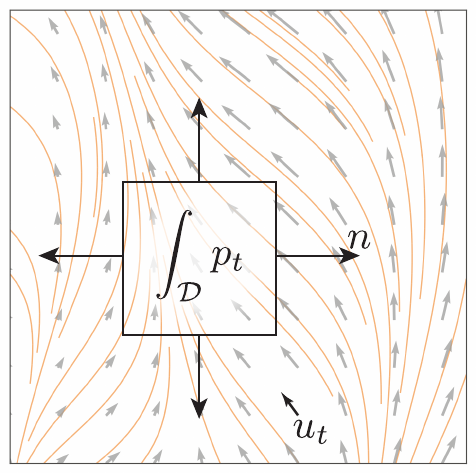}
  \end{center}
  \caption{The continuity equation asserts that the local change in probability equals minus the net outgoing probability flux.}\label{fig:continuity}
\end{wrapfigure} 
where $n(y)$ is a unit-norm normal field pointing outward to the domain's boundary $\partial\gD$, and $\dd s_y$ is the boundary's area element.
To apply these insights to the Continuity Equation, let us integrate \eqref{e:continuity} over a small domain $\gD\subset\Real^d$ (for instance, a cube) and apply the Divergence Theorem to obtain
\begin{equation}\label{ea:continuity_intuitive}
    \frac{\dd}{\dd t} \int_\gD p_t(x) \dd x = -\int_\gD \divv(p_t u_t)(x) \dd x = -\int_{\partial \gD}  \ip{p_t(y) u_t(y),n(y)}\dd s_y.
\end{equation}
This equation expresses the rate of change of total probability mass in the volume $\gD$ (left-hand side) as the negative probability \emph{flux} leaving the domain (right-hand side).
The probability flux, defined as $j_t(y)=p_t(y)u_t(y)$, is the probability mass flowing through the hyperplane orthogonal to $n(y)$ per unit of time and per unit of (possibly high-dimensional) area.
See \cref{fig:continuity} for an illustration.

\subsection{Instantaneous Change of Variables}

One important benefit of using flows as generative models is that they allow the tractable computation of \emph{exact} likelihoods $\log p_1(x)$, for all $x \in \Real^d$.
This feature is a consequence of the Continuity Equation called the \emph{Instantaneous Change of Variables}~\citep{chen2018neural}:
\begin{equation}
    \label{e:instant_div} \frac{\dd}{\dd t} \log p_t(\psi_t (x)) = -\divv(u_t)(\psi_t(x)).%
\end{equation}
This is the ODE governing the change in log-likelihood, $\log p_t(\psi_t(x))$, along a sampling trajectory $\psi_t(x)$ defined by the flow ODE \eqref{e:flow_flow}.
To derive \eqref{e:instant_div}, differentiate $\log p_t(\psi_t(x))$ with respect to time, and apply both the Continuity Equation \eqref{e:continuity} and the flow ODE \eqref{e:flow_flow}.
Integrating \eqref{e:instant_div} from $t=0$ to $t=1$ and rearranging, we obtain 
\begin{equation}\label{e:div_int}
    \log p_1(\psi_1(x)) = \log p_0(\psi_0(x))-\int_0^1 \divv(u_t)(\psi_t(x))\dd t.
\end{equation}
In practice, computing $\divv (u_t)$, which equals the trace of the Jacobian matrix $\partial_x u_t(x) \in \Real^{d\times d}$, is increasingly challenge as the dimensionality $d$ grows. 
Because of this reason, previous works employ unbiased estimators such as Hutchinson's trace estimator~\citep{grathwohl2018ffjord}:
\begin{equation}
    \divv (u_t)(x) = \trace \brac{\partial_x u_t(x)} = \E_Z\, \trace \brac{ Z^T \partial_x u_t(x) Z}, 
\end{equation}
where $Z\in\Real^{d\times d}$ is any random variable with $\E\brac{Z}=0$ and $\Cov\parr{Z,Z}=I$, (for example, $Z\sim\mathcal{N}(0,I)$), and $\trace[Z] = \sum_{i=1}^d Z_{i,i}$.
By plugging the equation above into \eqref{e:div_int} and switching the order of integral and expectation, we obtain the following unbiased log-likelihood estimator:
\begin{equation}
    \log p_1(\psi_1(x)) = \log p_0(\psi_0(x))-\E_Z\, \int_0^1  \trace \brac{ Z^T \partial_x u_t(\psi_t(x)) Z} \dd t.
\end{equation}
In contrast to $\divv(u_t)(\psi_t(x))$ in \eqref{e:instant_div}, computing  $\trace\brac{Z^T \partial_x u_t(\psi_t(x)) Z}$ for a fixed sample $Z$ in the equation above can be done with a single backward pass via a vector-Jacobian product (JVP)\footnote{E.g., see \url{https://pytorch.org/docs/stable/generated/torch.autograd.functional.vjp.html}.}.

In summary, computing an unbiased estimate of $\log p_1(x)$ entails simulating the ODE 
\begin{subequations}\label{e:log_p_1_ode_unbiased}
    \begin{align}
    \frac{\dd}{\dd t} \brac{\begin{matrix}
        f(t) \\ g(t)
    \end{matrix}} &= \brac{\begin{matrix}
         u_t(f(t)) \\
         -\trace\brac{Z^T \partial_x u_t(f(t))Z}
    \end{matrix}},\\
    \brac{\begin{matrix}
        f(1) \\ g(1)
    \end{matrix}} &= \brac{\begin{matrix}
        x\\ 0
    \end{matrix}},
    \end{align}
\end{subequations}
backwards in time, from $t=1$ to $t=0$, and setting:
\begin{equation}\label{e:log_p_1_unbiased}
    \widehat{\log p}_1(x) = \log p_0(f(0)) - g(0).
\end{equation}
See~\cref{ex:sample_with_likelihood} for an example on how to obtain log-likelihood estimates from a flow model using the \fmlibrary.

\begin{pbox}[label={ex:sample_with_likelihood}]{Computing the likelihood}
\begin{minted}[linenos, breaklines, mathescape, fontsize=\footnotesize, xleftmargin=2em]{python}
from flow_matching.solver import ODESolver
from flow_matching.utils import ModelWrapper
from torch.distributions.normal import Normal

velocity_model: ModelWrapper = ...  # Train the model parameters s.t. model(x_t, t) = $u_t(x_t)$

x_1 = torch.randn(batch_size, *data_dim)  # Point $X_1$ where we wish to compute $\log p_1(x)$

# Define $\log p_0(x)$
gaussian_log_density = Normal(torch.zeros(size=data_dim), torch.ones(size=data_dim)).log_prob

solver = ODESolver(velocity_model=velocity_model) 
num_steps = 100
x_0, log_p1 = solver.compute_likelihood(
    x_1=x_1, 
    method='midpoint', 
    step_size=1.0 / num_steps, 
    log_p0=gaussian_log_density
)
\end{minted}
\end{pbox}

\subsection{Training flow models with simulation}

The Instantaneous Change of Variables, and the resulting ODE system \eqref{e:log_p_1_ode_unbiased}, allows training a flow model by maximizing the log-likelihood of training data~\citep{chen2018neural,grathwohl2018ffjord}.
Specifically, let $u^\theta_t$ be a velocity field with learnable parameters $\theta\in\Real^p$, and consider the problem of learning $\theta$ such that 
\begin{equation}
    p^\theta_1\approx q.
\end{equation}
We can pursue this goal, for instance, by minimizing the KL-divergence of $p^\theta_1$ and $q$: 
\begin{equation}\label{e:loss_kl}
    \gL(\theta) = D_{\text{KL}}(q  , p^\theta_1) = -\E_{Y\sim q} \log p^\theta_1(Y) + \text{constant},
\end{equation}
where $p^\theta_1$ is the distribution of $X_1=\psi^\theta_1(X_0)$, $\psi_t^\theta$ is defined by $u_t^\theta$, and we can obtain an unbiased estimate of $\log p^\theta_1(Y)$ via the solution to the ODE system~\eqref{e:log_p_1_ode_unbiased}.
However, computing this loss---as well as its gradients---requires precise ODE simulations during training, where only errorless solutions constitute unbiased gradients.
In contrast, \emph{Flow Matching}, presented next, is a simulation-free framework to train flow generative models without the need of solving ODEs during training.

\pagebreak

\section{Flow Matching}\label{s:fm_continuous}
Given a source distribution $p$ and a target distribution $q$, Flow Matching (FM) \citep{lipman2022flow,liu2022flow,albergo2022building} is a scalable approach for training a flow model, defined by a learnable velocity $u_t^\theta$, and solving the \highlight{Flow Matching Problem}:
\begin{myframe}
    \begin{equation}\label{prob:gen_flow}
    \text{Find } u^\theta_t \text{ generating } p_t, \text{ with } p_0=p \text{ and } p_1=q.
\end{equation}
\end{myframe}
In the equation above, ``generating'' is in the sense of \cref{def:generates}.
Revisiting the Flow Matching \emph{blueprint} from~\cref{fig:blueprint}, the FM framework (a) identifies a known source distribution $p$ and an unknown data target distribution $q$, (b) prescribes a probability path $p_t$ interpolating from $p_0=p$ to $p_1=q$, (c) learns a velocity field $u^\theta_t$ implemented in terms of a neural network and generating the path $p_t$, and (d) samples from the learned model by solving an ODE with $u^\theta_t$. 
To learn the velocity field $u^\theta_t$ in step (c), FM minimizes the regression loss: 
\begin{equation}\label{e:loss_fm}
     \gL_{\FM}(\theta) = \E_{X_t\sim p_t} D\parr{ u_t(X_t) , u^\theta_t(X_t)}, 
\end{equation}
where $D$ is a dissimilarity measure between vectors, such as the squared $\ell_2$-norm $D(u, v) = \norm{u-v}^2$.
Intuitively, the FM loss encourages our learnable velocity field $u_t^\theta$ to match the ground truth velocity field $u_t$ known to generate the desired probability path $p_t$.
\Cref{fig:diagram} depicts the main objects in the Flow Matching framework and their dependencies.
Let us start our exposition of Flow Matching by describing how to build $p_t$ and $u_t$, as well as a practical implementation of the loss~\eqref{e:loss_fm}.

\subsection{Data} 

To reiterate, let source samples be a RV $X_0 \sim p$ and target samples a RV $X_1 \sim q$.
Commonly, source samples follow a known distribution that is easy to sample, and target samples are given to us in terms of a dataset of finite size.
Depending on the application, target samples may constitute images, videos, audio segments, or other types of high-dimensional, richly structured data.
Source and target samples can be independent, or originate from a general joint distribution known as the \highlight{coupling}
\begin{align}
    (X_0,X_1) &\sim \pi_{0,1} (X_0,X_1), %
\end{align}
where, if no coupling is known, the source-target samples are following the independent coupling $\pi_{0,1} (X_0,X_1) = p(X_0)q(X_1)$.
One common example of independent source-target distributions is to consider the generation of images $X_1$ from random Gaussian noise vectors $X_0\sim \gN(0,I)$.
As an example of a dependent coupling, consider the case of producing high-resolution images $X_1$ from their low resolution versions $X_0$, or producing colorized videos $X_1$ from their gray-scale counterparts $X_0$.

\subsection{Building probability paths} 

Flow Matching drastically simplifies the problem of designing a probability path $p_t$---together with its corresponding velocity field $u_t$---by adopting a \emph{conditional} strategy.
As a first example, consider conditioning the design of $p_t$ on a single target example $X_1=x_1$, yielding the \highlight{conditional probability path} $p_{t|1}(x|x_1)$ illustrated in \cref{fig:p_t}.
Then, we may construct the overall, \highlight{marginal probability path} $p_t$ by aggregating such conditional probability paths $p_{t|1}$:
\begin{equation}\label{e:p_t}
    p_t(x) = \int p_{t|1}(x|x_1)q(x_1) \dd x_1, %
\end{equation}
as illustrated in \cref{fig:marginal_p_t}.
To solve the Flow Matching Problem, we would like $p_t$ to satisfy the following {boundary conditions}:
\begin{equation}\label{e:p_q_interp}
    p_0=p, \quad p_1=q, %
\end{equation}
that is, the marginal probability path $p_t$ interpolates from the source distribution $p$ at time $t=0$ to the target distribution $q$ at time $t=1$. 
These boundary conditions can be enforced by requiring the conditional probability paths to satisfy
\begin{equation}\label{e:p_t_cond_boundary}
    p_{0|1}(x|x_1) = \pi_{0|1}(x|x_1), \text{ and  } p_{1|1}(x|x_1)=\delta_{x_1}(x),
\end{equation}
where the conditional coupling $\pi_{0|1}(x_0|x_1)=\pi_{0,1}(x_0,x_1)/q(x_1)$ and $\delta_{x_1}$ is the delta measure centered at $x_1$.
For the independent coupling $\pi_{0,1}(x_0,x_1)=p(x_0)q(x_1)$, the first constraint above reduces to $p_{0|1}(x|x_1)=p(x)$.
Because the delta measure does not have a density, the second constraint should be read as $\int p_{t|1}(x|y)f(y)\dd y \to f(x)$ as $t \to 1$ for continuous functions $f$.
Note that the boundary conditions \eqref{e:p_q_interp} can be verified plugging \eqref{e:p_t_cond_boundary} into \eqref{e:p_t}.

A popular example of a conditional probability path satisfying the conditions in \eqref{e:p_t_cond_boundary} was given in \eqref{e:condot_path}:
\begin{equation*}
  \gN(\cdot\, |\, t x_1, (1-t)^2I)\too \delta_{x_1}(\cdot) \text{ as } t \to 1.
\end{equation*}

\begin{figure}[t]
\begin{center}
\resizebox{\textwidth}{!}{%
\begin{tikzpicture}
  \newcommand{\sep}{3.5}
  \pgfmathsetmacro{\myinnersep}{2}%
  \tikzset{
  box/.style={fill=metabg, rounded corners=0.2cm, minimum width=2cm, inner sep=0.25cm, minimum height=0.8cm}
}
  \node[box] (psi0) at (0, +2.25) {Flow};
  \node[] (psi1)    at (0, -0) {$\psi_t(x)$};
  \node[] (psi2)    at (0, -3) {$\psi_t(x | x_1)$};
  \node[] (psi3)    at (0, -6) {$t x_1 + (1 - t) x$};
  
  \node[box] (u0)   at (\sep, +2.25) {Velocity field};
  \node[] (u1)      at (\sep, -0) {$u_t(x)$};
  \node[] (u2)      at (\sep, -3) {$u_t(x | x_1)$};
  \node[] (u3)      at (\sep, -6) {$(x_1 - x) / (1-t)$};
  
  \node[box] (p0)   at (2*\sep, +2.25) {{Probability path}};
  \node[] (p1)      at (2*\sep, -0) {$p_t(x)$};
  \node[] (p2)      at (2*\sep, -3) {$p_t(x | x_1)$};
  \node[] (p3)      at (2*\sep, -6) {$\gN(x | tx_1, (1-t)^2 I)$};
  
  \node[box] (b0)   at (3*\sep, +2.25)   {{Boundary conds.}};
  \node[] (b1)      at (3*\sep, -0)   {$p_0=p$};
  \node[] (b1b)     at (3*\sep, -0.5) {$p_1=q$};
  \node[] (b2)      at (3*\sep, -3)   {$p_0=p$};
  \node[] (b2b)     at (3*\sep, -3.5) {$p_1 = \delta_{x_1}$};
  \node[] (b3)      at (3*\sep, -6)   {$p_0=\gN(0, I)$};
  \node[] (b3b)     at (3*\sep, -6.5) {$p_1 = \delta_{x_1}$};
  
  \node[box] (l0)   at (4*\sep, +2.25) {{Loss}};
  \node[] (l1)      at (4*\sep, -0) {\highlight{Flow Matching (FM)} \eqref{e:fm_loss}};
  \node[] (l1)      at (4*\sep, -0.5) {$D\parr{u_t(X_t), u^\theta_t(X_t)}$};
  \node[] (l2)      at (4*\sep, -3) {\highlight{Conditional FM (CFM)} \eqref{e:cfm_loss_main}};
  \node[] (l2)      at (4*\sep, -3.5) {$D\parr{u_t(X_t|X_1), u^\theta_t(X_t)}$};
  \node[] (l3)      at (4*\sep, -6) {\highlight{OT, Gauss CFM} \eqref{eq:gauss}};
  \node[] (l3)      at (4*\sep, -6.5) {$\| u^\theta_t(X_t) - (X_1 - X_0)\|^2$};

  \draw[->, black, thick]    (psi1.north east) -- (u1.north west)   node [pos=0.5, above] {\scriptsize differentiation};
  \draw[->, metablue, thick] (u1.south west)   -- (psi1.south east) node [pos=0.5, below] {\scriptsize solve ODE};
  
  \draw[->, metablue, thick] (psi2.north east) -- (u2.north west)   node [pos=0.5, above] {\scriptsize differentiation};
  \draw[->, black, thick]    (u2.south west)   -- (psi2.south east) node [pos=0.5, below] {\scriptsize solve ODE};
  
  \draw[->, black, thick] (u1.north east) -- (p1.north west)   node [pos=0.5, above] {\scriptsize Continuity \eqref{e:continuity}};
  \draw[->, black, thick] (p1.south west) -- (u1.south east)   node [pos=0.5, below] {\scriptsize non-unique solution};
  
  \draw[->, black, thick] (u2.north east) -- (p2.north west)   node [pos=0.5, above] {\scriptsize Continuity \eqref{e:continuity}};
  \draw[->, black, thick] (p2.south west) -- (u2.south east)   node [pos=0.5, below] {\scriptsize non-unique solution};
  
  \draw[->, metablue, thick] (u2.north) -- (u1.south)   node [sloped, pos=0.5, below] {\scriptsize cond. expectation};
  
  \draw[->, black   , thick, dotted, decorate,decoration={snake,amplitude=3pt,pre length=2pt,post length=3pt}] (psi2.north) -- (psi1.south)   node [sloped, pos=0.5, above, yshift=0.2cm] {\scriptsize ???};
  
  \draw[->, black, thick]    ($(p1.south) - (0.25, 0)$) -- ($(p2.north) - (0.25, 0)$) node [sloped, pos=0.5, below] {\scriptsize conditioning}; 
  \draw[->, metablue, thick] ($(p2.north) + (0.25, 0)$) -- ($(p1.south) + (0.25, 0)$) node [sloped, pos=0.5, below] {\scriptsize marginalization}; 
  
  \draw[->, black, thick, dotted] (psi2.south) -- (psi3.north);
  \draw[->, black, thick, dotted] (u2.south)   -- (u3.north);
  \draw[->, black, thick, dotted] (p2.south)   -- (p3.north);
  \draw[->, black, thick, dotted] (b2b.south)  -- (b3.north);

  \draw[black, thick]     (psi1.north)                   -- ($(psi1.north) + (0.0, 0.75)$);
  \draw[->, black, thick] ($(p1.north) + (0.0, 0.75)$)   -- (p1.north);
  \draw[black, thick]     ($(psi1.north) + (0.0, 0.75)$) -- ($(p1.north) + (0.0, 0.75)$) node [pos=0.5, above] {\scriptsize push-forward $X_0$};
  
  \draw[metablue, thick]     ($(psi2.south) + (0.25, -0)$)    -- ($(psi2.south) + (0.25, -0.75)$);
  \draw[->, metablue, thick] ($(p2.south) - (0.25, +0.75)$)   -- ($(p2.south) - (0.25, 0)$);
  \draw[metablue, thick]     ($(psi2.south) + (0.25, -0.75)$) -- ($(p2.south) - (0.25, +0.75)$) node [pos=0.5, below] {\scriptsize \colorbox{white}{push-forward $X_0$}};
\end{tikzpicture}}
\end{center}
\caption{
  Main objects of the Flow Matching framework and their relationships.
  A \colorbox{metabg} {Flow} is represented with a \colorbox{metabg}{Velocity field} defining a random process generating a \colorbox{metabg}{Probability path}. The main idea of Flow Matching is to break down the construction of a complex flow satisfying the desired \colorbox{metabg}{Boundary conditions} (top row) to  conditional flows (middle row) satisfying simpler \colorbox{metabg}{Boundary conditions} and consequently easier to solve. The arrows indicate dependencies between different objects;  \highlight{Blue arrows} signify relationships employed by the Flow Matching framework. The \colorbox{metabg}{Loss} column lists the losses for learning the \colorbox{metabg}{Velocity field}, where the CFM loss (middle and bottom row) is what used in practice. The bottom row lists the simplest FM algorithm instantiation as described in \cref{s:quick_tour}.   
  }
\label{fig:diagram}
\end{figure}

\subsection{Deriving generating velocity fields}\label{sec:Deriving generating velocity fields}

Equipped with a marginal probability path $p_t$, we now build a velocity field $u_t$ generating $p_t$.
The generating velocity field $u_t$ is an average of multiple \highlight{conditional velocity fields} $u_t(x|x_1)$, illustrated in \cref{fig:u_t}, and satisfying: 
\begin{equation}\label{e:u_t|1_generates_p_t|1}
    u_{t}(\cdot|x_1) \text{ generates } p_{t|1}(\cdot|x_1).
\end{equation}
Then, the \highlight{marginal velocity field} $u_t(x)$, generating the marginal path $p_t(x)$,  illustrated in \cref{fig:marginal_u_t}, is given by averaging the conditional velocity fields $u_t(x|x_1)$ across target examples: 
\begin{equation}\label{e:u_t}
    u_t(x) = \int u_t(x|x_1)p_{1|t}(x_1|x) \dd x_1. %
\end{equation}
To express the equation above using known terms,  recall Bayes' rule 
\begin{equation} 
p_{1|t}(x_1|x) = \frac{p_{t|1}(x|x_1)q(x_1)}{p_t(x)},
\end{equation}
defined for all $x$ with $p_t(x)>0$. \Cref{e:u_t} can be interpreted as the weighted average of the conditional velocities $u_t(x|x_1)$, with weights $p_{1|t}(x_1|x)$ representing the posterior probability of target samples $x_1$ given the current sample $x$.
Another interpretation of \eqref{e:u_t} can be given with conditional expectations (see \cref{sec:conditional_densities_and_expectations}). Namely, if $X_t$ is \emph{any} RV such that $X_t\sim p_{t|1}(\cdot|X_1)$, or equivalently, the joint distribution of $(X_t,X_1)$ has density $p_{t,1}(x,x_1)=p_{t|1}(x|x_1)q(x_1)$ then using \eqref{e:f_x_y_cond_y} to write \eqref{e:u_t} as a conditional expectation, we obtain
\begin{equation}\label{e:u_t_cond_E}
  u_t(x) = \E\brac{ u_t(X_t|X_1)  \, \vert \, X_t=x },
\end{equation}
which yields the useful interpretation of $u_t(x)$ as the least-squares approximation to $u_t(X_t|X_1)$ given $X_t=x$, see \cref{sec:conditional_densities_and_expectations}. Note, that the $X_t$ in \eqref{e:u_t_cond_E} is in general a different RV that $X_t$ defined by the final flow model \eqref{e:flow_model}, although they share the same marginal probability $p_t(x)$. 

\subsection{General conditioning and the Marginalization Trick}\label{s:general_conditioning_and_main_theorem}

To justify the constructions above, we need to show that the marginal velocity field $u_t$ from \cref{e:u_t,e:u_t_cond_E} generates the marginal probability path $p_t$ from \cref{e:p_t} under mild assumptions.
The mathematical tool to prove this is the Mass Conservation Theorem (\cref{thm:continuity}).
To proceed, let us consider a slightly more general setting that will be useful later in the manuscript.
In particular, there is nothing special about building conditional probability paths and velocity fields by conditioning on $X_1=x_1$.
As noted in~\cite{tong2023improving}, the analysis from the previous section carries through to conditioning on any arbitrary RV $Z\in\Real^m$ with PDF $p_Z$.
This yields the \highlight{marginal probability path}
\begin{equation}\label{e:p_t_general}
    p_t(x) = \int p_{t|Z}(x|z) p_Z(z) dz,  %
\end{equation}
which in turn is generated by the \highlight{marginal velocity field} 
\begin{align}\label{e:u_t_general}
   u_t(x) = \int u_t(x|z) p_{Z|t}(z|x) dz = \E\brac{u_t(X_t|Z) \ \vert \ X_t=x},
\end{align}
where $u_t(\cdot|z)$ generates $p_{t|Z}(\cdot|z)$, $p_{Z|t}(z|x) = \frac{p_{t|Z}(x|z)p_Z(z)}{p_t(x)}$ follows from Bayes' rule given $p_t(x)>0$, and $X_t\sim p_{t|Z}(
\cdot|Z)$.
Naturally, we can recover the constructions in previous sections by setting $Z=X_1$.
Before we prove the main result, we need some regularity assumptions, encapsulated as follows.
\begin{myframe}
\begin{assumption}\label{as:p_t}
$p_{t|Z}(x|z)$ is $C^1([0,1)\times \Real^d)$ and $u_t(x|z)$ is $C^1([0,1)\times \Real^d,\Real^d)$ as a function of $(t,x)$.
Furthermore, $p_Z$ has bounded support, that is, $p_Z(x)=0$ outside some bounded set in $\Real^m$.
Finally, $p_t(x)>0$ for all $x\in\Real^d$ and $t\in[0,1)$.
\end{assumption}    
\end{myframe}
These are mild assumptions.
For example, one can show that $p_t(x) > 0$ by finding a condition $z$ such that $p_Z(z)>0$ and $p_{t|Z}(\cdot|z)>0$.
In practice, one can satisfy this by considering $(1-(1-t)\eps)p_{t|Z} + (1-t)\eps \gN(0,I)$ for an arbitrarily small $\eps > 0$.
One example of $p_{t|Z}(\cdot|z)$ satisfying this assumption is the path in \eqref{e:condot_path}, where we let $Z=X_1$.
We are now ready to state the main result:
\begin{myframe}[nobreak=true]
    \begin{theorem}[Marginalization Trick]\label{thm:fm_main} Under \cref{as:p_t}, if $u_t(x|z)$ is conditionally integrable and generates the conditional probability path $p_t(\cdot|z)$, then the marginal velocity field $u_t$ generates the marginal probability path $p_t$, for all $t\in[0,1)$.
\end{theorem}
\end{myframe}
In the theorem above, \emph{conditionally integrable} refers to a conditional version of the integrability condition from the Mass Conservation Theorem \eqref{e:integrable}, namely:
\begin{align}\label{e:conditional_integrable}
    \int_0^1\int\int \norm{u_t(x|z)} p_{t|Z}(x|z) p_Z(x) dz \dd x \dd t &<  \infty.
\end{align}
\begin{proof} 
The result follows from verifying the two conditions of the Mass Conservation in \cref{thm:continuity}.
First, let us check that the pair $(u_t, p_t)$ satisfies the Continuity Equation \eqref{e:continuity}.
Because $u_t(\cdot|x_1)$ generates $p_t(\cdot|x_1)$, we have that
\begin{align}\label{ea:marginal_trick_proof}
\frac{\dd}{\dd t}p_t(x) &\,\overset{({i})}{=}    \int \frac{\dd}{\dd t}p_{t|Z}(x|z)p_Z(x)dz \\
                   &\overset{({ii})}{=}  -\int \divv_x\brac{u_t(x|z) p_{t|Z}(x|z)} p_Z(z) dz\\
                   &\,\overset{({i})}{=}   -\divv_x \int u_t(x|z) p_{t|Z}(x|z)p_Z(z) dz\\ 
                   &\overset{({iii})}{=} -\divv_x \brac{u_t(x)p_t(x)}.
\end{align}
Equalities (i) follows from switching differentiation ($\frac{\dd}{\dd t}$ and $\divv_x$, respectively) and integration, as justified by Leibniz's rule, the fact that $p_{t|Z}(x|z)$ and $u_t(x|z)$ are $C^1$ in $t,x$, and the fact that $p_Z$ has bounded support (so all the integrands are integrable as continuous functions over bounded sets).
Equality (ii) follows from the fact that $u_t(\cdot|z)$ generates $p_{t|Z}(\cdot|z)$ and \cref{thm:continuity}.
Equality (iii) follows from multiplying and dividing by $p_t(x)$ (strictly positive by assumption) and using the formula \eqref{e:u_t_general} for $u_t$.

To verify the second and last condition from \cref{thm:continuity}, we shall prove that $u_t$ is integrable and locally Lipschitz.
Because $C^1$ functions are locally Lipschitz, it suffices to check that $u_t(x)$ is $C^1$ for all $(t, x)$.
This would follow from verifying that $u_t(x|z)$ and $p_{t|Z}(x|z)$ are $C^1$ and $p_t(x)>0$, which hold by assumption.
Furthermore, $u_t(x)$ is integrable because $u_t(x|z)$ is conditionally integrable:
\begin{align}
    \int_0^1 \int \norm{u_t(x)}p_t(x) \dd x \dd t &\leq \int_0^1\int\int \norm{u_t(x|z)}p_{t|Z}(x|z)p_Z(z) dz \dd x \dd t  < \infty,
\end{align}
where the first inequality follows from vector Jensen's inequality.
\end{proof}

\subsection{Flow Matching loss}\label{s:flow_matching_loss}

After having established that the target velocity field $u_t$ generates the prescribed probability path $p_t$ from $p$ to $q$, the missing ingredient is a tractable loss function to learn a velocity field model $u_t^\theta$ as close as possible to the target $u_t$.
One major roadblock towards stating this loss function directly is that computing the target $u_t$ is infeasible, as it requires marginalizing over the entire training set (that is, integrating with respect to $x_1$ in \cref{e:u_t} or with respect to $z$ in \cref{e:u_t_general}).
Fortunately, a family of loss functions known as \highlight{Bregman divergences} provides unbiased gradients to learn $u_t^\theta(x)$ in terms of \emph{conditional} velocities $u_t(x|z)$ alone.

\begin{wrapfigure}[10]{r}{0.27\textwidth}
  \begin{center}    \includegraphics[width=0.25\textwidth]{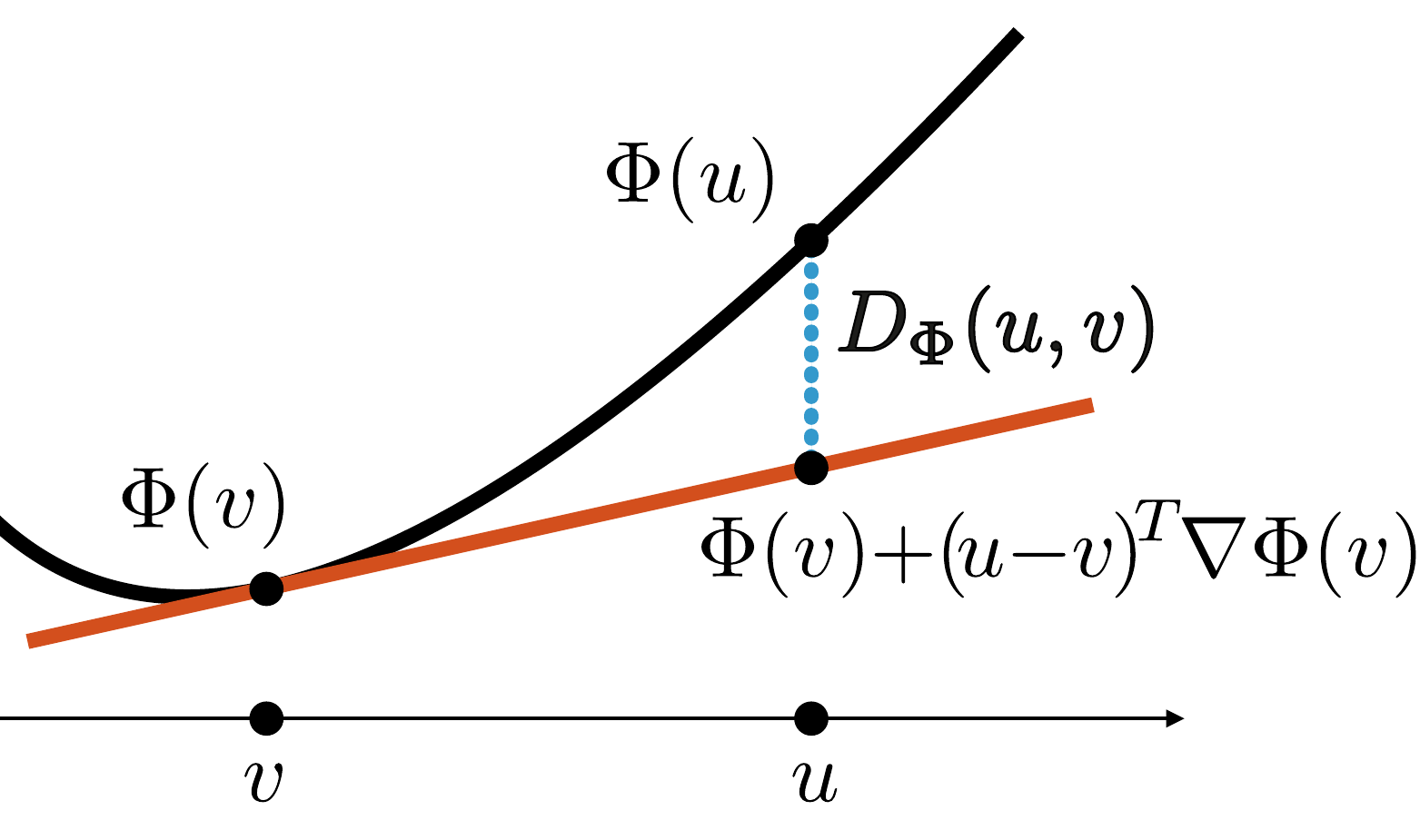}
  \end{center}
  \caption{Bregman divergence.}\label{fig:bregman}
\end{wrapfigure}

Bregman divergences measure dissimilarity between two vectors $u,v\in\Real^d$ as 
\begin{equation}\label{e:bd}
    D(u,v) \defe \Phi(u) - \brac{ \Phi(v) + \ip{u-v, \nabla \Phi(v)} },
\end{equation}
where $\Phi:\Real^d\too\Real$ is a strictly convex function defined over some convex set $\Omega\subset\Real^d$.
As illustrated in \cref{fig:bregman}, the Bregman divergence measures the difference between $\Phi(u)$ and the linear approximation to $\Phi$ developed around $v$ and evaluated at $u$.
Because linear approximations are global lower bounds for convex functions, it holds that $D(u,v)\geq 0$. Further, as $\Phi$ is strictly convex, it follows that $D(u,v)=0$ if and only if $u=v$.
The most basic Bregman divergence is the squared Euclidean distance $D(u,v) = \norm{u-v}^2$, esulting from choosing $\Phi(u)=\norm{u}^2$.
The key property that makes Bregman divergences useful for Flow Matching is that their gradient with respect to the second argument is \emph{affine invariant} \citep{holderrieth2024gm}:
\begin{equation}\label{e:bd_affine_grad}
    \nabla_v D(au_1 + bu_2, v) = a\nabla_v D(u_1,v) + b\nabla_v D(u_2,v), \text{ for any } a+b=1,
\end{equation}
as it can be verified from \cref{e:bd}. Affine invariance allows us to swap expected values with gradients as follows: 
\begin{equation}\label{e:swap_grad_and_exp}
    \nabla_v D(\mathbb{E}[Y], v) = \mathbb{E}[\nabla_v D(Y, v)]\, \text{ for any RV } Y\in\mathbb{R}^d.
\end{equation}

The \highlight{Flow Matching loss} employs a Bregman divergence to regress our learnable velocity $u_t^\theta(x)$ onto the target velocity $u_t(x)$ along the probability path $p_t$:
\begin{equation}\label{e:fm_loss}
  \gL_{\FM}(\theta) = \E_{t, X_t\sim p_t} D(u_t(X_t), u_t^\theta(X_t)), %
\end{equation}
where time $t \sim U[0,1]$.
As mentioned above, however, the target velocity $u_t$ is not tractable, so the loss above cannot be computed as is.
Instead, we consider the simpler and tractable \highlight{Conditional Flow Matching (CFM) loss}:
\begin{equation}\label{e:cfm_loss_main}
    \gL_{\CFM}(\theta) = \E_{t, Z, X_t\sim p_{t|Z}(\cdot|Z)} D(u_t(X_t|Z), u_t^\theta(X_t)). %
\end{equation}
The two losses are equivalent for learning purposes, since their gradients coincide \citep{holderrieth2024gm}:
\begin{myframe}
\begin{theorem}\label{thm:cfm}
    The gradients of the Flow Matching loss and the Conditional Flow Matching loss coincide:
    \begin{equation}
        \nabla_\theta \gL_{\FM}(\theta) = \nabla_\theta \gL_{\CFM}(\theta).
    \end{equation}
    In particular, the minimizer of the Conditional Flow Matching loss is the marginal velocity $u_t(x)$.
\end{theorem}
\end{myframe}
\begin{proof}
The proof follows a direct computation:
    \begin{align*}
    \nabla_\theta \gL_{\FM}(\theta) &= \nabla_\theta\E_{t, X_t\sim p_t} D(u_t(X_t), u_t^\theta(X_t))\\
&= \E_{t, X_t\sim p_t}\nabla_\theta D(u_t(X_t), u_t^\theta(X_t))\\
&\overset{(i)}{=} \E_{t, X_t\sim p_t}\nabla_v D(u_t(X_t), u_t^\theta(X_t))\nabla_\theta u_t^\theta(X_t)\\
&\overset{ \eqref{e:u_t_general}}{=} \E_{t, X_t\sim p_t}\nabla_v D(\textcolor{metablue}{\E_{Z\sim p_{Z|t}(\cdot|X_t)}\brac{\textcolor{black}{u_t(X_t|Z)}}}, u_t^\theta(X_t))\nabla_\theta u_t^\theta(X_t)\\
&\overset{({ii})}{=} \E_{t, X_t\sim p_t}\textcolor{metablue}{\E_{Z\sim p_{Z|t}(\cdot|X_t)}\brac{\textcolor{black}{\nabla_v D(u_t(X_t|Z), u_t^\theta(X_t))\nabla_\theta u_t^\theta(X_t)}}}\\
&\overset{(iii)}{=} \E_{t, X_t\sim p_t}\E_{Z\sim p_{Z|t}(\cdot|X_t)}[\nabla_\theta D(u_t(X_t|Z), u_t^\theta(X_t))]\\
&\overset{(iv)}{=} \nabla_\theta \E_{t, Z\sim q, X_t\sim p_{t|Z}(\cdot|Z)}[D(u_t(X_t|Z), u_t^\theta(X_t))]\\
&= \nabla_\theta \mathcal{L}_{\text{CFM}}(\theta)
\end{align*}
where in (i),(iii) we used the chain rule; (ii) follows from \cref{e:swap_grad_and_exp} applied conditionally on $X_t$; and in (iv) we use Bayes' rule.
\end{proof}

\paragraph{Bregman divergences for learning conditional expectations.} \Cref{thm:cfm} is a particular instance of a more general result utilizing Bregman divergences for learning conditional expectations described next. It will be used throughout this manuscript and provide the basis for all scalable losses behind Flow Matching:

\begin{myframe}
    \begin{proposition}[Bregman divergence for learning conditional expectations]\label{prop:bregman_gradient}
        Let $X\in\gS_X,Y\in\gS_Y$ be RVs over state spaces $\gS_X,\gS_Y$ and $g:\Real^p\times \gS_X\too \Real^n$, $(\theta,x)\mapsto g^\theta(x)$, where $\theta\in \Real^p$ denotes learnable parameters. Let $D_x(u,v)$, $x\in \gS_X$ be a Bregman divergence over a convex set $\Omega\subset \Real^n$ that contains the image of $f$. Then,
        \begin{equation}
            \textcolor{black}{\nabla_\theta} \E_{X,Y} D_X \parr{ \textcolor{metablue}{Y}, g\textcolor{black}{^\theta}(X) } = \textcolor{black}{\nabla_\theta} \E_X D_X \parr{ \textcolor{metablue}{\E\brac{ Y \, \vert \, X}}, g\textcolor{black}{^\theta}(X) }.
        \end{equation}
        In particular, for all $x$ with  $p_X(x)>0$, the global minimum of $g^\theta(x)$ w.r.t.~$\theta$ satisfies 
        \begin{equation}
            g\textcolor{black}{^\theta}(x) = \textcolor{black}{\E \brac{ Y \ \vert \ X=\textcolor{black}{x} }}.
        \end{equation}
    \end{proposition}
\end{myframe}
\pagebreak
\begin{proof}
  We assume $g^\theta$ is differentiable w.r.t.~$\theta$ and that the distributions of $X$ and $Y$, as well as $D_x$, and $g$ allow switching differentiation and integration, develop:
    \begin{align*}
        \nabla_\theta \E_{X,Y} D_X \parr{ Y, g^\theta(X) } &\overset{({i})}{=}  \E_X \brac{ \textcolor{metablue}{ \E \brac { \textcolor{black}{ \nabla_v D_X \parr{ Y, g^\theta(X) } \nabla_\theta g^\theta(X)} \ \vert 
 \ X } } }\\
        &\overset{({ii})}{=} \E_{X} \brac{ \nabla_v D_X \parr{ \textcolor{metablue}{\E\brac{ \textcolor{black}{Y} \ \vert \ X}}, g^\theta(X) } \nabla_\theta g^\theta(X) } \\
        &\overset{({iii})}{=} \E_{X} \brac{ \nabla_\theta D_X \parr{ \textcolor{metablue}{\E\brac{ \textcolor{black}{Y} \ \vert \ X}}        
        , g^\theta(X) } } \\
        &\,\,= \nabla_\theta  \E_{X}  D_X \parr{ \textcolor{metablue}{\E\brac{ \textcolor{black}{Y} \ \vert \ X}}, g^\theta(X) },
    \end{align*}
    where (i) follows from the chain rule and the tower property of expectations \eqref{e:tower}.
  Equality (ii) follows from \eqref{e:swap_grad_and_exp}. 
  Equality (iii) uses the chain rule again.
  Lastly, for every $x\in\gS_X$ with $p_X(x)>0$ we can choose $g^\theta(x)=\E\brac{Y|X=x}$, obtaining $\E_X D_X \parr{ \textcolor{black}{\E\brac{ Y \ \vert \ X}}, g\textcolor{black}{^\theta}(X) } = 0$, which must be the global minimum with respect to $\theta$.
\end{proof}
\Cref{thm:cfm} is readily shown from  \cref{prop:bregman_gradient} by making the choices $X=X_t$, $Y=u_t(X_t|Z)$, $g^\theta(x)=u_t^\theta(x)$, and taking the expectation with respect to $t \sim U[0,1]$.

\paragraph{General time distributions} One useful variation of the FM loss is to sample times $t$ from a distribution other than Uniform.
Specifically, consider $t\sim \omega(t)$, where $\omega$ is a PDF over $[0,1]$.
This leads to the following weighted objective:
\begin{equation}
    \gL_{\CFM} (\theta) = \E_{t\sim \textcolor{metablue}{\omega}, Z, X_t} D(u_t(X_t|Z),u_t^\theta(X_t)) = \E_{t\sim U, Z, X_t} \textcolor{metablue}{\omega(t)} D(u_t(X_t|Z),u_t^\theta(X_t)).
\end{equation}
Although mathematically equivalent, sampling $t\sim \omega$ leads to better performance than using weights $\omega(t)$ in large scale image generation tasks \citep{esser2024scaling}.

\subsection{Solving conditional generation with conditional flows}\label{s:conditional_flows}

So far, we have reduced the problem of training a flow model $u_t^\theta$ to: (i) Find conditional probability paths $p_{t|Z}(x|z)$ yielding a marginal probability path $p_t(x)$ satisfying the boundary conditions in \eqref{e:p_q_interp}. (ii) Find conditional velocity fields $u_t(x|z)$ generating the conditional probability path. (iii) Train using the Conditional Flow Matching loss (see \cref{e:cfm_loss_main}). We now discuss a concrete options on how to do step (i) and (ii), \ie, design such conditional probability paths and velocity fields.

We will now propose a flexible method to design such conditional probability paths and velocity fields using a specific construction via \highlight{conditional flows}. The idea is as follows: %
\emph{Define} a flow model $X_{t|1}$ (similarly to \eqref{e:flow_model}) satisfying the boundary conditions \eqref{e:p_t_cond_boundary}, and extract the velocity field from $X_{t|1}$ by differentiation  \eqref{e:u_from_psi}.
This process defines both $p_{t|1}(x|x_1)$ and $u_t(x|x_1)$.
In more detail, define the \highlight{conditional flow model}
\begin{equation}
    X_{t|1} = \psi_t(X_0|x_1), \quad \text{ where } X_0\sim \pi_{0|1}(\cdot\,|\, x_1), %
\end{equation}    
where $\psi:[0,1)\times \Real^d \times \Real^d \too \Real^d$ is a \highlight{conditional flow} defined by
    \begin{equation}\label{e:psi}
    \psi_t(x|x_1) = \begin{cases}
        x & t=0 \\
        x_1 & t=1
    \end{cases}, %
\end{equation}
smooth in $(t, x)$, and a diffeomorphism in $x$.
(Smooth here means that all derivatives of $\psi_t(x|x_1)$ with respect to $t$ and $x$ exist and are continuous: $C^\infty([0,1)\times \Real^d,\Real^d)$). These conditions could be further relaxed to $C^2([0,1)\times \Real^d,\Real^d)$ at the expense of simplicity.)
The push-forward formula \eqref{e:push-forward_p} defines the probability density of $X_{t|1}$ as 
\begin{equation}\label{e:p_t|1}
    p_{t|1}(x|x_1) \defe \brac{\psi_t(\cdot | x_1)_\sharp \pi_{0|1}(\cdot|x_1)}(x),
\end{equation}
although we will not need this expression in practical optimization of the CFM loss it is used theoretically to show that $p_{t|1}$ satisfies the two boundary conditions \eqref{e:p_t_cond_boundary}.
First, and according to \eqref{e:psi}, $\psi_0(\cdot|x_1)$ is the identity map, keeping $\pi_{0|1}(\cdot|x_1)$ intact at time $t=0$.
Second, $\psi_1(\cdot|x_1)=x_1$ is the constant map, concentrating all probability mass at $x_1$ as $t\too 1$.
Furthermore, note that $\psi_t(\cdot|x_1)$ is a smooth diffeomorphism for $t\in [0,1)$.
Therefore, by the equivalence of flows and velocity fields (\cref{sec:equivalence_flows_velocities}), there exists a unique smooth conditional velocity field (see \cref{e:u_from_psi}) taking form:
\begin{equation}\label{e:psi_and_u}
    u_t(x|x_1) = \dot{\psi}_t(\psi_t^{-1}(x|x_1)|x_1).
\end{equation}
To summarize: we have further reduced the task of finding the conditional path and a corresponding  generating velocity to simply building a conditional flow $\psi_{t}(\cdot|x_1)$ satisfying \eqref{e:psi}.  %
In \cref{s:kinetic_optimal} we will pick a particularly simple  $\psi_t(x|x_1)$ with some desirable properties (conditional Optimal Transport flow) that leads to the standard Flow Matching algorithm as seen in \cref{section:intro}, and in \cref{sec:affine_conditional_flows} we will discuss a particular and well-known family of conditional flows, namely affine flows that include some known examples from the diffusion models' literature. In \cref{s:fm_non_euclidean} we will use conditional flows to define Flow Matching on manifold which showcase the flexibility of this approach.

\subsubsection{The Conditional Flow Matching loss, revisited}

Let us revisit the CFM loss \eqref{e:cfm_loss_main} by setting $Z=X_1$ and using the conditional flows way of defining the conditional probability path and velocity, 
\begin{align}\nonumber
    \gL_{\CFM}(\theta) &= \E_{t, X_1, X_t\sim p_t(\cdot|X_1)}  D\parr{u_t(X_t|X_1), u_t^\theta(X_t)} \\  \label{e:cfm_psi}  &\overset{\eqref{e:law_of_uncon_stat}}{=} \E_{t, (X_0,X_1)\sim \pi_{0,1}} D\parr{ \dot{\psi}_t(X_0|X_1), u_t^\theta(X_t)}
\end{align}
where in the second equality we used the Law of Unconscious Statistician with $X_t=\psi_t(X_0|X_1)$ and 
\begin{equation}\label{e:u_t_X_t_cond_X_1}
 u_t(X_t|X_1) \overset{\eqref{e:psi_and_u}}{=} \dot{\psi}_t\parr{
 \psi_t^{-1}\parr{ \psi_t\parr{X_0 | X_1} \big| X_1} \Big | X_1
 } = \dot{\psi}_t(X_0|X_1).  
\end{equation}
The minimizer of the loss \eqref{e:cfm_psi} according to \cref{prop:bregman_gradient} takes the form as in \citep{liu2022flow},
\begin{equation}\label{e:u_t_dpsi}
    u_t(x) = \E\brac{\dot{\psi}_t(X_0|X_1) \, \Big \vert \, X_t=x}.
\end{equation}

In the \fmlibrary{} the \texttt{ProbPath} object defines a probability path. This probability path can be sampled at ($t$, $X_0$, $X_1$) to obtain $X_t$ and $\dot{\psi}_t(X_0|X_1)$. Then, one can  compute a Monte Carlo estimate of the CFM loss $\gL_{\CFM}(\theta)$. An example training loop with the CFM objective is shown in \cref{ex:cfm_loss}.

\begin{pbox}[label={ex:cfm_loss}]{Training with the conditional flow matching (CFM) loss}
\begin{minted}[linenos, breaklines, mathescape, fontsize=\footnotesize, xleftmargin=2em]{python}
import torch
from flow_matching.path import ProbPath
from flow_matching.path.path_sample import PathSample

path: ProbPath = ...  # The flow_matching library implements the most common probability paths
velocity_model: torch.nn.Module = ...  # Initialize the velocity model
optimizer = torch.optim.Adam(velocity_model.parameters())

for x_0, x_1 in dataloader:  # Samples from $\pi_{0,1}$ of shape [batch_size, *data_dim]
    t = torch.rand(batch_size)  # Randomize time $t \sim U[0,1]$
    sample: PathSample = path.sample(t=t, x_0=x_0, x_1=x_1) 
    x_t = sample.x_t
    dx_t = sample.dx_t  # dX_t is $\dot{\psi}_t(X_0|X_1)$.
    # If $D$ is the Euclidean distance, the CFM objective corresponds to the mean-squared error
    cfm_loss = torch.pow(velocity_model(x_t, t) - dx_t, 2).mean()  # Monte Carlo estimate
    optimizer.zero_grad()
    cfm_loss.backward()
    optimizer.step()
\end{minted}
\end{pbox}

\pagebreak
\subsubsection{The Marginalization Trick for probability paths built from conditional flows}

Next, we introduce a version of the Marginalization trick for probability paths that are built from conditional flows.
To this end, note that if $\pi_{0|1}(\cdot|x_1)$ is $C^1$, then $p_t(x|x_1)$ is also $C^1$ by construction; moreover, $u_t(x|x_1)$ is conditionally integrable if 
\begin{equation}\label{e:psi_integrable}
    \E_{t, (X_0,X_1)\sim \pi_{0,1}} \norm{\dot{\psi}_t(X_0|X_1)}  < \infty.
\end{equation}
Therefore, by setting $Z=X_1$, the following corollary to \cref{thm:fm_main} is obtained.
\begin{myframe}
    \begin{corollary}\label{cor:cond_flows} Assume that $q$ has bounded support, $\pi_{0|1}(\cdot|x_1)$ is $C^1(\Real^d)$ and strictly positive for some $x_1$ with $q(x_1)>0$, and $\psi_t(x|x_1)$ is a conditional flow satisfying equations \eqref{e:psi} and \eqref{e:psi_integrable}. Then $p_{t|1}(x|x_1)$ and $u_{t}(x|x_1)$, defined in \eqref{e:p_t|1} and \eqref{e:psi_and_u}, respectively, define a marginal velocity field $u_t(x)$ generating the marginal probability path $p_t(x)$ interpolating $p$ and $q$.
\end{corollary}
\end{myframe}
\begin{proof}
  If $\pi_{0|1}(\cdot|x_1)>0$ for some $x_1\in\Real^d$ such that $q(x_1)>0$, it follows that $p_{t|1}(x|x_1)>0$ for all $x\in \Real^d$ and is $C^1([0,1)\times \Real^d)$ (see \eqref{e:p_t|1} and \eqref{e:push-forward_p} for definitions).
  Furthermore, $u_t(x|x_1)$ (defined in \eqref{e:psi_and_u}) is smooth and satisfies
    \begin{align*}
        \int_0^1\int \norm{u_t(x|x_1)} p_{t|1}(x|x_1)q(x_1)\dd x_1 \dd x \dd t 
        &\,\,= \E_{t, X_1\sim q, X_t\sim p_{t|1}(\cdot|X_1)}\norm{u_t(X_{t}|X_1)} \\
        & \overset{\eqref{e:law_of_uncon_stat}}{=} \E_{t, X_1\sim q, X_0\sim \pi_{0|1}(\cdot|X_1)} \norm{u_t(\psi_t(X_0|X_1)|X_1)} \\
        &\overset{\eqref{e:u_t_X_t_cond_X_1}}{=}
        \E_{t, (X_0,X_1)\sim \pi_{0,1}} \norm{\dot{\psi}_t(X_0|X_1)}\\
        &\,\,\,\,< \infty.
    \end{align*}
    Therefore, $u_t(x|x_1)$ is conditionally integrable (see \eqref{e:conditional_integrable}).
    By \cref{thm:fm_main}, the marginal $u_t$ generates $p_t$.
    Because $p_{t|1}(x|x_1)$ as defined by \eqref{e:p_t|1} satisfies \eqref{e:p_t_cond_boundary}, it follows that $p_t$ interpolates $p$ and $q$.
\end{proof}
This theorem will be used as a tool to show that particular choices of conditional flows lead to marginal velocity $u_t(x)$ generating the marginal probability path $p_t(x)$.

\begin{figure}
\begin{equation*}\renewcommand{\arraystretch}{1.5}
    \begin{matrix}
        \\ p_t(x) = \\  u_t(x) =
    \end{matrix} \  \overset{X_1\text{ conditioning}}{\overset{\includegraphics[width=0.25\textwidth]{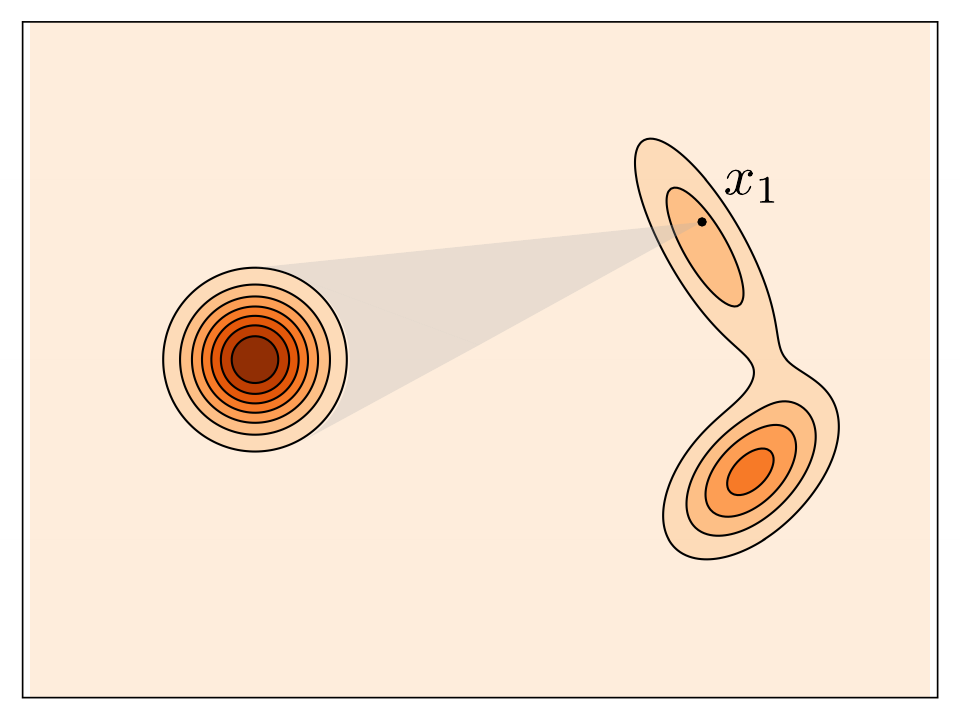}}{\begin{matrix}
    \psi_t(X_0|x_1) \sim p_{t|1}(\cdot|x_1)  \\
    \int \textcolor{metablue}{p_{t|1}(x|x_1)}q(x_1)\dd x_1 \\
    \E \brac{ \textcolor{metablue}{u_{t}(X_t|X_1)} \vert X_t=x }
    \end{matrix}
    }} \ \begin{matrix}
      \\   = \\  =
    \end{matrix} \ \overset{X_0\text{ conditioning}}{\overset{\includegraphics[width=0.25\textwidth]{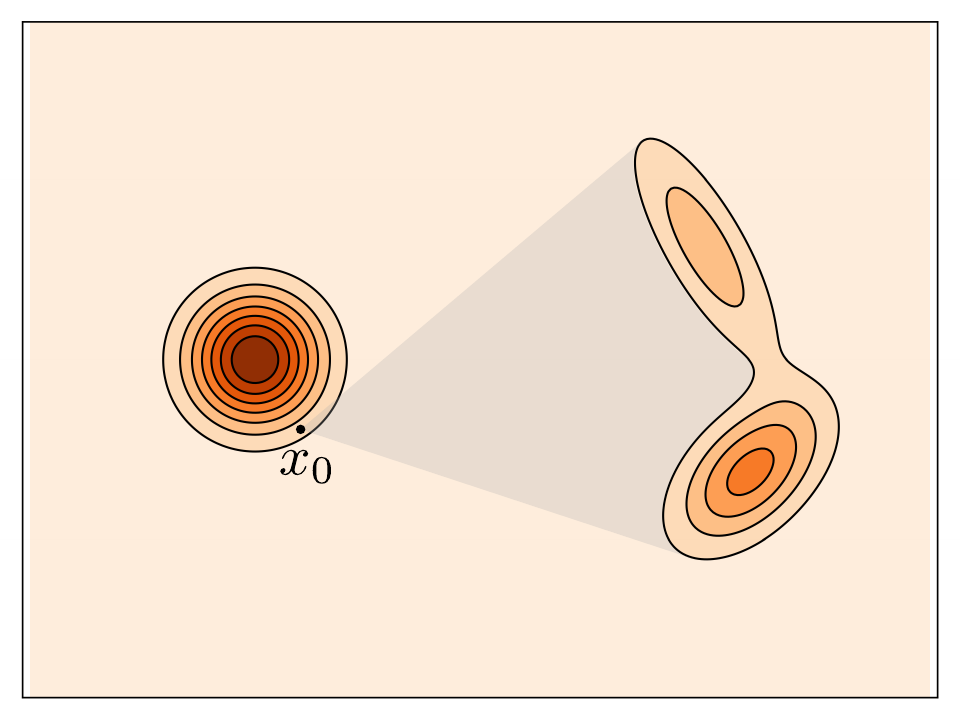}}{
    \begin{matrix}
    \psi_t(X_1|x_0) \sim p_{t|0}(\cdot|x_0)  \\
    \int \textcolor{metablue}{p_{t|0}(x|x_0)}p(x_0)\dd x_1 \\
        \E \brac{ \textcolor{metablue}{u_{t}(X_t|X_0)} \vert X_t=x }
    \end{matrix}
    }} \ \begin{matrix}
         \\ = \\  =
    \end{matrix} \ \overset{(X_0,X_1)\text{ conditioning}}{\overset{\includegraphics[width=0.25\textwidth]{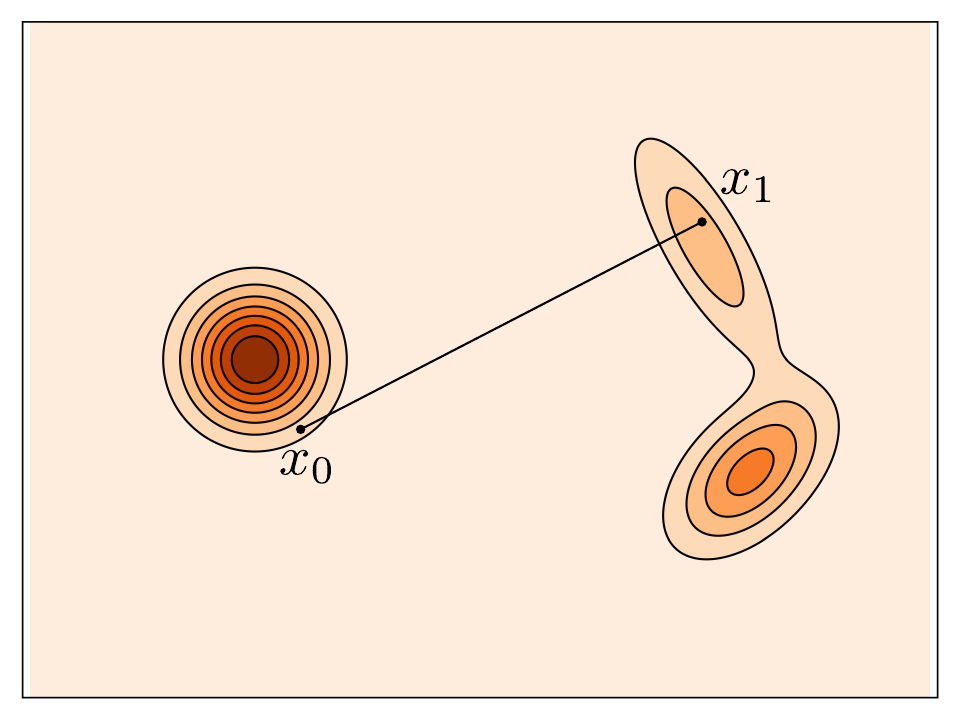}}{\begin{matrix} 
    \psi_t(x_0,x_1) \sim p_{t|0,1}(\cdot|x_0,x_1) \\
    \int \textcolor{metablue}{p_{t|0,1}(x|x_0,x_1)}\pi_{0,1}(x_0,x_1)\dd x_0 \dd x_1 \\
    \E \brac{ \textcolor{metablue}{u_{t}(X_t|X_0,X_1)} \vert X_t=x }\end{matrix}}}
\end{equation*}
  \caption{Different forms of conditioning in Flow Matching  and path design with corresponding conditional flows. When the conditional flows are a diffeomorphism, all constructions are equivalent. When, they are not, extra conditions are required to validate that the marginal velocity generates the marginal path, see text for more details. }\label{fig:conditioning}
\end{figure}

\pagebreak
\subsubsection{Conditional flows with other conditions}
Different conditioning choices $Z$ exist but are essentially all equivalent. %
As illustrated in~\cref{fig:conditioning}, main options include fixing target samples $Z=X_1$ \citep{lipman2022flow}, source samples $Z=X_0$ \citep{esser2024scaling}, or two-sided $Z=(X_0, X_1)$~\citep{albergo2022building,liu2022flow,pooladian2023multisample,tong2023improving}.

Let us focus on the two-sided condition $Z=(X_0, X_1)$. Following the FM blueprint described above, we are now looking to build a conditional probability path $p_{t|0,1}(x|x_0,x_1)$ and a corresponding generating velocity $u_t(x|x_0,x_1)$ such that 
\begin{equation}\label{e:si_boundary_constraints}
    p_{0|0,1}(x|x_0,x_1)=\delta_{x_0}(x), \text{ and } p_{1|0,1}(x|x_0,x_1)=\delta_{x_1}(x).
\end{equation}
We will keep this discussion formal as it requires usage of delta functions $\delta$ and our existing derivations so far only deals with probability densities (and not general distributions). To build such a path we can consider an \highlight{interpolant}  \citep{albergo2022building} defined by $X_{t|0,1}=\psi_t(x_0,x_1)$ for a function $\psi:[0,1]\times\Real^d\times\Real^d\too \Real^d$ satisfying conditions similar to \eqref{e:psi},
\begin{equation}\label{e:si_interpolation}
    \psi_t(x_0,x_1) = \begin{cases}
        x_0 & t=0 \\ x_1 & t=1.
    \end{cases}
\end{equation}
Therefore, $\psi_t(\cdot,x_1)$ pushes $\delta_{x_0}(x)$ to $\delta_{x_1}(x)$. We now, similarly to before, define the conditional probability path to be
\begin{equation}%
    p_{t|0,1}(\cdot|x_0,x_1) \defe \psi_t(\cdot,x_1)_\sharp \delta_{x_0}(\cdot)
\end{equation}
which satisfies the boundary constraints in \eqref{e:si_boundary_constraints}. \cite{albergo2022building}'s  \highlight{stochastic interpolant} is defined by 
\begin{equation}
    X_t=\psi_t(X_0,X_1)\sim p_t(\cdot)=\int p_{t|0,1}(\cdot|x_0,x_1)\pi_{0,1}(x_0,x_1)\dd x_0 \dd x_1.
\end{equation} %
Next, the conditional velocity along this path can also be computed with \eqref{e:u_from_psi} giving
\begin{equation}
    u_t(x|x_0,x_1) = \dot{\psi}_t(x_0,x_1)
\end{equation}
which is defined only for $x=\psi_t(x_0,x_1)$. 
Ignoring for a second the extra conditions, \Cref{thm:fm_main} now presumably implies that the marginal velocity generating $p_t(x)$ is
\begin{align*}
    u_t(x) &= \E\brac{u_t(X_t|X_0,X_1) \, \vert \, X_t=x} \\
    &= \E\brac{\dot{\psi}_t(X_0,X_1) \, \vert \, X_t=x},
\end{align*}
which leads to the same marginal formula as the $X_1$-conditioned case \eqref{e:u_t_dpsi}, but with a seemingly more permissive conditional flow $\psi_t(x_0,x_1)$ which is only required to be an interpolant now, weakening the more stringent diffeomorphism condition. However, A more careful look reveals that some extra conditions are still required to make $u_t(x)$ a generating velocity for $p_t(x)$ and simple interpolation (as defined in \eqref{e:si_interpolation}) is not enough to guarantee this, not even with extra smoothness  conditions, as required in \Cref{thm:fm_main}.
To see this, consider
\begin{equation*} 
\psi_t(x_0,x_1) = \parr{1-2t}_+^\tau x_0 + \parr{2t - 1}_+^\tau x_1, \text{ where } (s)_+=\text{ReLU}(s), \tau > 2, 
\end{equation*}
a $C^2([0,1])$ interpolant (in time) concentrating all probability mass at location $0$ at time $t=0.5$ for all $x_0,x_1$. That is $\sP(X_{\frac{1}{2}}=0) = 1$. Therefore, assuming $u_t(x)$ indeed generates $p_t(x)$ its marginal at $t=\frac{1}{2}$ is $\delta_0$ and since a flow is both Markovian (as shown in \eqref{e:flow_is_markov}) and deterministic its marginal has to be a delta function for all $t>0.5$ leading to a contradiction since $X_1=\psi_1(X_0,X_1)\sim q$, which is generally not a delta function. \citet{albergo2022building} and \citet{liu2022flow} provide some extra conditions that guarantee that $u_t(x)$ indeed geenrates $p_t(x)$ but these are somewhat harder to verify compared to the conditions of \Cref{thm:fm_main}. Below we will show how to practically check the conditions of \Cref{thm:fm_main} to validate that particular paths of interest are guaranteed to be generated by the respective marginal velocities. 

Nevertheless, when $\psi_t(x_0,x_1)$ is in addition a diffeomorphism in $x_0$ for a fixed $x_1$, and in $x_1$ for a fixed $x_0$, the three constructions leads to the same marginal velocity, defined by \eqref{e:u_t_dpsi}, and same marginal probability path $p_t$, defined by $X_t=\psi_t(X_0,X_1)=\psi_t(X_0|X_1)=\psi_t(X_1|X_0)$, see Figure \ref{fig:conditioning}.

\subsection{Optimal Transport and linear conditional flow}\label{s:kinetic_optimal}
We now ask: how to find a useful conditional flow $\psi_t(x|x_1)$? One approach is to choose it as a minimizer of a natural cost functional, ideally with some desirable properties.  
One popular example of such cost functional is the dynamic Optimal Transport problem with quadratic cost~\citep{villani2009optimal,villani2021topics,peyre2019computational}, formalized as
\begin{subequations}\label{e:optimal_transport}
    \begin{align}
    (p_t^\star, u_t^\star) = \argmin_{p_t,u_t} & \int_0^1 \int \norm{u_t(x)}^2 p_t(x) \dd x \dd t &\text{(Kinetic Energy)}\label{e:kinetic_energy}\\
    \text{s.t. }                               & p_0=p, p_1=q                               &\text{(interpolation)}\label{e:ot_boundary}\\
                                               & \frac{\dd}{\dd t} p_t + \divv(p_t u_t) =0.      &\text{(continuity equation)}\label{e:ot_continuity_equation}
    \end{align} 
\end{subequations} 

The $(p_t^\star,u_t^\star)$ above defines a flow (via \cref{e:flow}) with the form 
\begin{equation}
    \psi^\star_t(x) = t\phi(x) + (1-t)x, 
\end{equation}
called the \highlight{OT displacement interpolant}~\citep{mccann1997convexity}, where $\phi:\Real^d\too\Real^d$ is the Optimal Transport map.
The OT displacement interpolant also solves the Flow Matching Problem \eqref{prob:gen_flow} by defining the random variable
\begin{equation}
    X_t = \psi_t^\star(X_0)\sim p^\star_t \quad \text{ when } \quad  X_0 \sim p.
\end{equation}
The Optimal Transport formulation promotes straight sample trajectories 
\begin{equation*}
  X_t=\psi^\star_t(X_0)=X_0 + t(\phi(X_0)-X_0),
\end{equation*}
with a constant velocity $\phi(X_0)-X_0$, which are in general easier to sample with ODE solvers---in particular, the target sample $X_1$ is here perfectly solvable with a single step of the Euler Method \eqref{e:euler_method}.

We can now try to plug our marginal velocity formula (\cref{e:u_t_dpsi}) into the Optimal Transport problem \eqref{e:optimal_transport} and search for an optimal $\psi_t(x|x_1)$. While this seems like a challenge, we can instead find a bound for the Kinetic Energy for which such a minimizer is readily found  \citep{liu2022flow}: 
\begin{align}
    \int_0^1 \E_{X_t\sim p_t} \norm{u_t(X_t)}^2 \dd t &= \int_0^1 \E_{X_t\sim p_t} \norm{\E\brac{\dot{\psi}_t(X_0|X_1) \, \big\vert \ X_t}}^2 \dd t \\
    &\overset{(i)}{\leq} \int_0^1 \E_{X_t\sim p_t} \E\brac{\norm{\dot{\psi}_t(X_0|X_1)}^2 \, \Big\vert \ X_t} \dd t \\ \label{e:ke_bound}
    &\overset{(ii)}{=} \E_{(X_0,X_1)\sim \pi_{0,1}}\int_0^1 \norm{\dot{\psi}_t(X_0|X_1)}^2 \dd t,
\end{align}
where in the (i) we used Jensen's inequality, and in (ii) we used the tower property of conditional expectations (see \cref{e:tower}) and switch integration of $t$ and expectation. Now the integrand in \eqref{e:ke_bound} can be minimized individually for each $(X_0,X_1)$ --- this leads to the following variational problem for $\gamma_t=\psi_t(x|x_1)$:
\begin{subequations}
    \begin{align}
    \min_{\gamma:[0,1]\too\Real^d} &\quad \int_0^1 \norm{\dot{\gamma}_t}^2 \dd t\\
    \text{s.t.} &\quad \gamma_0 = x, \gamma_1 = x_1.
    \end{align}
\end{subequations}
This problem can be solved using Euler-Lagrange equations~\citep{gelfand2000calculus}, which in this case take the form $\frac{d^2}{\dd t^2}{\gamma}_t=0$.
By incorporating the boundary conditions, we obtain the minimizer:
\begin{equation}\label{e:linear}
    \psi_t(x|x_1) =  t x_1 + (1-t)x.
\end{equation}  
Note that although not constrained to be, this choice of $\psi_t(x|x_1)$ is a diffeomorphism in $x$ for $t\in [0,1)$ and smooth in $t,x$, as required from conditional flows.

\pagebreak
Several conclusions can be drawn:
\begin{enumerate}
    \item The linear conditional flow minimizes a bound of the Kinetic Energy among \emph{all} conditional flows. 
    \item In case the target $q$ consists of a \emph{single} data point  $q(x)=\delta_{x_1}(\cdot)$ we have that the linear conditional flow in \eqref{e:linear} is the Optimal Transport \citep{lipman2022flow}. Indeed, in this case $X_t=\psi_t(X_0|x_1)\sim p_t$ and $X_0=\psi^{-1}(X_t|x_1)$ is a function of $X_t$ which makes $\E\brac{\dot{\psi}_t(X_0|x_1) \, \big\vert \ X_t}=\dot{\psi}_t(X_0|x_1)$ and therefore (ii) becomes an equality.

    \begin{myframe}
\begin{theorem}\label{thm:cond_ot}
    If $q=\delta_{x_1}$, then the dynamic OT problem \eqref{e:optimal_transport} has an analytic solution given by the OT displacement interpolant in \eqref{e:linear}.
\end{theorem}    
\end{myframe} 
    \item Plugging the linear conditional flow in \eqref{e:ke_bound} we get 
    \begin{equation}\label{e:KE_leq_coupling}
        \int_0^1 \E_{X_t\sim p_t} \norm{u_t(X_t)}^2 \dd t \leq \E_{(X_0,X_1)\sim \pi_{0,1}} \int_0^1 \norm{X_1-X_0}^2 \dd t
    \end{equation}
    showing that the Kinetic Energy of the marginal velocity $u_t(x)$ is not bigger than that of the original coupling $\pi_{0,1}$ \citep{liu2022flow}.
\end{enumerate}
The conditional flow in \eqref{e:linear} is in particular affine and consequently motivates investigating the family of \emph{affine} conditional flows, discussed next.

\subsection{Affine conditional flows}\label{sec:affine_conditional_flows}

In the previous section we discovered the linear (Conditional-OT) flows as a minimizer to a bound of the Kinetic Energy among \emph{all} conditional flows. The linear conditional flow is a particular instance the wider family of \emph{affine conditional flows}, explored in this section.  
\begin{equation}\label{e:psi_affine}
    \psi_t(x|x_1) = \alpha_t x_1 + \sigma_t x,
\end{equation}
where $\alpha_t,\sigma_t : [0,1] \to [0,1]$ are smooth functions satisfying 
\begin{equation}\label{e:scheduler}
  \alpha_0=0=\sigma_1, \text{ } \alpha_1=1=\sigma_0, \text{ and } \dot{\alpha}_t,-\dot{\sigma}_t>0 \text{ for } t\in(0,1).
\end{equation} 
We call the pair $(\alpha_t,\sigma_t)$ a \emph{scheduler}.
The derivative condition above ensures that $\alpha_t$ is strictly monotonically increasing, while $\sigma_t$ is strictly monotonically decreasing.
The conditional flow \eqref{e:psi_affine} is a simple affine map in $x$ for each $t\in[0,1)$, which satisfies the conditions \eqref{e:psi}.
The associated marginal velocity field \eqref{e:u_t_dpsi} is
\begin{equation}\label{e:u_t_affine}
    u_t(x) = \E\brac{ \dot{\alpha}_t X_1 + \dot{\sigma}_t X_0 \vert X_t=x }.
\end{equation}
By virtue of \cref{cor:cond_flows}, we can prove that, if using the independent coupling and a smooth and strictly positive source density $p$ with finite second moments---for instance, a Gaussian $p = \gN(\cdot|0,I)$---then $u_t$ generates a probability path $p_t$ interpolating $p$ and $q$.
We formally state this result, significant for Flow Matching applications, as the following theorem. 
\begin{myframe}
    \begin{theorem}
      Assume that $q$ has bounded support, $p$ is $C^1(\Real^d)$ with strictly positive density with finite second moments, and these two relate by the independent coupling $\pi_{0,1}(x_0,x_1)=p(x_0)q(x_1)$.
      Let $p_t(x)=\int p_{t|1}(x|x_1)q(x_1)\dd x_1$ be defined by \cref{e:p_t|1}, with $\psi_t$ defined by \cref{e:psi_affine}.
      Then, the marginal velocity \eqref{e:u_t_affine} generates $p_t$ interpolating $p$ and $q$.
    \end{theorem}
\end{myframe}
\begin{proof}
    We apply \cref{cor:cond_flows}.
    First, note that $\pi_{0|1}(\cdot|x_1)=p(\cdot)$ is $C^1$ and positive everywhere by assumption.
    Second, $\psi_t$, defined in \eqref{e:psi_affine}, satisfies \eqref{e:psi}.
    Third, we are left with checking \eqref{e:psi_integrable}:
    \begin{align*}
        \E_{t,(X_0,X_1)} \norm{\dot{\psi}(X_0|X_1)} &= \E_{t,(X_0,X_1)} \norm{\dot{\alpha}_t X_1 + \dot{\sigma}_t X_0} \\
        &\leq \E_t |\dot{\alpha}_t|\, \E_{X_1} \norm{X_1} + \E_t |\dot{\sigma}_t|\, \E_{X_0} \norm{X_0} \\
        &= \E_{X_1} \norm{X_1} + \E_{X_0} \norm{X_0} \\
        & < \infty,
    \end{align*}
    where the last inequality follows from the fact that $X_1\sim q$ has bounded support and $X_0\sim p$ has bounded second moments.
\end{proof}
In this affine case, the CFM loss \eqref{e:cfm_psi} takes the form
\begin{equation}
    \gL_{\CFM}(\theta) = \E_{t, (X_0,X_1)\sim \pi_{0,1}} D( \dot{\alpha}_t X_1 + \dot{\sigma}_t X_0 , u_t^\theta(X_t)) \,.
\end{equation}

\begin{pbox}[label={ex:fm_affine}]{Examples of affine probability paths in the \fmlibrary{}}
\begin{minted}[linenos, breaklines, mathescape, fontsize=\footnotesize, xleftmargin=2em]{python}
from flow_matching.path import AffineProbPath, CondOTPath
from flow_matching.path.scheduler import (
    CondOTScheduler, PolynomialConvexScheduler, LinearVPScheduler, CosineScheduler)

# Conditional Optimal Transport schedule with $\alpha_t=t$, $\sigma_t=1-t$
path = AffineProbPath(scheduler=CondOTScheduler())
path = CondOTPath()  # Shorthand for the affine path with the CondOTScheduler above

# Polynomial schedule with $\alpha_t=t^n$, $\sigma_t=1-t^n$
path = AffineProbPath(scheduler=PolynomialConvexScheduler(n=1.0))

# Linear variance preserving schedule with $\alpha_t=t$, $\sigma_t=\sqrt{1-t^2}$
path = AffineProbPath(scheduler=LinearVPScheduler())

# Cosine schedule with $\alpha_t=\sin(0.5t\pi)$, $\sigma_t=\cos(0.5t\pi)$
path = AffineProbPath(scheduler=CosineScheduler())
\end{minted}
\end{pbox}

\subsubsection{Velocity parameterizations}\label{s:velocity_param}

In the affine case, the marginal velocity field $u_t$ admits multiple parametrizations, each of them learnable using the Flow Matching losses introduced in~\cref{s:flow_matching_loss}.
To derive these parametrizations, use the equivalent formulations of the affine paths 
\begin{equation}\label{e:X_t_equiv}
    X_t=\alpha_t X_1 + \sigma_t X_0 \quad \Leftrightarrow \quad X_1 = \frac{X_t - \sigma_t X_0}{\alpha_t} \quad \Leftrightarrow \quad X_0 = \frac{X_t - \alpha_t X_1}{\sigma_t},
\end{equation}
in the marginal velocity formula \eqref{e:u_t_affine}, obtaining 
\begin{align}\label{e:u_t_affine_x_1t_x_0t}
    u_t(x) &= \dot{\alpha}_t \textcolor{metablue}{\E\brac{X_1 \vert X_t=x}} + \dot{\sigma}_t \textcolor{red}{\E\brac{X_0 \vert X_t=x}}  \\ \label{e:x_1_param}
    &= \frac{\dot{\sigma}_t}{\sigma_t}x + \brac{\dot{\alpha}_t - \alpha_t\frac{\dot{\sigma}_t}{\sigma_t}}\textcolor{metablue}{\E\brac{X_1 \vert X_t=x}}\\ \label{e:x_0_param}
    &= \frac{\dot{\alpha}_t}{\alpha_t}x + \brac{\dot{\sigma}_t - \sigma_t\frac{\dot{\alpha}_t}{\alpha_t}}\textcolor{red}{\E\brac{X_0 \vert X_t=x}},
\end{align}
where we have used the fact that $\E\brac{Z|Z=z}=z$.
Then, denote the deterministic functions:
\begin{align}\label{e:x_1_t}
x_{1|t}(x) &= \E\brac{ X_1 \vert X_t=x} \text{ as the } \textcolor{metablue}{x_1\text{-prediction (target)}}, \\ \label{e:x_0_t}
x_{0|t}(x) &=  \E\brac{ X_0 \vert X_t=x}   \text{ as the } \textcolor{red}{x_0\text{-prediction (source)}}.
\end{align}
These provides two more opportunities to parameterize $u_t$: via the $x_1$-prediction $x_{1|t}$ \eqref{e:x_1_param} and via the $x_0$-prediction $x_{0|t}$ \eqref{e:x_0_param}.
\Cref{tab:conversion} offers conversion formulas between the parameterizations.
These parameterizations can also be learned using a Conditional Matching loss, similar to \eqref{e:cfm_loss_main}.
In particular, any function
\begin{equation}
    g_t(x) \defe \E\brac{f_t(X_0,X_1) \vert X_t=x},
\end{equation}
where $f_t(X_0,X_1)$ is a RV defined as a time-dependent function of $X_0$ and $X_1$, can be learned by minimizing a Matching loss of the form 
\begin{equation}
    \gL_{\M}(\theta) = \E_{t,X_t\sim p_t} D\parr{g_t(X_t), g_t^\theta(X_t)}.
\end{equation}
This loss has the same gradients as the Conditional Matching loss
\begin{equation}
    \gL_{\CM}(\theta) = \E_{t,(X_0,X_1)\sim \pi_{0,1}} D\parr{f_t(X_0,X_1), g_t^\theta(X_t)}.
\end{equation}
To learn $x_{1|t}$, the Conditional Matching loss employs $f_t(x_0,x_1)=x_1$, and similarly for $x_{0|t}$.
This procedure is justified by  \cref{thm:cm}, which is an immediate result from \cref{prop:bregman_gradient} when letting $X=X_t$, $Y=f_t(X_0,X_1)$, and integrating with respect to $t\sim U[0,1]$.
\begin{myframe}
\begin{theorem}\label{thm:cm}
    The gradients of the Matching loss and the Conditional Matching loss coincide for arbitrary functions $f_t(X_0,X_1)$ of $X_0,X_1$:
    \begin{equation}
        \nabla_\theta \gL_{\M}(\theta) = \nabla_\theta \gL_{\CM}(\theta).
    \end{equation}
    In particular, the minimizer of the Conditional Matching loss is the conditional expectation
    \begin{equation}        g_t^\theta(x)=\E\brac{f_t(X_0,X_1)\vert X_t=x}.
    \end{equation} 
\end{theorem}
\end{myframe}
\Cref{ex:fm_denoiser} shows how to train with $x_1$-prediction using the \fmlibrary{}.
\begin{pbox}[label={ex:fm_denoiser}]{Training an $X_1$-prediction model using the Conditional Matching (CM) objective}
\begin{minted}[linenos, breaklines, mathescape, fontsize=\footnotesize, xleftmargin=2em]{python}
import torch
from flow_matching.path import AffineProbPath
from flow_matching.solver import ODESolver
from flow_matching.utils import ModelWrapper

path: AffineProbPath = ...  
denoiser_model: torch.nn.Module = ...  # Initialize the denoiser
optimizer = torch.optim.Adam(velocity_model.parameters())

for x_0, x_1 in dataloader:  # Samples from $\pi_{0,1}$ of shape [batch_size, *data_dim]
    t = torch.rand(batch_size)  # Randomize time $t \sim U[0,1]$
    sample = path.sample(t=t, x_0=x_0, x_1=x_1)  # Sample the conditional path
    cm_loss = torch.pow(model(sample.x_t, t) - sample.x_1, 2).mean()  # CM loss 
    optimizer.zero_grad() 
    cm_loss.backward()
    optimizer.step()

# Convert from denoiser to velocity prediction
class VelocityModel(ModelWrapper):
    def __init__(self, denoiser: nn.Module, path: AffineProbPath):
        super().__init__(model=denoiser)
        self.path=path

    def forward(self, x: torch.Tensor, t: torch.Tensor, **extras) -> torch.Tensor:
        x_1_prediction = super().forward(x, t, **extras)
        return self.path.target_to_velocity(x_1=x_1_prediction, x_t=x, t=t) 

# Sample $X_1$
velocity_model = VelocityModel(denoiser=denoiser_model, path=path)
x_0 = torch.randn(batch_size, *data_dim)  # Specify the initial condition
solver = ODESolver(velocity_model=velocity_model)
num_steps = 100
x_1 = solver.sample(x_init=x_0, method='midpoint', step_size=1.0 / num_steps)
\end{minted}
\end{pbox}

\paragraph{Singularities in the velocity parameterizations.} Seemingly, the coefficients of \eqref{e:x_1_param} would blow up as $t \to 1$, and similarly for \eqref{e:x_0_param} as $t \to 0$.
If $\E\brac{X_1|X_0=x}$ and $\E\brac{X_0|X_1=x}$ exist, which is the case for $p(x)>0$ and $q(x)>0$, these are not essential singularities \emph{in theory}, meaning that the singularities in $x_{1|t}$ and $x_{0|t}$ would cancel with the singularities of the coefficients of the parameterization.
However, these singularities could be still problematic \emph{in practice} when the learnable $x^\theta_{1|t}$ and $x^\theta_{0|t}$ are by construction continuous and therefore do not perfectly regress their targets $x_{1|t}$ and $x_{0|t}$.
To understand how to fix these potential issues, recall \eqref{e:u_t_affine_x_1t_x_0t} and consider $u_0(x)=\dot{\alpha}_0\E\brac{X_1|X_0=x} + \dot{\sigma}_0 x$ as $t\too 0$, and $u_1(x)=\dot{\alpha}_1 x + \dot{\sigma}_1 \E\brac{X_0|X_1=x}$ as $t\too 1$.
These can be computed in many cases of interest.
Returning to our example $\pi_{0,1}(x_0,x_1)=\gN(x_0|0,I)q(x_1)$ and assuming $\E_{X_1}X_1=0$, it follows that $u_0(x) = \dot{\sigma}_0 x $ and $u_1(x)=\dot{\alpha}_1 x$.
These expressions can be used to fix singularities when converting from $x_{1|t}$ and $x_{0|t}$ to $u_t(x)$ as $t \to 1$ or $t \to 0$, respectively.

\subsubsection{Post-training velocity scheduler change}

Affine conditional flows admit a closed-form transformation from a marginal velocity field $u_t(x)$, based on a scheduler $(\alpha_t,\sigma_t)$ and an arbitrary data coupling $\pi_{0,1}$, to a marginal velocity field $\bar{u}_r(x)$, based on a different scheduler $(\bar{\alpha}_r,\bar{\sigma}_r)$ and the same data coupling $\pi_{0,1}$.
Such a transformation is useful to adapt a trained velocity field to a different scheduler, potentially improving sample efficiency and quality generation~\citep{karras2022elucidating,shaul2023bespoke,pokle2023training}.
To proceed, define the \emph{scale-time (ST) transformation} $(s_r,t_r)$ between the two conditional flows:
\begin{equation}\label{ea:st_between_flows}
  \bar{\psi}_r(x_0|x_1) = s_r\psi_{t_r}(x_0|x_1),
\end{equation}
where $\psi_t(x_0|x_1)=\alpha_t x_1 + \sigma_t x_0$, $\bar{\psi}_r(x_0|x_1) = \bar{\alpha}_r x_1 + \bar{\sigma}_r x_0$, and $s,t:[0,1]\too \Real_{\geq 0}$ are time-scale reparametrizations.
Solving \eqref{ea:st_between_flows} yields 
\begin{equation}\label{e:conversion}
    \begin{aligned}
    t_r &= \rho^{-1}(\bar{\rho}(r)) \\ s_r &= {\bar{\sigma}_r}/{\sigma_{t_r}},
    \end{aligned}
\end{equation}
where we define the signal-to-noise ratio by 
\begin{equation}\label{e:snr}
\begin{aligned}
    \rho(t)& =\frac{\alpha_t}{\sigma_t} \\
    \bar{\rho}(t)& =\frac{\bar{\alpha}_t}{\bar{\sigma}_t},    
\end{aligned}
\end{equation}
assumed to be an invertible function.
The marginal velocity $\bar{u}_r(x)$ for the new scheduler $(\bar{\alpha}_r,\bar{\sigma}_r)$ follows the expression
\begin{align*}\nonumber
    \bar{u}_r(x) &= \E\brac{\dot{\bar{X}}_r \big| \bar{X}_r=x} \\ \nonumber
    &\overset{(\ref{ea:st_between_flows})}{=} \E\brac{ \dot{s}_rX_{t_r} + s_r\dot{X}_{t_r}\dot{t}_r \big\vert s_rX_{t_r} = x } \\ \nonumber & = \dot{s}_r\E\brac{X_{t_r} \Big \vert X_{t_r}=\frac{x}{s_r}} + s_r\dot{t}_r \E\brac{\dot{X}_{t_r}  \Big \vert X_{t_r}=\frac{x}{s_r}} 
    \\  &= \frac{\dot{s}_r}{s_r}x + s_r \dot{t}_r u_{t_r}\parr{\frac{x}{s_r}},  
\end{align*}
where as before $\bar{X}_r=\bar{\psi}_r(X_0|X_1)$ and $X_t=\psi_t(X_0|X_1)$. This last term can be used to change a schedular post-training. \Cref{ex:fm_scheduler} shows how to change the scheduler of a velocity field trained with a variance preserving schedule to the conditional Optimal Transport schedule using the \fmlibrary{}.

\paragraph{Equivalence of schedulers.} One additional important consequence of the above formula is that all schedulers \emph{theoretically} lead to the \emph{same sampling} at time $t=1$ \citep{shaul2023kinetic}. That is,  
\begin{equation}\label{e:schedulers_equivalence}
 \bar{\psi}_1(x_0)=\psi_1(x_0), \text{ for all } x_0\in \Real^d.    
\end{equation}
To see that, denote $\bar{\psi}_r(x)$ the flow defined by $\bar{u}_t(x)$, and  differentiate $\tilde{\psi}_r(x) \defe s_r\psi_{t_r}(x)$ w.r.t.~$r$ and note that it also satisfies 
\begin{equation}
    \frac{\dd }{\dd t} \tilde{\psi}_r(x) = \bar{u}_r(\tilde{\psi}_r(x)). 
\end{equation}
Therefore, from uniqueness of ODE solutions we have that $\bar{\psi}_r(x)=\tilde{\psi}_r(x)=s_r\psi_{t_r}(x)$. Now, to avoid dealing with infinite signal-to-noise ratio assume the schedulers satisfy $\sigma_1=\eps=\bar{\sigma}_1$ for arbitrary $\eps>0$ (in addition to \eqref{e:scheduler}), then for $r=1$ we have $t_1=1$ and $s_1=1$ and therefore \cref{e:schedulers_equivalence} holds. 
\begin{pbox}[label={ex:fm_scheduler}]{Post-training scheduler change}
\begin{minted}[linenos, breaklines, fontsize=\footnotesize, xleftmargin=2em]{python}
import torch
from flow_matching.path import AffineProbPath
from flow_matching.path.scheduler import ScheduleTransformedModel, CondOTScheduler, VPScheduler
from flow_matching.solver import ODESolver
from flow_matching.utils import ModelWrapper

training_scheduler = VPScheduler()  # Variance preserving schedule
path = AffineProbPath(scheduler=training_scheduler)
velocity_model: ModelWrapper = ...  # Train a velocity model with the variance preserving schedule

# Change the scheduler from variance preserving to conditional OT schedule
sampling_scheduler = CondOTScheduler()
transformed_model = ScheduleTransformedModel(
    velocity_model=velocity_model,
    original_scheduler=training_scheduler,
    new_scheduler=sampling_scheduler
)

# Sample the transformed model with the conditional OT schedule
solver = ODESolver(velocity_model=transformed_model)
x_0 = torch.randn(batch_size, *data_dim)  # Specify the initial condition
solver = ODESolver(velocity_model=velocity_model)
num_steps = 100
x_1 = solver.sample(x_init=x_0, method='midpoint', step_size=1.0 / num_steps)
\end{minted}
\end{pbox}

\subsubsection{Gaussian paths} 
\label{subsubsec:gaussian_paths}
At the time of writing, the most popular class of affine probability paths is instantiated by the independent coupling $\pi_{0,1}(x_0,x_1)=p(x_0)q(x_1)$ and a Gaussian source distribution $p(x) = \gN(x|0, \sigma^2 I)$.
Because Gaussians are invariant to affine transformations, the resulting conditional probability paths take form 
\begin{equation}\label{e:p_t_kernel_gaussian}
    p_{t|1}(x|x_1) = \gN(x \vert \alpha_t x_1, \sigma_t^2 I).
\end{equation}
This case subsumes probability paths generated by standard diffusion models (although in diffusion the generation is stochastic and follows an SDE, it has the same marginal probabilities).
Two examples are the \highlight{Variance Preserving (VP)} and \highlight{Variance Exploding (VE)} paths \citep{song2021sde}, defined by choosing the following schedulers: 
\begin{align*}
  \alpha_t &\equiv 1, \sigma_0\gg 1, \sigma_1=0; & \text{(VP)}\\ %
  \alpha_t &= e^{-\frac{1}{2}\beta_t}, \sigma_t = \sqrt{1-e^{-\beta_t}}, \beta_0\gg 1, \beta_1=0. & \text{(VE)} %
\end{align*}
In the previous equations, ``$\gg 1$'' requires a sufficiently large scalar such that $p_0(x)=\int p_{0|1}(x|x_1)q(x_1)\dd x_1$ is close to a known Gaussian distribution for $t=0$---that is, the Gaussian $\gN(\cdot|0,\sigma_0^2 I)$ for VE, and $\gN(\cdot|0,I)$ for VP.
Note that in both cases, $p_t(x)$ does not exactly reproduce $p$ at $t=0$, in contrast to the FM paths in \eqref{e:scheduler}.

One useful quantity admitting a simple form in the Gaussian case is the \highlight{score}, defined as the gradient of the log probability.
Specifically, the score of the conditional path in \eqref{e:p_t_kernel_gaussian} follows the expression 
\begin{equation}\label{e:score_cond}
    \nabla \log p_{t|1}(x|x_1) = -\frac{1}{\sigma_t^2}\parr{x -  \alpha_t x_1}.
\end{equation}
The score of the corresponding \emph{marginal} probability path \eqref{e:p_t} is 
\begin{align}
    \nabla \log p_t(x) &= \int  \frac{\nabla p_{t|1}(x|x_1) q(x_1)}{p_t(x)} \dd x_1\\ &= 
    \int  \nabla \log  p_{t|1}(x|x_1) \frac{p_{t|1}(x|x_1) q(x_1)}{p_t(x)} \dd x_1  \\
    &= \E\brac{\nabla \log p_{t|1}(X_t|X_1) \, \vert \, X_t=x}\\
    &\overset{\eqref{e:score_cond}}{=}  \E\brac{-\frac{1}{\sigma_t^2}\parr{X_t-\alpha_t X_1} \, \vert \, X_t=x} \\
    &\overset{\eqref{e:X_t_equiv}}{=} \E\brac{-\frac{1}{\sigma_t} X_0 \, \Big\vert \, X_t=x} \\   
    \label{e:score_x_0_identity}
    &\overset{\eqref{e:x_0_t}}{=} -\frac{1}{\sigma_t}x_{0|t}(x),
\end{align}
where we borrow the notation $x_{0|t}$ from \eqref{e:x_0_t}.
The literature on diffusion refers to $x_0$-prediction ($x_{0|t}$) as \emph{noise}-prediction, or $\eps$-prediction.
The formula above shows that the score is proportional to the $x_0$-prediction, and provides a conversion rule---for the Gaussian path case---from score to other parametrizations, as shown in~\cref{tab:conversion}.

\paragraph{Kinetic optimality of marginal velocity.} 
A consequence of the conversion formulas developed above (\cref{tab:conversion}) is that the marginal velocity for Gaussian paths can be written in the form 
\begin{align} \label{e:u_t_score}
    u_t(x) &= \frac{\dot{\alpha}_t}{\alpha_t}x - \frac{\dot{\sigma}_t\sigma_t\alpha_t - \dot{\alpha}_t\sigma_t^2}{\alpha_t}\nabla\log p_t(x) \\ \label{e:u_t_grad}
    &= \nabla \brac{
    \frac{\dot{\alpha}_t}{2\alpha_t}\norm{x}^2 - \frac{\dot{\sigma}_t\sigma_t\alpha_t - \dot{\alpha}_t\sigma_t^2}{\alpha_t}\log p_t(x)} 
\end{align}
that shows $u_t(x)$ is a gradient and therefore Kinetic Optimal for the fixed marginalized Gaussian probability path $p_t(x)$ defined by $p_{t|1}(x|x_1)=\gN(x|\alpha_t x_1, \sigma_t^2I)$ (see \eg, \cite{villani2021topics} Section 8.1.2, or \cite{neklyudov2023action} Theorem 2.1).

\begin{table}
    \centering
    \renewcommand{\arraystretch}{1.5}
    \begin{tabular}{ccccc}
    \toprule \\[-4ex] %
       \backslashbox{$B$}{$A$} & velocity  & $x_1$-prediction & $x_0$-prediction & score \\ \midrule
       velocity & \cellcolor{black!10}{$0,1$} & \cellcolor{metablue!15}{$\frac{\dot{\sigma}_t}{\sigma_t},\frac{\dot{\alpha}_t\sigma_t-\dot{\sigma}_t\alpha_t}{\sigma_t}$} & \cellcolor{metablue!15}{
       $\frac{\dot{\alpha}_t}{\alpha_t},\frac{\dot{\sigma}_t\alpha_t-\dot{\alpha}_t\sigma_t}{\alpha_t}$}  & 
       \cellcolor{red!15}{$\frac{\dot{\alpha}_t}{\alpha_t},-\frac{\dot{\sigma}_t\sigma_t\alpha_t-\dot{\alpha}_t\sigma^2_t}{\alpha_t}$} \\
       $x_1$-prediction & & \cellcolor{black!10}{$0,1$} & \cellcolor{metablue!15}{$\frac{1}{\alpha_t},-\frac{\sigma_t}{\alpha_t}$} & \cellcolor{red!15}{$\frac{1}{\alpha_t},\frac{\sigma^2_t}{\alpha_t}$} \\
       $x_0$-prediction & & & \cellcolor{black!10}{$0,1$} & \cellcolor{red!15}{$0, -\sigma_t$} \\ score & & &  & \cellcolor{black!10}{$0,1$} \\ 
       \bottomrule
    \end{tabular}
    \caption{Conversion between different model parameterizations: $(a_t,b_t)$ corresponds to  $f^B_t(x) = a_t x + b_t f^A_t(x)$. The colors indicate the tranformation is relevant for \colorbox{black!10}{all paths}, \colorbox{metablue!15}{Affine paths} and \colorbox{red!15}{Gaussian paths}. The lower diagonal is computed from the upper diagonal using the inverse transformation: $f_t^A(x) = \frac{1}{b_t}\parr{-a_t x + f_t^B(x)}$ which can be expressed as the pair $-\frac{a_t}{b_t}, \frac{1}{b_t}$. Note some of these conversions have singularities, as discussed at the end of Section \ref{s:velocity_param}.}
    \label{tab:conversion}
\end{table}

\subsection{Data couplings}

In developing the Flow Matching training algorithm, we have assumed we can draw samples $(X_0,X_1)\sim \pi_{0,1}(X_0,X_1)$ from some coupling $\pi_{0,1}$ $(x_0,x_1)$ of the source $p$ and target $q$ distributions. For example, independent samples $\pi_{0,1}(x_0,x_1)=p(x_0)q(x_1)$, the simplest coupling preserving the marginal distributions $p$ and $q$, or paired samples $(X_0,X_1)\sim \pi_{0,1}$ provided as part of the dataset.
This section explores two examples of concrete couplings that can be used to train Flow Matching models.

\subsubsection{Paired data}

Dependent couplings arise naturally in learning tasks on paired data.
Consider, for instance, the task of image in-painting, where $q$ is a distribution of natural images, and $p$ is the distribution of those same images with a square region masked-out. 
Rather than transforming noise into data, our goal here is to learn a mapping from masked-out images $x_0$ to their filled counterparts $x_1$.
As this is an ill-defined problem---many filled images $x_1$ are compatible with each masked-out image $x_0$---solving this task can casted as learning to sample from the unknown, data-dependent coupling $\pi_{1|0}(x_1|x_0)$.

Based on these insights, \citet{liu2023i2sb,albergo2024datadependent} propose learning a \emph{bridge} or flow model with data-dependent couplings, a simple modification enabling a new regime of applications.
While the object of interest $\pi_{1|0}(x_1|x_0)$ is unavailable to sample, it is often the case that one can sample from the reverse dependency, $\pi_{0|1}(x_0|x_1)$.
Returning to the example of image in-painting, it is easy to mask out a filled image $X_1 \sim q$ (target sample) to produce a source sample $X_0 \sim p$.
To this end, specify
\begin{equation}
    \pi_{0,1}(x_0,x_1)=\pi_{0|1}(x_0|x_1)q(x_1).\label{e:data_dependent_couplings}
\end{equation}
Thus, we can obtain a pair $(X_0, X_1)$ by (i) drawing $X_1 \sim q$, and (ii) applying a predefined randomized transformation to obtain $X_0$ from $X_1$.
To satisfy the conditions of \cref{cor:cond_flows} (making sure the source is a density) and to     encourage diversity, we add noise when sampling from $\pi_{0|1}(x_0|x_1)$.
\citep{liu2023i2sb,albergo2024datadependent} demonstrated the capability of this approach on various applications, such as image super-resolution, in-painting, and de-blurring, outperforming methods based on guided diffusion~\citep{saharia2022palette}.

\subsubsection{Multisample couplings}

As discussed in \cref{s:kinetic_optimal}, straight probability paths yield ODEs simulations with smaller errors.
Therefore, it is natural ask: how could we change the training algorithm, so the learned velocity field induces straight(er)  trajectories?

As hinted above, straight trajectories are related to the Optimal Transport (OT) problem.
Specifically, consider a convex cost functional $c:\Real^d\rightarrow \Real_{\geq0}$ and the conditional OT flow $\psi_t(x_0|x_1)=t x_1 + (1-t)x_0$.
Then, the transport cost of the coupling admits an upper bound on the marginal transport cost \citep{liu2022flow,pooladian2023multisample}, that is:
\begin{equation}
    \E\brac{c(\psi_1(X_0)-X_0)} \leq \E \brac{c(X_1-X_0)},
\end{equation}
where the case of $c(x)=\norm{x}^2$ can be understood from the bound in \eqref{e:KE_leq_coupling} after plugging the OT solution in the l.h.s.~that satisfies $u_t(X_t)=u_t(\psi_t(x)) = \phi(x)-x$, where $\phi$ is the OT map.  
Therefore, one could construct low-cost marginal transport maps by reducing the coupling cost.
To this end, \citet{pooladian2023multisample} propose \highlight{multisample couplings}, a process to implicitly construct non-trivial joints $\pi_{0,1}(x_0, x_1)$ introducing dependencies between source and target distributions: 
\begin{enumerate}
    \item Sample  $X_0^{(i)}\sim p$ and $X_1^{(i)}\sim q$, $i\in [k]$ independently.
    \item Construct $\pi^k\in B_k$ by $\pi^k\defe\argmin_{\pi\in B_k} \E_{\pi}\brac{c(X_0^{(i)}-X_1^{(j)})}$.
    \item Sample a pair $(X_0^{(i)},X_0^{(j)})$ uniformly at random from $(X_0^{(i)},X_1^{(j)})$ for which $\pi^k(i,j)=1$.
\end{enumerate}
where $B_k$ is the polytope of $k\times k$ doubly stochastic matrices. 

The process above implicitly defines a joint distribution $\pi_{0,1}^k(x_0,x_1)$ by means of sampling.
This implicit joint preserves the marginals and obeys an optimality constraint (step 2) \citep{pooladian2023multisample}.
For $k=1$, the method reduces to independent couplings.
For $k>1$, \citet{pooladian2023multisample} show that the transport cost is reduced compared to independent couplings, that is, $\E_{(x_0,x_1)\sim\pi_{0,1}^k(x_0,x_1) } \brac{c(x_1-x_0)} \leq \E_{X_0\sim p, X_1\sim q} \brac{c(X_1-X_0)}$.
Furthermore, for the quadratic cost function, multisample couplings approach the Optimal Transport cost and induces straight trajectories as $k\to\infty$ \citep{pooladian2023multisample,tong2023improving}.

\subsection{Conditional generation and guidance}\label{sec:guided_generation}

We now consider training a generative model under a guiding signal to further control the produced samples.
This technique has proved valuable in numerous practical applications, such as image-to-image translation \citep{saharia2022palette} and text-to-image generation \citep{nichol2022glide, esser2024scaling}.
In this subsection, we assume access to \emph{labeled} target samples $(x_1, y)$, where $y\in\gY\subseteq\Real^k$ is a label or guidance variable.

\subsubsection{Conditional models} \label{sec:conditional_model}

One natural way to train a generative model under guidance is to learn to sample from the conditional distribution $q(x_1|y)$, as demonstrated by both diffusion and FM models~\citep{zheng2023guided}.
Following the FM blueprint in~\cref{fig:blueprint}, consider samples from the conditional target distribution $q(x_1|y)$ and prescribe a simple---typically but not necessarily Gaussian---source distribution $p$.
Next, design a \highlight{guided probability path} as the aggregation of conditional probability paths: 
\begin{equation}\label{e:pt_guidance}
    p_{t|Y}(x|y)=\int p_{t|1}(x|x_1)q(x_1|y)\dd x_1. %
\end{equation}
where we assume $p_{t,1|Y}(x,x_1|y)=p_{t|1}(x|x_1)q(x_1|y)$, meaning that the conditional path does not depend on $Y$. The resulting guided probability path is conditioned on the guidance variable $Y\sim p_Y$, and satisfies the marginal endpoints 
\begin{equation}
    p_{0|Y}(\cdot|y)=p(\cdot), \quad p_{1|Y}(\cdot|y)= q(\cdot|y).
\end{equation}
The \highlight{guided velocity field} takes form
\begin{equation}\label{e:u_t_guidance}
    u_t(x|y) = \int u_t(x|x_1)p_{1|t,Y}(x_1|x,y) \dd x_1, %
\end{equation}
where, by Bayes' Rule, it follows 
\begin{equation} 
p_{1|t,Y}(x_1|x,y) = \frac{p_{t|1}(x|x_1)q(x_1|y)}{p_{t|Y}(x|y)}.
\end{equation}
To show that $u_t(x|y)$ generates $p_{t|Y}(x|y)$, plug \eqref{e:pt_guidance} and \eqref{e:u_t_guidance} into \eqref{ea:marginal_trick_proof}, and realize that the FM/CFM losses remain unchanged for the guided case, and enable the same steps appearing in the proof of \cref{thm:cfm}.
In practice, we train a single neural network $u_t^\theta: \Real^d\times\Real^k\rightarrow\Real^d$ to model the guided marginal velocity field for all values of $y$.
Then, the guided version of the CFM loss \eqref{e:cfm_psi} follows the expression 
\begin{equation}\label{e:cfm_psi_guidance}
    \gL_{\CFM}(\theta) = \E_{t, (X_0,X_1,Y)\sim \pi_{0,1,Y}} D\parr{\dot{\psi}_t(X_0|X_1), u_t^\theta(X_t|Y)}.
\end{equation}

In practice, the literature in diffusion models shows that guidance is most effective in applications where a large amount of target samples $X_1$ share the same guiding signal $Y$, such as in class guidance \citep{nichol2021improved}.
However, guiding is more challenging in settings where the guidance variable $Y$ is non-repeating and complex, such as image captions.

\subsubsection{Classifier guidance and classifier-free guidance}

For flows trained with Gaussian paths, classifier guidance \citep{song2021sde,dhariwal2021beat} and classifier-free guidance \citep{ho2021classifierfree} can be applied utilizing the transformations between velocity fields and score functions for conditional distributions shown in \cref{tab:conversion} \citep{zheng2023guided}:
\begin{equation}\label{e:score_to_velocity}
    u_t(x|y) = a_t x + b_t \nabla \log p_{t|Y}(x|y).
\end{equation}
Using Bayes' rule over the guided probability path yields 
\begin{equation}
    p_{t|Y}(x|y) = \frac{p_{Y|t}(y|x)p_t(x)}{p_Y(y)}.
\end{equation}
Taking logarithms and gradients with respect to $x$, $\nabla=\nabla_x$, we arrive at the fundamental relation between the scores of the probability path $p_t(x)$ and its guided counterpart $p_{t|Y}(x|y)$:
\begin{equation}\label{e:score_classifier_guidance}
     \overbrace{\nabla\log p_{t|Y}(x|y)}^{\text{{conditional score}}} = \nabla \overbrace{\log p_{Y|t}(y|x)}^{\text{{classifier}}} +  \overbrace{\nabla\log p_t(x)}^{\text{{unconditional score}}}\hspace{-10pt}.
\end{equation}
Namely, the two are related by means of the score of a classifier model $p_{Y|t}(y|x)$ attempting to predict the guidance variable $y$ given a sample $x$.

Based on this relation, \citet{song2021sde} propose \highlight{classifier guidance}, that is sampling from the conditional model $q(x_1|y)$ by guiding an unconditional model (parameterized with $\nabla\log p_t(x)$) with a time-dependent classifier (predicting the guiding variable $y$ given $x\sim p_t(x)$).
The corresponding \mbox{velocity field then translates to:}
\begin{equation}\label{e:ut_classifier_guidance}
    \tilde{u}_t^{\theta,\phi}(x|y) = a_t x + b_t \parr{\nabla \log p_{Y|t}^\phi(y|x) + \nabla \log p_t^\theta (x)} = u_t^\theta(x) + b_t \nabla \log p_{Y|t}^\phi(y|x),
\end{equation}
where $u_t^\theta(x)$ is a velocity field trained on the unconditional target $q(x)$, and $\log p_{Y|t}^\phi(y|x)$ is a time-dependent classifier with parameters $\phi\in\Real^{m}$.
\cite{dhariwal2021beat} show that this approach outperforms the conditional model from \cref{sec:conditional_model} for both class- and text-conditioning \citep{nichol2022glide}.
In practice, because the classifier and the unconditional score are learned separately, it is often necessary to calibrate the classifier guidance as 
\begin{equation}\label{e:ut_classifier_guidance_calibrated}
    \tilde{u}_t^{\theta,\phi}(x|y) = u_t^\theta(x) + b_t w\nabla \log p_{Y|t}^\phi(y|x),
\end{equation}
where $w\in \Real$ is the classifier scale, typically chosen to be $w>1$ \citep{dhariwal2021beat}.

In a later work, \citep{ho2021classifierfree} propose a pure generative approach called \highlight{classifier-free guidance}.
By simply re-arranging \eqref{e:score_classifier_guidance}, we obtain 
\begin{equation}\label{e:score_classifier_free_guidance}
     \nabla \overbrace{\log p_{Y|t}(y|x)}^{\text{{classifier}}} = \overbrace{\nabla\log p_{t|Y}(x|y)}^{\text{{conditional score}}} \ \  - \overbrace{\nabla\log p_t(x)}^{\text{{unconditional score}}}\hspace{-10pt},
\end{equation}
revealing that the score of the classifier can be implicitly approximated by the difference between the scores of the vanilla and guided probability paths.
Then, the authors propose to learn the conditional and unconditional scores simultaneously using the same model.
In terms of velocities, \cite{zheng2023guided} show one can also plug \ref{e:score_classifier_free_guidance} into \ref{e:ut_classifier_guidance_calibrated} and use the conversion from scores to velocities as in \cref{tab:conversion} to get:
\begin{equation}\label{e:ut_cfg}
    \tilde{u}_t^{\theta}(x|y) = (1-w)u_t^\theta(x|\emptyset) + wu_t^\theta(x|y),
\end{equation}
where $w$ is once again the guidance calibration scale. Now, only a single model is trained, $u_t^\theta(x|y)$, where $y\in\set{\gY,\emptyset }$, $\emptyset$ is a place-holder value denoting the null-condition, and $u_t^\theta(x|\emptyset)$ is the velocity field generating the unconditional probability path $p_t(x)$.
The resulting loss reads:
\begin{equation}\label{e:cfm_psi_cfg}
    \gL_{\CFM}(\theta) = \E_{t,\xi, (X_0,X_1,Y)\sim \pi_{0,1,Y}} \brac{D\parr{\dot{\psi}_t(X_0|X_1), u_t^\theta(X_t|(1-\xi)\cdot Y + \xi \cdot \emptyset)}},
\end{equation}
where $\xi\sim\text{Bernoulli}(p_{\mathrm{uncond}})$, and $p_{\mathrm{uncond}}$ is the probability of drawing the null condition $\emptyset$ during training.
The exact distribution which CFG samples from is unknown, with some works proposing different intuitive or theoretical justifications for CFG sampling~\citep{dieleman2022guidance,guo2024gradient,chidambaram2024does,bradley2024classifier}. Despite this, at the time of writing, CFG is the most popular approach to training a conditional model.
\citet{esser2024scaling,polyak2024moviegencastmedia} show the application of classifier-free guidance to train large-scale guided FM models.

\pagebreak
\section{Non-Euclidean Flow Matching}\label{s:fm_non_euclidean}

This section extends Flow Matching from Euclidean spaces $\Real^d$ to general \emph{Riemannian manifolds} $\gM$.
Informally, Riemannian manifolds are spaces behaving locally like Euclidean spaces, and are equipped with a generalized notion of distances and angles.
Riemannian manifolds are useful to model various types of data.
For example, probabilities of natural phenomena on Earth can be modeled on the sphere \cite{mathieu2020riemannian}, and protein backbones are often parameterized inn terms of matrix Lie groups \cite{jumper2021highly}.
The extension of flows to Riemannian manifolds is due to \citet{mathieu2020riemannian,lou2020riemannian}.
However, their original training algorithms required expensive ODE simulations.
Following \citet{chen2024flow}, the Flow Matching solutions in this section provide a scalable, simulation-free training algorithm to learn generative models on Riemannian manifolds.

\subsection{Riemannian manifolds}

We consider complete connected, smooth Riemannian manifolds $\gM$ with a metric $g$.
The tangent space at point $x\in\gM$, a vector space containing all tangent vectors to $\gM$ at $x$, is denoted with $T_x\gM$.
The Riemannian metric $g$ defines an inner product over $T_x \gM$ denoted by $\ip{u,v}_g$, for $u,v\in T_x\gM$.
Let $T\gM=\cup_{x\in \gM} \set{x}\times T_x\gM$ be the tangent bundle that collects all the tangent planes of the manifold.
In the following, vector fields defined on tangent spaces are important objects to build flows on manifolds with velocity fields.
We denote by $\gU=\set{u_t}$ the space of time-dependent smooth vector fields (VFs) $u_t :[0,1]\times \gM\too T\gM$, where $u_t(x)\in T_x\gM$ for all $x\in\gM$.
Also, $\divv_g(u_t)$ is the Riemannian divergence with respect to the spatial ($x$) argument.
Finally, we denote by $\dd \vol_x$ the volume element over $\gM$, and integration of a function $f:\gM\too\Real$ over $\gM$ is denoted $\int f(x) \dd \vol_x$.%

\subsection{Probabilities, flows and velocities on manifolds}\label{s:flows_on_manifolds}

Probability density functions over a manifold $\gM$ are continuous non-negative functions $p:\gM\rightarrow\Real_{\geq 0}$ integrating to $1$, namely $\int_\gM p(x) \dd \vol_x=1$.
We define a probability path in time $p_t$ as a time-dependent curve in probability space $\gP$, namely $p_t:[0,1]\too\gP$.
A time-dependent flow, $\psi:[0,1]\times \gM\too\gM$, similar to the Euclidean space, defines a global diffeomorphism on $\gM$ for every $t$.

Remarkably, constructing flow-based models via velocity fields naturally applies to general Riemannian manifolds.
Formally, and rephrasing Proposition 1 from \cite{mathieu2020riemannian}:
\begin{myframe}
    \begin{theorem}[Flow local existence and uniqueness]\label{thm:riemannian_ode_existence_and_uniqueness}
    Let $\gM$ a smooth complete manifold and a velocity field $u_t \in \gU$. If $u$ is $C^\infty([0,1]\times\gM,T\gM)$ (in particular, locally Lipschitz), then the ODE in \eqref{e:flow} has a unique solution which is a  $C^\infty(\Omega,\gM)$ diffeomorphism $\psi_t(x)$ defined over the open set $\Omega \supset \set{0}\times \gM$.
    \end{theorem}
\end{myframe}

Similar to \cref{thm:ode_existence_and_uniqueness}, flow ODEs generally only define a local diffeomorphism on the manifold, meaning that $\psi_t(x)$ may be defined on a maximal interval in time $[0,t_x])$ for different values of $x\in\gM$. Similar to the Euclidean case we will work with the semi-open time interval $t\in[0,1)$ to allow $q$ to have compact support (for which $u_t$ is not everywhere defined). To ensure existence for the desired time interval, $[0,1)$, we add the integrability constraint (see \cref{thm:continuity}) and rely on the Mass Conservation theorem once again.
For a Riemannian manifold with metric $g$, the \highlight{Riemannian continuity equation} reads
\begin{equation}\label{e:riemannian_continuity}
    \frac{\dd}{\dd t} p_t(x)+ \divv_g(p_t u_t)(x) = 0,%
\end{equation}
and the corresponding Manifold Mass Conservation theorem \citep{villani2009optimal} is stated as follows.
\begin{myframe}
\begin{theorem}[Manifold Mass Conservation]\label{thm:riemannian_continuity}
    Let $p_t$ be a probability path and $u_t\in\gU$ a locally Lipchitz integrable vector field over a Riemannian manifold $\gM$ with metric $g$. Then the following are equivalent
    \begin{enumerate}
        \item The Continuity Equation \eqref{e:riemannian_continuity} holds for $t\in [0,1)$.
        \item $u_t$ generates $p_t$ in the sense of \ref{def:generates}.
    \end{enumerate}
      
\end{theorem}
\end{myframe}
In the previous result, by integrable $u$ we mean 
\begin{equation}\label{e:manifold_integrable}
 \int_0^1\int_\gM \norm{u_t(x)}p_t(x)\dd \vol_x \dd t < \infty. %
\end{equation}
Note that the assumptions of \cref{thm:riemannian_continuity} yield a global diffeomorphism on $\gM$, giving rise to the \highlight{Riemannian instantaneous change of variables} formula:
\begin{equation}
    \label{e:r_instant_div} \frac{\dd}{\dd t} \log p_t(\psi_t (x)) = -\divv_g(u_t)(\psi_t(x)).%
\end{equation}
Finally, we say that $u_t$ generates $p_t$ from $p$ if 
\begin{equation}
    X_t=\psi_t(X_0)\sim p_t \text{ for } X_0\sim p.
\end{equation}

Having positioned flows as valid generative models on manifolds, it stands to reason that the FM principles can be transferred to this domain as well.
In the Riemannian version of FM we aim to find a velocity field $u_t^\theta\in\gU$ generating a target probability path $p_t:[0,1]\too\gP$ with marginal constraints $p_0=p$ and $p_1=q$, where $p,q$ denote the source and target distributions over the manifold $\gM$.
As the velocity field lies on the tangent spaces of the manifold, the \highlight{Riemannian Flow Matching loss} compares velocities using a Bregman divergence defined over the individual tangent planes of the manifold, 
\begin{equation}\label{e:rfm_loss}
    \gL_{\RFM}(\theta) = \E_{t,X_t\sim p_t}D_{X_t}\parr{u_t(X_t),u_t^\theta(X_t)}. %
\end{equation}
In the equation above, the expectation is now an integral over the manifold, that is $E[f(X)]=\int_\gM f(x)p_X(x)\dd \vol_x$, for a smooth function $f:\gM\too\gM$ and a random variable $X\sim p_X$.
The Bregman divergences, $D_x$, $x\in\gM$, are potentially defined with the Riemannian inner product and a strictly convex function assigned to each tangent space $\Phi_x:T_x\gM\too T_x\gM$, that is, $D_{x}(u,v) \defe \Phi_x(u) - \brac{ \Phi_x(v) + \langle u-v, \nabla_v \Phi_x(v) \rangle_g}$. For example, choosing the Riemannian metric $\Phi_x=\norm{\cdot}_g^2$ then  $D_x(u,v)=\norm{u-v}^2_g$  for $u,v\in T_x\gM$.

\subsection{Probability paths on manifolds}

\highlight{Marginal probability paths} are built as in the Euclidean case \eqref{e:p_t}:
\begin{equation}\label{e:riemannian_p_t}
    p_t(x) = \int_\gM p_t(x|x_1)q(x_1) \dd \vol_{x_1},%
\end{equation}
where $p_{t|1}(x|x_1)$ is the \highlight{conditional probability path} defined on the manifold.
We also require the boundary constraints
\begin{equation}\label{e:riemannian_p_q_interp}
    p_0=p, \quad p_1=q. %
\end{equation}
For instance, these constraints can be implemented by requiring the conditional path $p_{t|1}(x|x_1)$ to satisfy
\begin{equation}\label{e:riemannian_p_t_cond_boundary}
    p_{0|1}(x|x_1) = \pi_{0|1}(x|x_1), \text{ and  } p_{1|1}(x|x_1)=\delta_{x_1}(x),
\end{equation}
where $\pi_{0|1}$ is the conditional coupling, $\pi_{0|1}(x_0|x_1)=\pi_{0,1}(x_0,x_1)/q(x_1)$.

\subsection{The Marginalization Trick for manifolds}

The Marginalization Trick for the marginal velocity field (\cref{thm:fm_main}) readily applies to the Riemannian case.
Consider the \highlight{conditional velocity field} $u_t(x|x_1)\in\gU$ such that 
\begin{equation}\label{e:riemannian_u_t|1_generates_p_t|1}
    u_{t}(\cdot|x_1) \text{ generates } p_{t|1}(\cdot|x_1).
\end{equation}

Then, the \highlight{marginal velocity} field $u_t(x)$ is given by the following averaging of the conditional velocities,
\begin{equation}\label{e:riemanniam_u_t}
    u_t(x) = \int_\gM u_t(x|x_1)p_{1|t}(x_1|x) \dd \vol_{x_1}, %
\end{equation}
where, by Bayes' Rule for PDFs, we obtain
\begin{equation} 
p_{1|t}(x_1|x) = \frac{p_{t|1}(x|x_1)q(x_1)}{p_t(x)},
\end{equation}
which is defined for all $x\in\gM$ for which $p_t(x)>0$.

The Marginalization Trick (\cref{thm:fm_main}) for the Riemannian case requires adjusting \cref{as:p_t} as follows:
\begin{myframe}
\begin{assumption}\label{as:riemannian_p_t}
$p_{t|1}(x|x_1)$ is $C^\infty([0,1)\times \gM)$ and $u_t(x|x_1)$ is $C^\infty([0,1)\times \gM,\gM)$ as function of $(t,x)$. Furthermore, we assume either $q$ has bounded support, \ie, $q(x_1)=0$ outside some bounded set or $\gM$ is compact; and $p_t(x)>0$ for all $x\in\gM$ and $t\in[0,1)$.
\end{assumption}    
\end{myframe}
We are now ready to state the \highlight{Manifold  Marginalization Trick} theorem:
\begin{myframe}
    \begin{theorem}[Manifold Marginalization Trick]\label{thm:rfm_main} Under Assumption \ref{as:riemannian_p_t}, if $u_t(x|x_1)$ is conditionally integrable and generates the conditional probability path $p_t(\cdot|x_1)$ then the marginal velocity field $u_t(\cdot)$ generates the marginal probability path $p_t(\cdot)$.
\end{theorem}
\end{myframe}
By \highlight{conditionally integrable}, we mean a conditioned version of the integrability condition from the Mass Conservation Theorem \eqref{e:manifold_integrable}:
\begin{align}\label{e:riemannian_conditional_integrable}
    \int^1_0\int_\gM\int_\gM \norm{u_t(x|x_1)}_g p_{t|1}(x|x_1) q(x_1) \dd \vol_{x_1} \dd \vol_{x} \dd t &< \infty
\end{align}
The proof of \cref{thm:rfm_main} is repeating the arguments of \cref{thm:fm_main} and is given in \cref{a:manifold_marginalization_trick}.

\subsection{Riemannian Flow Matching loss}

The \highlight{Riemannian Conditional Flow Matching (RCFM) loss} reads
\begin{equation}\label{e:rcfm_loss}
    \gL_{\RCFM}(\theta) = \E_{t,X_1,X_t\sim p_{t|1}(\cdot|X_1)}D_{X_t}\parr{u_t(X_t|X_1),u_t^\theta(X_t)}.%
\end{equation}
Once again, we have the equivalence: 
\begin{myframe}
\begin{theorem}\label{thm:rcfm}
    The gradients of the Riemannian Flow Matching loss and the Riemannian Conditional Flow Matching loss coincide:
    \begin{equation}
        \nabla_\theta \gL_{\RFM}(\theta) = \nabla_\theta \gL_{\RCFM}(\theta).
    \end{equation}
\end{theorem}
\end{myframe}

The above theorem can be proved using \cref{prop:bregman_gradient} with $X=X_t$, $Y=u_t(X_t|X_1)$, $g^\theta(x)=u_t^\theta(x)$, and integrating w.r.t.~$t\in[0,1]$.

\subsection{Conditional flows through premetrics}

Having established how to learn a flow model with the RCFM loss, we are left with specifying the conditional probability path and its generating velocity field.
Similar  to \cref{s:conditional_flows}, we begin by stating the requirements for the corresponding conditional flow $\psi:[0,1)\times \gM \times \gM \too \gM$, such that $p_{t|1}(\cdot|x_1)$ satisfies the boundary conditions \eqref{e:riemannian_p_t_cond_boundary}.
The \highlight{conditional flow model} is
\begin{equation}
    X_{t|1} = \psi_t(X_0|x_1), \quad \text{ where } X_0\sim \pi_{0|1}(\cdot|x_1), %
\end{equation}    
where the \highlight{conditional flow} is
\begin{equation}\label{e:riemannian_psi}
    \psi_t(x|x_1) = \begin{cases}
        x & t=0 \\
        x_1 & t=1
    \end{cases}, \ \text{ is smooth in } t,x \text{ and diffeomorphism in $x$ on } \gM. %
\end{equation}

Our analysis in Euclidean space focused on affine conditional flows, as these served as a rich class of easily computable (simulation-free) conditional flows.
Unfortunately,  combinations $\alpha_t x_1 + \sigma_t x_0$ for $\alpha_t+\sigma_t\ne1$ are not naturally defined on manifolds. The manifold analog for the case $\alpha_t+\sigma_t=1$ would be using  geodesic interpolation. 
Indeed, \citet{chen2024flow} proposed building conditional flows by moving along geodesic curves, in particular, generalizing the conditional OT paths moving along straight lines in Euclidean space (see \cref{thm:cond_ot}). Geodesics represent the shortest paths between two points on a manifold, reducing to straight lines in Euclidean spaces.
For manifolds, we define the \highlight{geodesic conditional flow} as
\begin{equation}\label{e:exp_log}
    \psi_t (x_0|x_1)= \exp_{x_0} (\kappa(t) \log_{x_0}(x_1)), \quad t\in[0,1],%
\end{equation}
where $\kappa(t):[0,1]\too[0,1]$ is a monotonically increasing scheduler satisfying $\kappa(0)=0$ and $\kappa(1)=1$, making sure all $x_0$ are pushed to $x_1$ at $t=1$. The exponential map, evaluated at $x\in\gM$, $\exp_x:T_x\gM\too \gM$, $v\mapsto \exp_x(v)$, returns the endpoint at time $t=1$ of the unique geodesic starting at $x$ with initial speed $v$. The logarithmic map $\log_{x}:\gM\too  T_xM$, $y\mapsto \log_x(y)$, is the inverse of the exponential map. In Euclidean space, the exponential map is simply vector addition, and the logarithmic map is vector subtraction.
Now, if we plug these in \eqref{e:exp_log}, we get $\psi_t (x_0|x_1)= x_0 + \kappa(t)(x_1-x_0)$, and by choosing $\kappa(t)=t$ we recover the conditional OT flow.

For simple manifolds with closed-form exponential and logarithmic maps, this construction allows a simulation-free recipe for training flows on manifolds, an arguably clear advantage compared to diffusion models approaches built on manifolds \citep{de2022riemannian,huang2022rimannian,lou2023scaling}.
In particular, manifold diffusion models require in-training simulation to sample from $p_t$, and have to resort to approximations of the score function on the manifold.  

Nevertheless, while building geodesic conditional flows is a natural construction, geodesics may be hard to compute for general manifolds that do not have closed-form exponential and logarithmic maps {and/or introduce undesired bias such as concentrating probability at boundary points.} 
To overcome the difficulty in computing geodesics and/or inject a desired implicit bias, one may seek an alternative notion of smooth distance function, $\dist(\cdot,\cdot):\gM\times\gM\too\Real_{\geq 0}$, and require that the conditional flow satisfies
\begin{equation}\label{e:dist_cond_flow}
    \dist(\psi_t(x_0|x_1),x_1) = \bar{\kappa}(t)\dist(x_0,x_1),
\end{equation}
where $\bar{\kappa}(t)=1-\kappa(t)$. This will assure that the conditional flow concentrates all the probability at $x_1$ at time $t=1$ if the following conditions hold:
\begin{enumerate}
    \item \emph{Non-negative}: $\dist(x,y)\geq 0$ for all $x,y\in \gM$.
    \item \emph{Positive}: $\dist(x,y)=0$ if and only if $x=y$.
    \item \emph{Non-degenerate}: $\nabla \dist(x,y)\neq 0$ if and only if $x\neq y$.
\end{enumerate}

\citet{chen2024flow} showed that the minimal norm conditional velocity field corresponding to a flow that satisfies \eqref{e:dist_cond_flow} has the form:
\begin{equation}\label{e:riemannian_cond_ut}
    u_t(x|x_1) = \frac{d \log \bar{\kappa}(t)}{\dd t}\dist(x,x_1)\frac{\nabla \dist(x,x_1)}{\norm{\nabla \dist(x,x_1)}^2_g},
\end{equation}      
\begin{wrapfigure}[13]{r}{0.4\textwidth}
  \begin{center} \vspace{-10pt}   \includegraphics[width=0.38\textwidth]{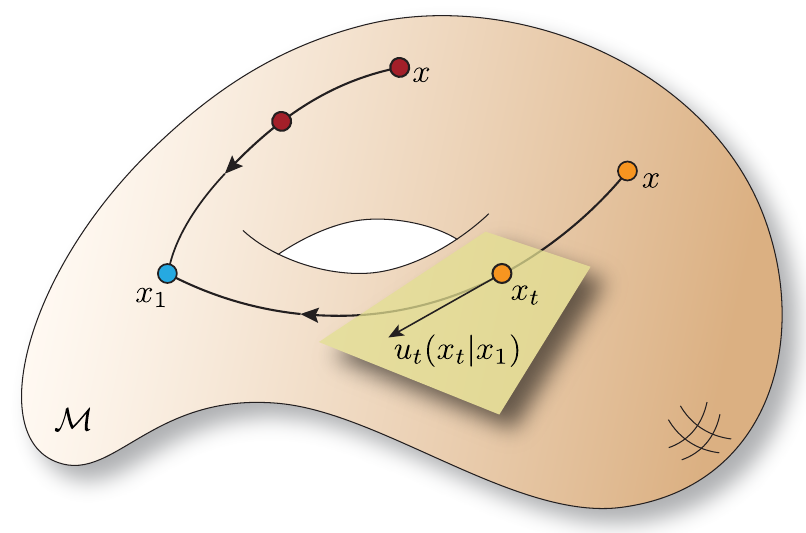}
  \end{center}
  \caption{Conditional flows on the manifold $\gM$.}\label{fig:rfm}
\end{wrapfigure}
where the non-degeneracy requirement of the premetric ensures that the velocity field has no discontinuities, since $u_t(x|x_1)\propto 1/{\norm{\nabla \dist(x,x_1)}_g}$. In particular, note that the geodesic conditional flow in \eqref{e:exp_log} satisfies \eqref{e:dist_cond_flow} for the choice $\dist=\dist_g$, where $\dist_g$ is the geodesic distance. An example of a choice of alternative premetrics is using spectral distances on general geometries \citep{chen2024flow}, where the conditional velocity offers a way to sample from $p_t(x|x_1)$ by simulation. Importantly, although conditional flows with premetrics require in-training simulation---like diffusion models on manifolds---the velocity field can still be accurately recovered compared to approximations of the score function.

Another issue, is that both conditional flows defined via geodesic interpolation and premetric can suffer from singularities, \eg, for compact manifolds. For example on the 2-sphere the geodesic function $\dist(x,x_1)$ is not differentiable at the antipodal point $x=-x_1$. Furthermore, any smooth function such as $x\mapsto \dist(x,x_1)$ will showcase at-least two critical points (maximum and minimum) where the velocity in \eqref{e:riemannian_cond_ut} is not-defined. However, the set of such  problematic points is generally very small (in fact of zero volume usually). Therefore, this issue does not cause problems in practice, at-least in use cases we are aware of.

In any case, to deal with this issue, we can include an augmented scheduler in the geodesic conditional flow. That is, use  $\bar{\kappa}(t,x,x_1)$, that depends also on $x,x_1$ to make \eqref{e:exp_log} globally smooth. To deal with the zero gradient issue of the premetric conditional flow we can relax the non-degeneracy requirement as follows:
\begin{itemize}
    \item[3.]  \emph{Non-degenerate (relaxed)}: The volume of the set $\gA_y = \set{x\in \gM \, \vert \,  \nabla \dist(x,y)=0 \text{ and } x\ne y } $ is $0$ for all $y\in \gM$.
\end{itemize}

\begin{pbox}[label={ex:fm_riemannian}]{Training with geodesic flows on a Sphere using the CFM objective}
\begin{minted}[linenos, breaklines, mathescape, fontsize=\footnotesize, xleftmargin=2em]{python}
import torch
from flow_matching.path import GeodesicProbPath, PathSample
from flow_matching.path.scheduler import CondOTScheduler
from flow_matching.utils.manifolds import Sphere

model = ...  # Define a trainable velocity model
optimizer = torch.optim.Adam(model.parameters())
loss_fn = torch.nn.MSELoss()  # Any Bregman divergence

manifold = Sphere()
scheduler = CondOTScheduler()
path = GeodesicProbPath(scheduler=scheduler, manifold=manifold)  

for x_0, x_1 in dataloader:  # Samples from $\pi_{0,1}$ of shape [batch_size, *data_dim]
    t = torch.rand(batch_size)  # Randomize time $t \sim U[0,1]$
    sample: PathSample = path.sample(t=t, x_0=x_0, x_1=x_1)  # Sample the conditional path
    
    model_output = model(sample.x_t, sample.t)
    projected_model_output = manifold.proju(sample.x_t, model_output)  # Project to tangent space
    
    loss = loss_fn(projected_model_output, sample.dx_t)  # CFM loss
    
    optimizer.zero_grad()  
    loss.backward()
    optimizer.step()
\end{minted}
\end{pbox}

\pagebreak
\section{Continuous Time Markov Chain Models}\label{sec:ctmcm}

This section presents the \highlight{Continuous Time Markov Chains (CTMCs)} as an alternative generative model to flow, with the use-case of generating discrete data, \ie, data residing in a discrete (and finite) state space. 
CTMC are Markov processes that form the building blocks behind the generative model paradigm of Discrete Flow Matching (DFM) \cite{campbell2024generative,gat2024discrete}, later discussed in~\cref{sec:discreteflow}.
Therefore, this section is analogous to \cref{s:flow_models}, where we presented flows as the building blocks behind the generative model paradigm of Flow Matching (FM).

\subsection{Discrete state spaces and random variables}

Consider a finite version of $\Real^d$ as our state space $\gS = \gT^d$, where $\gT = [K] = \set{1,2,\ldots,K}$, sometimes called \highlight{vocabulary}.
Samples and states are denoted by $x=(x^1,\ldots,x^d)\in \gS$, where $x^i\in \gT$ is single coordinate or a \highlight{token}. We will similarly use states $y,z\in\gS$.
Next, $X$ denotes a random variable taking values in the state space $\gS$, with probabilities governed by the \highlight{probability mass function (PMF)} $p_X:\gS\too\Real_{\geq 0}$, such that $\sum_{x\in \gS}p_X(x) = 1$, and the probability of an event $A\subset \gS$ being
\begin{equation}
    \sP(X\in A) = \sum_{x\in A} p_X(x).
\end{equation}
The notations $X\sim p_X$ or $X\sim p_X(X)$ indicate that $X$ has the PMF $p_X$.
The $\delta$ PMF in the discrete case is defined by 
\begin{equation}
    \delta(x,z) = \begin{cases}
        1 & x=z, \\ 0 & \text{else}.
    \end{cases}
\end{equation}
where we sometimes also define $\delta$ PMFs on tokens, such as in $\delta(x^i,y^i)$, for some $x^i,y^i\in \gT$.

\subsection{The CTMC generative model}

The \highlight{CTMC model} is an $\gS$-valued time-dependent family of random variables $(X_t)_{0\leq t \leq 1}$ that a form a Markov chain characterized by the \highlight{probability transition kernel} $\tkernel_{t+h|t}$ defined via
\begin{myframe}
\begin{equation}
    \tkernel_{t+h|t}(y|x) \defe \sP(X_{t+h}=y \vert X_t = x) = \delta(y,x) + h u_t(y,x) + o(h), \text{ and } 
    \sP(X_0 = x)= p(x),\label{e:ctmc_model} %
    \end{equation}
\end{myframe}

\begin{wrapfigure}[13]{r}{0.3\textwidth}
  \includegraphics[width=0.3\textwidth]{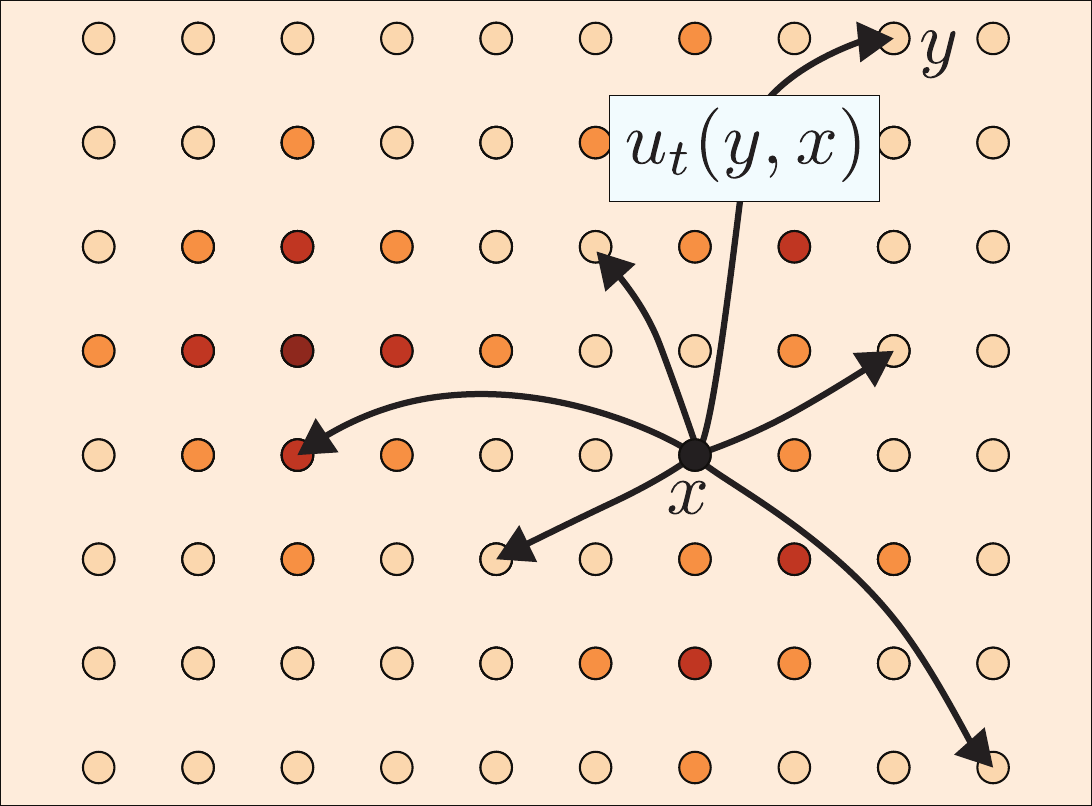}  
  \caption{The CTMC model is defined by prescribing rates (velocities) of probability between states.}\label{fig:ctmc_general}
\end{wrapfigure}

where the PMF $p$ indicates the initial distribution of the process at time $t=0$, and $o(h)$ is an arbitrary function satisfying $o(h)/h\too 0$ as $t\too 0$.
The values $u_t(y,x)$, called \highlight{rates} or \highlight{velocities}, indicate the speed at which the probability transitions between states as a function of time.
By fully characterized, we mean that all the joints $\sP(X_{t_1}=x_1,\ldots,X_{t_n}=x_n)$, for arbitrary $0\leq t_1 < \cdots < t_n \leq 1$ and $x_{i}\in\gS$, $i\in[n]$, are defined this way.

To make sure the transition probabilities $p_{t+h|t}(y|x)$ are defined via \eqref{e:ctmc_model}, velocities needs to satisfy the following \highlight{rate conditions}:
\begin{equation}
    u_t(y,x)\geq 0 \text{ for all } y\ne x\text{, and } \sum_y u_t(y,x)=0. \label{e:rate_conds}%
\end{equation}
If one of these conditions were to fail, then the transition probabilities $p_{t+h|t}(\cdot|x)$ would become negative or sum to $c \neq 1$ for arbitrary small $h>0$.
\Cref{e:ctmc_model} plays he same role as \cref{e:flow_model} and \cref{e:flow} when we were defining the flow generative modeling.
The \emph{marginal probability} of the process $X_t$ is denoted by the PMF $p_t(x)$ for time $t \in[0,1]$.
Then, similarly to \cref{def:generates} for the case of flows, we say that
\par %
\begin{myframe}
\begin{equation}\label{def:generates_ctmc}
    u_t \text{ \highlight{generates} } p_t \text{ if there exists } p_{t+h|t} \text{ satisfying } \eqref{e:ctmc_model} \text{ with marginals } p_t.
\end{equation}
\end{myframe} 

\paragraph{Simulating CTMC.}

To sample $X_t$, sample $X_0\sim p$ and take steps using the \highlight{(naive) Euler method}:
\begin{equation}\label{e:ctmc_euler_naive}
    \sP(X_{t+h} = y \ \vert \ X_t ) =  \delta(y,X_t) + hu_t(y,X_t). %
\end{equation}
According to \eqref{e:ctmc_model}, these steps introduce $o(h)$ errors to the update probabilities.
In practice, this means that we would need a sufficiently small $h>0$ to ensure that the right-hand side in \eqref{e:ctmc_euler_naive} remains a valid PMF.
One possible remedy to assure that any choice of $h>0$ results in a valid PMF, and maintains the $o(h)$ local error in probabilities is the following \highlight{Euler method}:
\begin{align}\label{e:ctmc_euler}
    \sP(X_{t+h} = y \ \vert \ X_t ) =  \begin{cases}
       \exp\brac{h u_t(X_t,X_t)} & y=X_t\\
       \frac{u_t(y,X_t)}{\abs{u_t(X_t,X_t)}}\parr{1-\exp\brac{hu_t(X_t,X_t)}} & y\ne X_t
    \end{cases}.%
\end{align}

\subsection{Probability paths and Kolmogorov Equation} 
Similarly to Continuity Equation in the continuous case, the marginal probabilities $p_t$ of the CTMC model $(X_t)_{0\leq t \leq 1}$  are characterized by the \highlight{Kolmogorov Equation}
\begin{equation}\label{e:kolmogorov}
    \frac{\dd}{\dd t}p_t(y) = \sum_{x} u_t(y,x) p_t(x). %
\end{equation}
The following classical theorem (see also Theorems 5.1 and 5.2 in \cite{coddington1956theory}) describes the existence of unique solutions for this linear homogeneous system of ODEs.
\begin{myframe}\begin{theorem}[Linear ODE existence and uniqueness]\label{thm:linear_system_ode_existence_and_uniqueness}
      If $u_t(y,x)$ are in $C([0,1))$ (continuous with respect to time), then there exists a unique solution $p_t(x)$ to the Kolmogorov Equation \eqref{e:kolmogorov}, for $t\in [0,1)$ and satisfying $p_0(x)=p(x)$.
    \end{theorem}
\end{myframe}
For the CTMC, the solution is guaranteed to exist for all times $t\in[0,1)$ and no extra conditions are required (unlike the non-linear case in \cref{thm:ode_existence_and_uniqueness}).
The Kolmogorov Equation has an intimate connection with the Continuity Equation \eqref{e:continuity}.
Rearranging the right-hand side of \eqref{e:kolmogorov} by means of the rate conditions yields 
\begin{align*}
    \sum_x u_t(y,x)p_t(x) &\overset{\eqref{e:rate_conds}}{=} 
    \overbrace{\sum_{x \ne y} u_t(y,x) p_t(x)}^{\text{{incoming flux}}} - \overbrace{\sum_{x \ne y} u_t(x,y) p_t(y)}^{\text{{outgoing flux}}} \\
    &\,\,\,= -\sum_{x\ne y} \brac{j_t(x,y) - j_t(y,x)},
\end{align*}
where $j_t(y,x)\defe u_t(y,x)p_t(x)$ is the \emph{probability flux} describing the probability of moving from state $x$ to state $y$ per unit of time.
The excess of outgoing flux is defined as the \highlight{divergence}, giving the Kolmogorov Equation the same structure as the one described in \cref{ss:continuity_equation} for the Continuity Equation \citep{gat2024discrete}. 

The following result is the main tool to build probability paths and velocities in the CTMC framework:
\begin{myframe}
    \begin{theorem}[Discrete Mass Conservation]\label{thm:discrete_mass_conservation}
    Let $u_t(y,x)$ be in $C([0,1))$ and $p_t(x)$ a PMF in $C^1([0,1))$ in time $t$. Then, the following are equivalent:
    \begin{enumerate}
        \item $p_t,u_t$ satisfy the Kolmogorov Equation \eqref{e:kolmogorov} for $t\in[0,1)$, and $u_t$ satisfies the rate conditions \eqref{e:rate_conds}.
        \item $u_t$ generates $p_t$ in the sense of \ref{def:generates_ctmc} for $t\in[0,1)$.
    \end{enumerate}
    \end{theorem}
\end{myframe}
The proof of \cref{thm:discrete_mass_conservation} is given in \cref{a:discrete_mass_conservation}.

\pagebreak
\subsubsection{Probability preserving velocities}\label{s:probability_preserving}

As a consequence of the Discrete Mass Conservation (\cref{thm:discrete_mass_conservation}), if velocity $u_t(y,x)$ generates the probability path $p_t(x)$, then 
\begin{equation}
    \tilde{u}_t(y,x) =  u_t(y,x)+v_t(y,x) \text{ generates } p_t(x),
\end{equation}
as long as $v_t(y,x)$ satisfies the rate conditions \eqref{e:rate_conds} and solves the \highlight{divergence-free velocity} equation 
\begin{equation}\label{e:discrete_div_free}
    \sum_x v_t(y,x) p_t(x) = 0. %
\end{equation}
In fact, $\tilde{u}_t(y,x)$ solves the Kolmogorov Equation:
\begin{equation*}
    \sum_x \tilde{u}_t(y,x) p_t(x) = \sum_x u_t(y,x) p_t(x) = \dot{p}_t(y),
\end{equation*}
showing that one may add divergence-free velocities during sampling without changing the marginal probability.
This will be a useful fact when sampling from discrete Flow Matching models, described next.

\section{Discrete Flow Matching}\label{sec:discreteflow}

Remarkably, the Flow Matching blueprint in~\cref{fig:blueprint} carries out seamlessly from the continuous case to the discrete case, yielding the \highlight{Discrete Flow Matching (DFM)} framework~\citep{campbell2024generative, gat2024discrete}.
In analogy to the continuous case, start by defining a probability path $p_t$ interpolating between a source PMF $p$ and a target PMF $q$.
Second, we would like to find a CTMC model $(X_t)_{0\leq t \leq 1}$, defined by a learnable velocity $u^\theta_t$, that generates the probability path $p_t$.
Finally, we train $u^\theta_t$ by minimizing a Bregman divergence that defines the Discrete Flow Matching loss.
In sum, this is to solve the discrete version of the Flow Matching problem~\eqref{prob:gen_flow}.

\subsection{Data and coupling}

Our goal is to transfer samples $X_0\sim p$ from a source PMF $p$ to samples $X_1\in q$ from a target PMF $q$, where $X_0,X_1 \in\gS$ are two RVs each taking values in the state space $\gS$.
Source and target samples can be related by means of the independent coupling $(X_0,X_1)\sim p(X_0)q(X_1)$, or associate by means of a general PMF coupling $\pi_{0,1}(x_0,x_1)$.
For example, text translation data considers coupled data $(x_0, x_1)$ representing the same document written in two different languages. 
Another application, such as text generation, concerns independent pairing where $p(x_0)$ is either the uniform probability over $\gS$ giving all states equal probability, or adding a special token $\mask$ to the vocabulary $\gT$, \ie, $\gT\cup \set{\mask}$, and considering $\pi_{0,1}(x_0,x_1) = \delta(x_0,\mask)q(x_1)$. Any RV $X_0\sim \delta(X_0,\mask)$ is the constant RV $X_0=(\mask,\ldots,\mask)$.

\subsection{Discrete probability paths}

The next step in the FM recipe is, as usual, to prescribe a probability path $p_t$ interpolating $p$ and $q$.
Following~\cref{s:general_conditioning_and_main_theorem}, we condition these objects on a general conditioning RV $Z \sim p_Z$ taking values in some arbitrary space $\gZ$. 
The \highlight{marginal probability path} takes form
\begin{equation}\label{e:p_t_discrete}
    p_t(x) = \sum_{z \in \gZ} p_{t|Z}(x|z) p_Z(z), %
\end{equation}
where $p_{t|Z}(\cdot|z)$ is a conditional PMF, and the marginal probability path satisfies the boundary constraints $p_0=p$ and $p_1=q$.

\subsection{The Marginalization Trick}

The Marginalization Trick (see \cref{s:general_conditioning_and_main_theorem}) transfers to the discrete case as-is \citep{campbell2024generative,gat2024discrete}.
Assuming that the \highlight{conditional velocity field} $u_t(\cdot,\cdot|z)$ generates $p_t(\cdot|z)$ in the sense of \eqref{def:generates_ctmc}, we obtain the \highlight{marginal velocity field}
\begin{equation}
    u_t(y,x) = \sum_{z} u_t(y,x|z) p_{Z|t}(z|x) = \E\brac{ u_t(y,X_t|Z)  \ \vert \ X_t=x },\label{e:u_t_discrete}
\end{equation}
defined for all $x,y\in \gS$ where $p_t(x)>0$, and RV $X_t\sim p_{t|Z}(\cdot|Z)$. 
By using Bayes' rule, we get
\begin{equation}
    p_{Z|t}(z|x) = \frac{p_{t|Z}(x|z)p_Z(z)}{p_t(x)}.
\end{equation}
To prove the discrete version of the Marginalization Trick Theorem (\cref{thm:fm_main}), assume: 
\begin{myframe}
\begin{assumption}\label{as:discrete_marginalization}
$p_{t|Z}(x|z) \in C^1([0,1))$, $u_t(y,x|z) \in C([0,1))$, and $p_{t}(x)>0$ for all $x\in\gS$ and $t\in [0,1)$.
\end{assumption}    
\end{myframe}
As it happened in the continuous case, the assumption $p_t>0$ is in practice mild, because we can always use $(1-(1-t)\eps)\cdot p_{Z|t} + (1-t)\eps\cdot p_{\text{uni
}}$, where $p_{\text{uni}}$ is the uniform distribution over $\gS$, and $\eps>0$ is arbitrary small.
We are no ready to state and prove the result.
\begin{myframe}
\begin{theorem}[Discrete Marginalization Trick]\label{thm:dfm_main}
    Under Assumption \ref{as:discrete_marginalization}, if $u_t(y,x|z)$ generates $p_{t|Z}(x|z)$ then the marginal velocity $u_t(y,x)$ in \eqref{e:u_t_discrete} generates $p_t(x)$ in \eqref{e:p_t_discrete} for $t\in[0,1)$.
\end{theorem}    
\end{myframe}

\begin{proof}
The proof is conceptually similar to the continuous case. Start by computing: 
    \begin{align*}
        \frac{\dd}{\dd t}p_t(y) &= \sum_{z} \frac{\dd}{\dd t}p_{t|Z}(y|z) p_Z(z) \\ &\overset{({i})}{=} \sum_{z} \brac{\sum_x u_t(y,x|z) p_{t|Z}(x|z)}p_Z(z) \\
        &\overset{({ii})}{=} \sum_{x} \brac{ \sum_{z} u_t(y,x|z)\frac{p_{t|Z}(x|z)p_Z(z)}{p_t(x)}} p_t(x) \\
        &\overset{({Bayes})}{=} \sum_{x} \overbrace{\sum_z u_t(y,x|z)p_{Z|t}(z|x)}^{{u_t(y,x)}} p_t(x),
    \end{align*}
    Equality (i) follows from \cref{thm:discrete_mass_conservation} and the fact that $u_t(y,x|z)$ generates $p_{t|Z}(y|z)$.
    Equality ({ii}) follows from multiplying and dividing by $p_t(x)$ which is assumed positive.
    Therefore, $u_t(y,x)$ satisfies the Kolmogorov Equation with $p_t$.
    Also, $u_t(y,x)$ satisfies the rate conditions \eqref{e:rate_conds}, because each $u_t(y,x|z)$ satisfies them.
    Lastly, $u_t(y,x) \in C([0,1))$ because both $u_t(y,x|z)$ and $p_{Z|t}(z|x)$ are in $C([0,1))$.
    In particular, $p_{Z|t}(z|x) \in C([0,1))$ follows from assuming  $p_t(x)>0$ for $t\in [0,1)$.
    By \cref{thm:discrete_mass_conservation}, and because $u_t(x,y)$ satisfies the Kolmogorov Equation with $p_t$ and the rate conditions, it generates $p_t$ in the sense of \eqref{def:generates_ctmc}.
\end{proof}

\subsection{Discrete Flow Matching loss}
To construct a CTMC generative model $(X_t)_{0\leq t\leq 1}$ we parameterize a velocity field $u_t^\theta(y,x)$ with parameters $\theta$, \eg, using a neural network. One would construct the neural network to satisfy the rate conditions \cref{e:rate_conds}. The \highlight{Discrete Flow Matching loss}  to train the CTMC model is defined as:
\begin{equation}\label{e:dfml}
    \gL_{\DFM}(\theta) = \E_{t,X_t\sim p_t} D_{X_t}(u_t(\cdot,X_t), u_t^\theta(\cdot,X_t)), %
\end{equation}
for $t\sim U[0,1]$ and $u_t(\cdot,x)\in \Real^{\gS}$ satisfying the rate conditions.
This means that $u_t(\cdot,x) \in \Omega_x$, where
\begin{equation}
\label{eq:dfm_omega_x}
    \Omega_x = \set{ v\in \Real^{\gS} \ \Bigg\vert \  v(y)\geq 0 \ \forall  y\ne x, \text{ and } v(x)=-\sum_{y\ne x} v(y) } \subset \Real^\gS,
\end{equation}
is a convex set, and $D_x(u,v)$ is a Bregman divergence defined using a convex function $\Phi_x:\Omega_x \too \Real$.
The \highlight{Conditional Discrete Flow Matching loss} takes form 
\begin{equation}\label{e:cdfm}
   \gL_{\CDFM}(\theta) = \E_{t,Z, X_t\sim p_{t|Z}} D_{X_t}(u_t(\cdot,X_t|Z), u_t^\theta(\cdot,X_t)). %
\end{equation}

Once again, the two losses \eqref{e:dfml} and \eqref{e:cdfm} both provide the same learning gradients.
\begin{myframe}
\begin{theorem}\label{thm:cdfm}
    The gradients of the Discrete Flow Matching loss and the Conditional Discrete Flow Matching loss coincide:
    \begin{equation}
        \nabla_\theta \gL_{\DFM}(\theta) = \nabla_\theta \gL_{\CDFM}(\theta).
    \end{equation}
    In particular, the minimizer of the Conditional Discrete Matching loss is the marginal velocity  
    \begin{equation}
        u_t^\theta(y,x) = \E\brac{ u_t(y,X_t|Z) \ \vert \ X_t=x }.
    \end{equation}
\end{theorem}
\end{myframe}
The proof follows by applying \cref{prop:bregman_gradient} when setting $X=X_t$, $Y=(X_t,Z)$, defining $f:\gS^2\too \Real^\gS$ as $(x,z)\mapsto u_t(\cdot,x|z)\in \Real^\gS$, and integrating with respect to $t\in[0,1]$.

\subsection{Factorized paths and velocities} 
\begin{wrapfigure}[16]{r}{0.3\textwidth}
  \begin{center}    \includegraphics[width=0.3\textwidth]{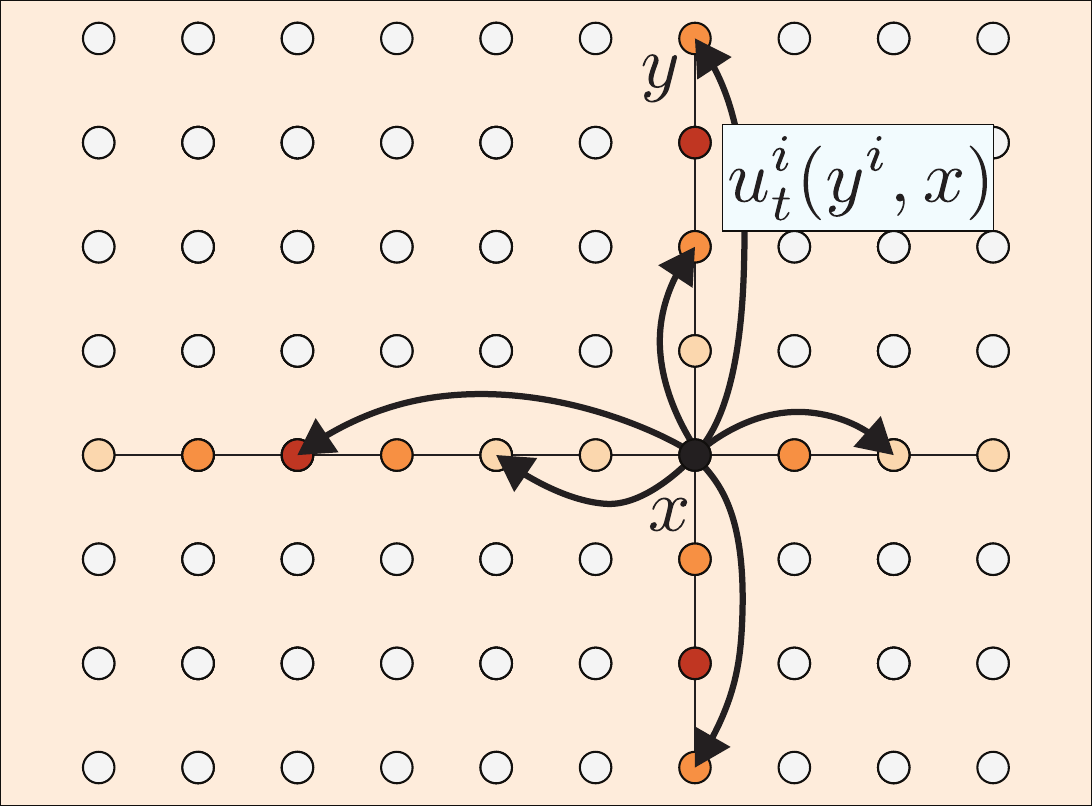}  
  \caption{Factorized CTMC model allows non-zero rates (velocities) only between states that differ in at-most one coordinate (token). }\label{fig:ctmc_factorized}
  \end{center}
\end{wrapfigure}

If implementing DFM as presented, we would require a learnable model $u_t^\theta(y, x)$---for instance, a neural network---that outputs a rate for \emph{all} possible states $y\in \gS = \gT^d$.
This would result in a huge output dimension $K^d$, infeasible for common sequence lengths $d$ and vocabulary sizes $K$.
One remedy to this issue is to consider \highlight{factorized velocities} \citep{campbell2022continuous},
\begin{equation}\label{e:factorized_velocity} 
    u_t(y,x) = \sum_i \delta(y^{\bar{i}}, x^{\bar{i}}) u_t^i(y^i,x), %
\end{equation}
where $\bar{i}=(1,\ldots,i-1,i+1,\ldots,d)$ denotes all indices excluding $i$.
Therefore, the factorized velocity above connects state $x$ to state $y$ only if these differ in at most one single token.
When using factorized velocities, we only require to model $u_t^i(y^i,x)$, as these fully define $u_t(y,x)$.
In turn, each $u_t^i(y^i,x)$ is a learnable model accepting $x\in \gS$ and returning a scalar $u_t^i(y^i,x)\in \Real$, for all $i\in [d]=\set{1,2,\ldots,d}$ and $y^i\in \gT$.
Therefore, the output of the model has a tractable dimension $d\cdot K$. The rate conditions for factorized velocities $u_t^i(y,x)$ are now required per dimension $i\in [d]$:
\begin{equation}
    u_t^i(y^i,x)\geq 0 \text{ for all } y^i\ne x^i\text{, and } \sum_{y^i\in\gT}u^i_t(y^i,x)=0\quad \text{ for all }x\in\gS.
\end{equation}

\subsubsection{Simulating CTMC with factorized velocities}\label{s:Simulating CTMC with factorized velocities}

When using factorized velocities, we can sample CTMC models coordinate-wise \citep{campbell2024generative}:
\begin{align*}
    \sP(X_{t+h}=y \ \vert \ X_t=x) &= \delta(y,x) + h\sum_{i}\delta(y^{\bar{i}},x^{\bar{i}})u_t^i(y^i,x) + o(h) \\
    &= \prod_i\brac{\delta(y^i,x^i) + hu_t^i(y^i,x) + o(h)},
\end{align*}
where the second equality follows from $\delta(y,x)=\prod_i \delta(y^i,x^i)$ and the identity
\begin{equation*}
 \prod_{i}\brac{a^i+h b^i} = \prod_{i}a^i + h\sum_i (\prod_{j\ne i}a^{j}) b^i + o(h).   
\end{equation*}
Therefore, and up to an $o(h)$ order, the transition kernel factorizes to coordinate-wise independent transitions
\begin{align}\label{e:sampling_factorized_per_coord}
    \sP(X_{t+h}^i = y^i \ \vert \ X_t=x) &= \delta(y^i,x^i) + hu^i(y^i,x) + o(h).
\end{align}
These can be sampled with the Euler method \eqref{e:ctmc_euler} per coordinate.
Interestingly, continuous Flow Matching also enjoys a similar factorization $u_t(x) = [u_t^1(x),\ldots, u_t^d(x)]\in \Real^d$, where $\dot{X}^i_t(x)=u^i_t(X_t)$ determines the change for coordinate $i$, and can be sampled independently (the ``samples'' in continuous FM are just deterministic).

\subsubsection{Building probability paths with factorized velocities}\label{s:building_paths_with_factorized_velocities}
If we construct probability paths in a certain way, it turns out to have factorized velocities (\cref{e:factorized_velocity}) by construction. We explain this construction next. For this, we define a \highlight{factorized probability path} as a probability path of the form: 
\begin{equation}\label{e:p_t_factorized}
    q_t(x) = \prod_i q^i_t(x^i). %
\end{equation}
Then, the following result shows that these factorized probability paths have factorized velocities. 
\begin{myframe}
\begin{proposition}\label{prop:factorized_path}
  Let $q_t(x)$ be a factorized probability path as in \eqref{e:p_t_factorized}, where $u_t^i(y^i,x^i) \in C([0,1))$ generates $q_t^i(x^i)$.
  Then $q_t$ has a  factorized generating velocity of the form
  \begin{equation}
      u_t(y,x)=\sum_i \delta(y^{\bar{i}}, x^{\bar{i}})  u_t^i(y^i,x^i).
  \end{equation}
\end{proposition}
\end{myframe}
To proceed with the proof, let us denote the marginal distributions of a PMF $q(x)$ by
\begin{equation}
q^i(x^i)\defe\sum_{x^{\bar{i}}}q(x)\, \quad q^{\bar{i}}(x^{\bar{i}}) \defe \sum_{x^i} q(x)
\end{equation}
\begin{proof}
    Let $q_t$ be a factorized probability path \eqref{e:p_t_factorized}.
    Let $u_t^i(y^i,x^i)$ be the generating velocity of $q_t^i(x^i)$.
    Differentiating with respect to $t$ yields
    \begin{align*}
    \frac{\dd}{\dd t} q_t(y) &= \sum_i q_t^{\bar{i}}(y^{\bar{i}}) \frac{\dd}{\dd t}q_t^i(y^i) \\
    &\overset{(i)}{=} \sum_i \brac{\sum_{x^{\bar{i}}} \delta(y^{\bar{i}}, x^{\bar{i}}) q_t^{\bar{i}}(x^{\bar{i}})} \brac{\sum_{x^i} u_t^i(y^i,x^i) q_t^i(x^i)} \\
    &\overset{(ii)}{=}   \sum_x \brac{ \sum_i \delta(y^{\bar{i}}, x^{\bar{i}})  u_t^i(y^i,x^i) } q_t(x),
\end{align*}
  Equality (i) follows from $q_t^{\bar{i}}(y^{\bar{i}})=\sum_{x^{\bar{i}}} \delta(y^{\bar{i}}, x^{\bar{i}}) q_t^{\bar{i}}(x^{\bar{i}})$ and the Kolmogorov Equation \eqref{e:kolmogorov}.
  Equality (ii) follows from changing the summation order and noting that, by the definition of $q_t$, we have $q_t^{\bar{i}}(x^{\bar{i}})q_t^i(x^i)=q_t(x)$ and $\sum_{x^{\bar{i}}}\sum_{x^i}=\sum_x$.
\end{proof}

We are now ready to show the main tool for constructing paths $p_t$ with factorized velocities that interpolate between arbitrary $p$ and $q$ \citep{campbell2024generative,gat2024discrete}.

\begin{myframe}\begin{theorem}[Discrete Factorized Marginalization Trick]\label{thm:factorized_conditional_marginals}
        Consider a marginal probability path constructed via
        \begin{equation}
    p_t(x)=\sum_z p_{t|Z}(x|z)p_Z(z), \text{ with } p_{t|Z}(x|z)=\prod_i p_{t|Z}^i(x^i|z),  
\end{equation}
\ie, where the conditional path factorizes in the sense of \cref{e:p_t_factorized}. Further, assume that $u_t^i(y^i,x^i|z)$ is $C([0,1))$ generates $p_{t|Z}^i(x^i|z)$ in $C^1([0,1))$, and $p_t(x)>0$ for all $x\in \gS$ and $t\in[0,1)$. Then, the marginal velocity is \begin{equation}\label{e:u_t_factorized}
            u_t(y,x) = \sum_i \delta(y^{\bar{i}}, x^{\bar{i}})   u_t^i(y^i,x)
\end{equation}     
with 
\begin{equation}\label{e:marginal_factorized_conditional_velocities}
u_t^i(y^i,x) =  \sum_z u_t^i(y^i,x^i|z)p_{Z|t}(z|x) =\E\brac{u_t^i(y^i,X_t^i|Z) \vert X_t=x}
\end{equation}
generates $p_t(x)$.
    \end{theorem}
\end{myframe}
\begin{proof}
According to \cref{prop:factorized_path}, the factorized conditional paths $p_{t|Z}(x|z)$ have factorized generating velocities $u_t(y,x|z) = \sum_i \delta(y^{\bar{i}}, x^{\bar{i}})  u_t^i(y^i,x^i|z)$.
Therefore, 
    \begin{align*}
        u_t(y,x) &\overset{(i)}{=} \sum_z u_t(y,x|z)p_{Z|t}(z|x) \\
        &\overset{(ii)}{=} \sum_z \brac{\sum_i \delta(y^{\bar{i}}, x^{\bar{i}}) u_t^i(y^i,x^i|z)}p_{Z|t}(x|z) \\
        &\overset{(iii)}{=} \sum_i \delta(y^{\bar{i}}, x^{\bar{i}})  \brac{ \sum_z u_t^i(y^i,x^i|z)p_{Z|t}(z|x) }. 
    \end{align*}
    Equality (i) follows from \eqref{e:u_t_discrete}.
    Equality (ii) follows from assuming that $p_{t|Z}$ has factorized velocity.
    Equality (iii) follows from changing summation order.
    Because $p^i_{t|Z}(x^i|z) \in C^1([0,1))$ and $p_t(x)>0$, it follows that $p_{t|Z}(x|z)\in C^1([0,1))$.
    Similarly, because $u_t^i(y^i,x^i|z) \in C([0,1))$, it follows that $u_t(y,x|z) \in C([0,1))$.
    Therefore, \cref{thm:dfm_main} implies that $u_t(y,x)$ generates $p_t(x)$, as required.
\end{proof}

By using \cref{thm:factorized_conditional_marginals}, we can design a probability path $p_t$ with factorized velocities interpolating between a source PMF $p$ and a target PMF $q$ as follows.
\begin{enumerate}%
    \item Find factorized probability conditional paths $p_{t|Z}(x|z)=\prod_i p^i_{t|Z}(x^i|z)$ such that the marginal $p_t(x)$ satisfies $p_0=p$ and $p_1=q$.

    \item Find generating velocities $u_t^i(y^i,x^i|z)$ to $p^i_{t|Z}(x^i|z)$. This can be done by finding solution $u_t^i(y^i,x^i|z)$ to the Kolmogorov Equation:
\begin{equation}\label{e:dfm_factorized_continuity}
    \sum_{x^i} {u_t^i(y^i,x^i|z)}p^i_{t|Z}(x^i|z) = \frac{\dd}{\dd t}p^i_{t|Z}(y^i|z), %
\end{equation}
for all $y^i\in\gT$, fixed values of $i\in[d], z\in\gZ$, and $t\in[0,1)$.
As a remark, \eqref{e:dfm_factorized_continuity} is an under-determined linear system of equations with $|\gT|$ unknowns (significantly less unknowns than the entire state space $|\gS|$). 
\end{enumerate}

\subsubsection{Conditional Discrete Flow Matching loss for factorized velocities}\label{s:conditional_loss_for_factorized_discrete}

Representing the marginal velocity $u_t^\theta$ in terms of factorized velocities $u_t^{\theta,i}$ enables the following Conditional Flow Matching loss
\begin{equation}\label{e:cdfm_factorized}
    \gL_{\CDFM}(\theta) = \E_{t,Z,X_t\sim p_{t|Z}} \sum_i D^i_{X_t}\parr{ u_t^i(\cdot,X_t|Z), u_t^{\theta,i}(\cdot,X_t) },
\end{equation}
where $t\sim U[0,1]$, and $u^i_t(\cdot,x|z),u_t^{\theta,i}(\cdot,x)\in \Real^{\gT}$ satisfy the rate conditions.
This means that $u^i_t(\cdot,x|z),u_t^{\theta,i}(\cdot,x) \in \Omega_{x^i}$ where, for $\alpha\in \gT$, we define 
\begin{equation}
    \Omega_{\alpha} = \set{ v\in \Real^{\gT} \ \Bigg\vert \  v(\beta)\geq 0\ \forall  \beta \in \gT\setminus\set{\alpha},  \text{ and } v(\alpha)=-\sum_{\beta\ne \alpha} v(\beta) }\subset \Real^\gT.
\end{equation}
This is a convex set, and $D^i_x(u,v)$ is a Bregman divergence defined by a convex function $\Phi^i_x:\Omega_{x^i} \too \Real$.
As before, we justify this loss using \cref{prop:bregman_gradient} and setting $X=X_t$, $Y=u_t^i(\cdot,X_t,Z)\in \Real^\gT$, letting $D_x^i(u,v)$ be a Bregman divergence over $\Omega_{x^i}\subset \Real^\gT$, and integrating with respect to $t\in [0,1]$. %

\subsubsection{Mixture paths}
It is time to implement \cref{s:building_paths_with_factorized_velocities} to build practical probability paths and their corresponding conditional velocities.
Following \citet{gat2024discrete}, we condition on $Z=(X_0,X_1)$ to accommodate arbitrary data couplings $(X_0,X_1)\sim \pi_{0,1}(X_0,X_1)$.
Then, we build the factorized conditional paths
\begin{equation}
p_{t|0,1}(x|x_0,x_1) = \prod_i p^i_{t|0,1}(x^i|x_0,x_1)
\end{equation}
as mixtures 
\begin{equation}\label{e:mixture_cond}
    p^i_{t|0,1}(x^i|x_0,x_1) = \kappa_t\delta(x^i,x_1^i) + (1-\kappa_t) \delta(x^i,x_0^i),
\end{equation}
where $\kappa:[0,1]\too [0,1]$ is a $C^1([0,1])$ scheduler. Note that a RV $X_t^i\sim p^i_{t|0,1}(\cdot|x_0,x_1)$ follows
\begin{equation}
    X_t^i = \begin{cases}
        x_1^i & \text{ with prob }\kappa_t \\
        x_0^i & \text{ with prob } (1-\kappa_t)
    \end{cases},
\end{equation}
i.e. it assumes either the source or the target states with a probability depending on the time $t$.

If $\kappa_0=0$ and $\kappa_1=1$, then the marginal $p_{t}(x)$ in \eqref{e:p_t_discrete} satisfies the boundary constraints. We also need generating velocities $u_t^i(y^i,x^i|x_0,x_1)$ for $p^i_{t|0,1}(x^i|x_0,x_1)$, which are solutions to \eqref{e:dfm_factorized_continuity}.
We derive these as follows: 
\begin{align*}
    \frac{\dd}{\dd t}p^i_{t|Z}(y^i|z) &\overset{\eqref{e:mixture_cond}}{=} \dot{\kappa}_t\brac{\delta(y^i,x_1^i)-\delta(y^i,x_0^i)} \\
    &\overset{\eqref{e:mixture_cond}}{=} {\dot{\kappa}_t}\brac{\delta(y^i,x_1^i) - \frac{p^i_{t|Z}(y^i|z) - \kappa_t \delta(y^i,x_1^i)}{1-\kappa_t}  } \\
    &\,\,\,\,= \frac{\dot{\kappa}_t}{1-\kappa_t}\brac{\delta(y^i,x_1^i) - p^i_{t|Z}(y^i|z)}\\
    &\,\,\,\,= \sum_{x^i}\frac{\dot{\kappa}_t}{1-\kappa_t}\brac{\delta(y^i,x_1^i) - \delta(y^i,x^i)}p^i_{t|Z}(x^i|z),
\end{align*}
where we have used $z=(x_0,x_1)$ and $Z=(X_0,X_1)$ interchangeably to keep notation concise.
In conclusion, we have found a conditional velocity generating the path in \eqref{e:mixture_cond}, namely
\begin{equation}\label{e:discrete_mixture_conditional_velocity}
    u_t^i(y^i,x^i|x_0,x_1) = \frac{\dot{\kappa}_t}{1-\kappa_t}\brac{\delta(y^i,x_1^i) - \delta(y^i,x^i)}.
\end{equation}
\Cref{ex:dfm_sample} shows how mixture paths are defined in the library \fmlibrary{}.
\begin{pbox}[label={ex:dfm_sample}]{Discrete probability path}
\begin{minted}[linenos, breaklines, mathescape, fontsize=\footnotesize, xleftmargin=2em]{python}
import torch
from flow_matching.path import MixtureDiscreteProbPath
from flow_matching.path.path_sample import DiscretePathSample

# Create a discrete probability path object
path = MixtureDiscreteProbPath(scheduler=PolynomialConvexScheduler(n=1.0)) 

# Sample the conditional path
# batch_size = 2 for $t$, $X_0$ and $X_1$
t = torch.tensor([0.25, 0.5])
x_0 = torch.tensor([0, 0])
x_1 = torch.tensor([1, 2])
sample: DiscretePathSample = path.sample(t=t, x_0=x_0, x_1=x_1) 
sample.x_0  # $X_0$ is [0, 0]
sample.x_1  # $X_1$ is [1, 2]
# $X_t$ is 
#   [0 with probability 0.75 and 1 with probability 0.25, 
#    0 with probability 0.5 and 2 with probability 0.5]
sample.x_t  
sample.t  # $t$ is [0.25, 0.5]
\end{minted}
\end{pbox}

\pagebreak
\paragraph{Velocity posterior parameterization.} Similar to the continuous case (\eg, \cref{s:velocity_param}), we can choose to parameterize our velocity $u_t^i(y^i,x)$ in different ways. The first approach is to parameterize it directly, akin to velocities in flows. Another way, which we take here, is  motivated by the following computation of the \highlight{mixture marginal velocity} following  \eqref{e:marginal_factorized_conditional_velocities}:  
\begin{align}\nonumber
    u_t^i(y^i,x) &= \sum_{x_0,x_1} \frac{\dot{\kappa}_t}{1-\kappa_t}\brac{\delta(y^i,x_1^i) - \delta(y^i,x^i)} p_{0,1|t}(x_0,x_1|x)\\ \label{e:mixture_marginal_path}
    &=  \sum_{x_1^i}\frac{\dot{\kappa}_t}{1-\kappa_t} \brac{ \delta(y^i,x_1^i) -  \delta(y^i,x^i) } p^i_{1|t} (x_1^i|x), %
\end{align}
where for the second equality we denote the marginal of the posterior $p_{0,1|t}$
\begin{align}\label{e:discrete_postreior_p_1|t}
 p_{1|t}^i(x_1^i|x) &= \sum_{x_0,x^{\bar{i}}_1}p_{0,1|t}(x_0,x_1|x) 
 \\  \label{e:cond_E_discrete_postreior_p_1|t}
 &= \E\brac{ \delta(x_1^i,X_1^i) \ \vert \ X_t=x }.
\end{align}
This derivation represents the marginal $u_t^i(y^i,x)$ using a learnable posterior $p_{1|t}^{\theta,i}(x_1^i|x)$, which can be understood as a discrete version of $x_1$-prediction (\cref{s:velocity_param}).
Next, we explore loss functions to learn this posterior.

\paragraph{CDFM losses for mixture paths}
We present two options for learning $p_{1|t}^{\theta,i}(x_1^i|x)$, both justified by \cref{prop:bregman_gradient}. 
First, the marginal posterior \eqref{e:discrete_postreior_p_1|t} and \eqref{e:cond_E_discrete_postreior_p_1|t} can be learned by the conditional matching loss
\begin{equation}
    \gL_{\CM}(\theta) = \E_{t,X_0,X_1,X_t} D_{X_t}\parr{ \delta(\cdot,X_1^i) , p_{1|t}^{\theta,i}(\cdot|X_t) }
\end{equation}
Since $\delta(\cdot,X_1^i),p_{1|t}^{\theta,i}(\cdot|X_t)$ are PMFs.
Therefore, we can set the Bregman divergence to be the KL-divergence $D(p,q)=\sum_{\alpha\in \gT} p(\alpha)\log\frac{p(\alpha)}{q(\alpha)}$ comparing PMFs, obtaining 
\begin{equation}
    \gL_{\CM}(\theta) = -\E_{t,X_0,X_1,X_t} \log p_{1|t}^{\theta,i}(X_1^i|X_t)  + \text{const}.
\end{equation}

Alternatively, we may follow \cref{s:conditional_loss_for_factorized_discrete} and use the factorized loss in \eqref{e:cdfm_factorized} with $u^{\theta,i}_{t}$ parametrized by $p_{1|t}^{\theta,i}$.
In this case, we can set the Bregman divergence to be the \emph{generalized} KL comparing general (not necessarily probability) vectors  $u,v\in \Real^m_{\geq 0}$:
\begin{equation}\label{e:generlized_kl}
    D(u,v) = \sum_{j} u^j \log \frac{u^j}{v^j} - \sum_j u_j + \sum_j v_j.
\end{equation}
For this choice of $D$, we get
\begin{align}\label{e:generlized_kl_mixture_path}
    D\parr{ u_t^i(\cdot,x^i|x_0,x_1) , u_t^{\theta,i}(\cdot,x)  } &= \frac{\dot{\kappa}_t}{1-\kappa_t}\brac{(\delta(x_1^i,x^i)-1)\log p_{1|t}^{\theta,i}(x_1^i|x) + \delta(x_1^i,x^i) -p_{1|t}^{\theta,i}(x^i|x)  }
\end{align}
which implements the loss \eqref{e:cdfm_factorized} when conditioning on $Z=(X_0,X_1)$. The generalized KL loss \eqref{e:generlized_kl_mixture_path} also provides an evidence lower bound (ELBO) on the likelihood of the target distribution \citep{shaul2024flowmatchinggeneraldiscrete},
\begin{equation}
    -\log p_1^{\theta}(x_1) \le \E_{t, X_0, X_t\sim p_{t|0,1}}\sum_{i} D\parr{ u_t^i(\cdot,X_t^i|X_0,x_1) , u_t^{\theta,i}(\cdot,X_t)  },
\end{equation}
where $p_1^{\theta}$ is the marginal generated by the model at time $t=1$. Hence, in addition to training, the generalized KL loss is commonly used for evaluation.

\paragraph{Sampling mixture paths}

The parametrization based on the posterior $p_{1|t}^{\theta,i}$ leads to the following sampling algorithm.
As indicated in \cref{s:Simulating CTMC with factorized velocities} working with factorized velocities enables the coordinate-wise sampling \eqref{e:sampling_factorized_per_coord}.
According to \eqref{e:marginal_factorized_conditional_velocities} and \eqref{e:mixture_marginal_path},
\begin{align}
    \sP(X_{t+h}^i = y^i \ \vert \ X_t=x) &= \delta(y^i,x^i) + hu^i(y^i,x) + o(h)      \\
    &= \sum_{x_1^i} \brac{ \delta(y^i,x^i) + h\frac{\dot{\kappa}_t}{1-\kappa_t}\brac{\delta(y^i,x_1^i) - \delta(y^i,x^i)} + o(h) }p^i_{1|t}(x_1^i|x).
\end{align}
Consequently, and given $X_t=x$, we may perform one step of sampling by performing the next two steps for each $i\in[d]$: (i) draw  $X_1^i \sim p_{1|t}^i(X_1^i|x)$; and (ii) update $X_{t+h}^i$ according to the Euler step in \eqref{e:ctmc_euler} with the velocity $\frac{\dot{\kappa}_t}{1-\kappa_t}\brac{\delta(y^i,X_1^i) - \delta(y^i,x^i)}$. Intuitively, (ii) decides whether to set $X_{t+h}^i=X_1^i$ or remain at $X_{t+h}^i = X_t^i$.

\paragraph{One-sided mixture paths and probability preserving velocities} 
It is often useful to extend the design space of the sampling algorithm by adding some divergence-free component, as described in \cref{s:probability_preserving}. For factorized paths, the divergence-free velocity $v_t^i$ needs to satisfy \eqref{e:dfm_factorized_continuity}, namely,   
\begin{equation}
    \sum_{x^i} v_t^i(y^i,x^i|z) p_{t|Z}^i(x^i|z) = 0.
\end{equation}
In general it could challenging to find such probability-preserving velocities without learning additional quantities, \eg, $p_{0|t}^i$. However, one useful case where a probability preserving velocity can be found in closed form is when assuming iid source distribution \ie, $p(x)=\prod_i p(x^i)$ and independent coupling $\pi_{0,1}(x_0,x_1)=p(x_0)q(x_1)$. In this case the  marginal mixture path takes the form
\begin{align*}
    p_t(x) &= \sum_{x_1} p_{t|1}(x|x_1) q(x_1), \text{ where } p_{t|1}(x|x_1)=\prod_i p_{t|1}^i(x^i|x_1), 
\end{align*}
where $p_{t|1}^i(x^i|x_1)=\brac{\kappa_t \delta(x^i,x_1^i) + (1-\kappa_t)p(x^i)}$.
The conditional velocity in \eqref{e:discrete_mixture_conditional_velocity}, \ie, 
\begin{equation}
    u_t^i(y^i,x^i|x_1) = \frac{\dot{\kappa}_t}{1-\kappa_t}\brac{\delta(y^i,x_1^i) - \delta(y^i,x^i)}
\end{equation}
also generates $p_{t|1}^i(x^i|x_1)$.
To find a divergence-free velocity we can, for example, subtract from this velocity a \emph{backward-time velocity} $\tilde{u}_t^i$ for $p_{t|1}^i(y^i,x^i|x_1)$ \citep{gat2024discrete}, in the sense that it satisfies the Kolmogorov equation with $p_{t|1}^i(y^i,x^i|x_1)$ and $-\tilde{u}_t^i$ satisfies the rate conditions.  Such a velocity can be found in a fashion to \cref{e:discrete_mixture_conditional_velocity},
\begin{equation}
    \tilde{u}_t^i(y^i,x^i|x_1)=\frac{\dot{\kappa}_t}{\kappa_t}\brac{\delta(y^i,x^i)-p(x^i)}.
\end{equation}
Therefore, a divergence-free velocity for $p_{t|1}^i(x^i|x_1)$ conditional path can be defined via 
\begin{equation}
    v_t^i(y^i,x^i|x_1) = u_t^i(y^i,x^i|x_1) - \tilde{u}_t^i(y^i,x^i|x_1).
\end{equation}

According to \cref{s:probability_preserving}, if we add a divergence-free field $v^i_t(y^i,x^i|x_1)$  to the velocity $u_t^i(y^i,x^i|x_1)$,  the latter still generates the same probability path $p_{t|1}^i(x^i|x_1)$. Consequently, \cref{thm:factorized_conditional_marginals} implies that the marginal velocity  $u_t^i(y^i,x)$ defined by
\begin{align*}
    u_t^i(y^i,x) &= \sum_{x_1} \brac{u_t^i(y^i,x^i|x_1) + c_t v_t^i(y^i,x^i|x_1)}p_{1|t}(x_1|x) \\
    &= \sum_{x_1^i}\brac{u_t^i(y^i,x^i|x^i_1) + c_t v_t^i(y^i,x^i|x^i_1)}p_{1|t}^i(x^i|x),
\end{align*}
still generates the same marginal path $p_t(x)$, where the second equality follows from $u_t^i(y^i,x^i|x_1)=u_t^i(y^i,x^i|x^i_1)$ for mixture paths, and similarly for $v_t^i(y^i,x^i|x^i_1)$.
In conclusion, and given $X_t=x$, a single step of the generalized sampling algorithm consists in (i) drawing $X_1^i \sim p_{1|t}^i(X_1^i|x)$ and (ii) taking an Euler step  \eqref{e:ctmc_euler} with the velocity 
\begin{equation*}
    u_t^i(y^i,x^i|x_1)=\frac{\dot{\kappa}_t}{1-\kappa_t}\brac{\delta(y^i,X_1^i) - \delta(y^i,x^i)} + c_t\brac{\frac{\dot{\kappa}_t}{1-\kappa_t}\brac{\delta(y^i,x_1^i) - \delta(y^i,x^i)} - \frac{\dot{\kappa}_t}{\kappa_t}\brac{\delta(y^i,x^i)-p(x^i)} },
\end{equation*}  
where $c_t>0$ is a time dependent constant.

Similar to the continuous flow matching example in~\cref{ex:fm_standalone}, we provide a standalone implementation of discrete flow matching in pure PyTorch in~\cref{ex:fm_discrete_standalone}. \Cref{ex:dfm_train} illustrates how to train a discrete flow with arbitrary data coupling using the \fmlibrary{}.

\begin{pbox}[label={ex:dfm_train}]{Training and sampling DFM with mixture paths and arbitrary data coupling.}
\begin{minted}[linenos, breaklines, mathescape, fontsize=\footnotesize, xleftmargin=2em]{python}
import torch

from flow_matching.path import MixtureDiscreteProbPath, DiscretePathSample
from flow_matching.path.scheduler import PolynomialConvexScheduler
from flow_matching.loss import MixturePathGeneralizedKL
from flow_matching.solver import MixtureDiscreteEulerSolver
from flow_matching.utils import ModelWrapper

model = ...  # Define a trainable velocity model
optimizer = torch.optim.Adam(model.parameters())

scheduler = PolynomialConvexScheduler(n=1.0)
path = MixtureDiscreteProbPath(scheduler=scheduler)
loss_fn = MixturePathGeneralizedKL(path=path)  # Generalized KL Bregman divergence

for x_0, x_1 in dataloader:  # Samples from $\pi_{0,1}$ of shape [batch_size, *data_dim]
    t = torch.rand(batch_size) * (1.0 - 1e-3)  # Randomize time $t \sim U[0,1-10^{-3}]$
    sample: DiscretePathSample = path.sample(t=t, x_0=x_0, x_1=x_1)  # Sample the conditional path 
    model_output = model(sample.x_t, sample.t)

    loss = loss_fn(logits=model_output, x_1=sample.x_1, x_t=sample.x_t, t=sample.t) # CDFM loss
    
    optimizer.zero_grad()  
    loss.backward()
    optimizer.step()

class ProbabilityDenoiser(ModelWrapper):
    def forward(self, x: torch.Tensor, t: torch.Tensor, **extras) -> torch.Tensor:
        logits = self.model(x, t, **extras)
        return torch.nn.functional.softmax(logits.float(), dim=-1)

# Sample $X_1$
probability_denoiser = ProbabilityDenoiser(model=model)
x_0 = torch.randint(size=[batch_size, *data_dim])  # Specify the initial condition
solver = MixtureDiscreteEulerSolver(
    model=probability_denoiser, 
    path=path, 
    vocabulary_size=vocabulary_size
)

step_size = 1 / 100
x_1 = solver.sample(x_init=x_0, step_size=step_size, time_grid=torch.tensor([0.0, 1.0-1e-3]))
\end{minted}
\end{pbox}

\begin{pbox}[label={ex:fm_discrete_standalone}]{Standalone Discrete Flow Matching code \\ \url{flow_matching/examples/standalone_discrete_flow_matching.ipynb}}
\begin{minted}{python}
import torch
import matplotlib.pyplot as plt
from torch import nn, Tensor
from sklearn.datasets import make_moons

class DiscreteFlow(nn.Module):
    def __init__(self, dim: int = 2, h: int = 128, v: int = 128):
        super().__init__()
        self.v = v
        self.embed = nn.Embedding(v, h)
        self.net = nn.Sequential(
            nn.Linear(dim * h + 1, h), nn.ELU(),
            nn.Linear(h, h), nn.ELU(),
            nn.Linear(h, h), nn.ELU(),
            nn.Linear(h, dim * v))
    
    def forward(self, x_t: Tensor, t: Tensor) -> Tensor:
        return self.net(torch.cat((t[:, None], self.embed(x_t).flatten(1, 2)), -1)).reshape(list(x_t.shape) + [self.v])

batch_size = 256
vocab_size = 128

model = DiscreteFlow(v=vocab_size)
optim = torch.optim.Adam(model.parameters(), lr=0.001) 

for _ in range(10000):
    x_1 = Tensor(make_moons(batch_size, noise=0.05)[0])
    x_1 = torch.round(torch.clip(x_1 * 35 + 50, min=0.0, max=vocab_size - 1)).long()
    x_0 = torch.randint(low=0, high=vocab_size, size=(batch_size, 2))

    t = torch.rand(batch_size)
    x_t = torch.where(torch.rand(batch_size, 2) <  t[:, None], x_1, x_0)

    logits = model(x_t, t)
    loss = nn.functional.cross_entropy(logits.flatten(0, 1), x_1.flatten(0, 1)).mean()
    optim.zero_grad()
    loss.backward()
    optim.step()

x_t = torch.randint(low=0, high=vocab_size, size=(200, 2))
t = 0.0
results = [(x_t, t)]
while t < 1.0 - 1e-3:
    p1 = torch.softmax(model(x_t, torch.ones(200) * t), dim=-1)
    h = min(0.1, 1.0 - t)
    one_hot_x_t = nn.functional.one_hot(x_t, vocab_size).float()
    u = (p1 - one_hot_x_t) / (1.0 - t)
    x_t = torch.distributions.Categorical(probs=one_hot_x_t + h * u).sample()
    t += h
    results.append((x_t, t))
fig, axes = plt.subplots(1, len(results), figsize=(15, 2), sharex=True, sharey=True)
for (x_t, t), ax in zip(results, axes):
    ax.scatter(x_t.detach()[:, 0], x_t.detach()[:, 1], s=10)
    ax.set_title(f't={t:.1f}')
plt.tight_layout()
plt.show()
\end{minted}
\vspace{0.5cm}
\includegraphics[width=\textwidth]{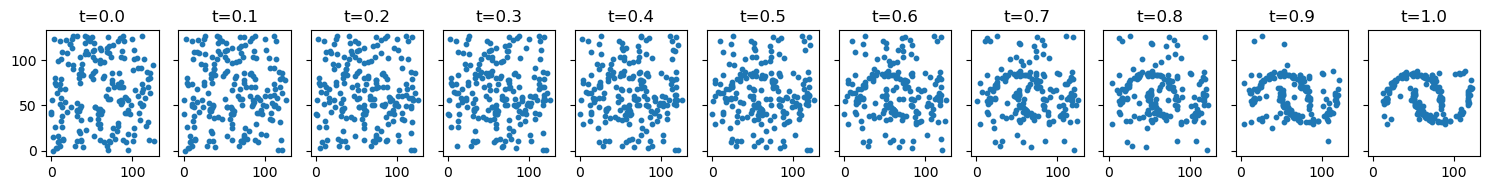}
\end{pbox}

\pagebreak
\section{Continuous Time Markov Process Models}\label{sec:fellerflow}
In the previous sections, we have developed a flow model on $\mathbb{R}^d$ and Riemannian manifolds (see \cref{s:flow_models} and \cref{s:fm_non_euclidean}) and a CTMC model for discrete data (see \cref{sec:ctmcm}). In this section, we want to unify and extend these models to a generative model that works for (1) general state spaces and (2) general Markov processes. This generative model will allow us to extend the principles of Flow Matching to a wide variety of generative models for a variety of modalities in \cref{sec:generatormatching}.

\subsection{General state spaces and random variables}

\paragraph{Working with general modalities.} Our explicit goal is not specify the modality we use. Hence, throughout this section, let $\gS$ be a general \highlight{state space}. Important examples  are $\gS=\mathbb{R}^d$ (\eg, images, vectors), $\gS$ discrete (\eg,  language), $\gS$ a Riemannian manifold (\eg, geometric data) or their products for generation of multiple data modalities jointly (multimodal models). For all modalities, we can define a metric (or distance function) $d:\gS\times\gS\to\mathbb{R}_{\geq 0},(x,y)\mapsto d(x,y)$ on $\gS$. For example, for $\gS$ discrete, the metric is simply $d(x,y)=1$ if $y\neq x$ and $d(x,x)=0$ for all $x\in\gS$. For $\gS=\mathbb{R}^d$, we use $d(x,y)=\|x-y\|$. We need to make a technical assumption that $(\gS,d)$ is a Polish metric space, \ie, it is complete (\ie, any Cauchy sequence converges) and separable (\ie, it has a countable dense subset). Any modality of interest for machine learning has that property.

\paragraph{Densities over general state spaces.} So far, in this work we assumed that a probability distribution $p$ over $\gS$ is represented by a density $p:\gS\to\mathbb{R}_{\geq 0}$. For general state spaces, we use a general \highlight{reference measure} $\nu$ and the density becomes the Radon-Nikodym derivative $\frac{\dd p}{\dd \nu}$. In other words, probabilities can be expressed as integrals with respect to $\nu$
\begin{align*}
    \sP(A) &= \int\limits_{A}p(x)\nu(\dd x)\text{ for all measurable }A\subset \gS
\end{align*}
For $\gS$ discrete, $\nu$ was the counting measure (so the integrals are just sums) and $p(x)$ is just the probability mass function (PMF). For $\gS=\mathbb{R}^d$, $\nu$ was the Lebesgue measure (so the integrals are just the ``usual'' integral) and $p(x)$ is just the probability density function (PDF). The above generalizes that to arbitrary state spaces.

\paragraph{(Optional) Working with arbitrary distributions.} It is important to note that not every probability distribution admits a density with respect to a reference measure. For the reader unfamiliar with general measure theory, it is safe ignore this possibility as a technical remark as long one works with distributions of interest that have a density $p(x)$. However, note that these are not just pathological examples but there are real cases of interest for machine learning applications where this matters: A simple  example is a probability path of the form $p_t=\delta_{(1-t)x+ty}$ on $\gS=\mathbb{R}^d$ connecting two points $x,y\in\mathbb{R}^d$ in a straight line - this cannot be represented by a density. Another example would be probability distributions over $\gS=C([0,1],\mathbb{R})$, \eg, for trajectory modeling, that often do not have a density with respect to a common reference measure. To mathematically handle such cases, we develop our framework for general \highlight{probability measures} $p$ over $\gS$. For this, we use the notation $p(\dd x)$ where ``$\dd x$'' is a \emph{symbolic} expression denoting integration with respect to $p$ in a variable $x$. For example, for a bounded measurable function $f:\gS\to\mathbb{R}$ we write 
\begin{align*}
    \mathbb{E}_{X\sim p}[f(X)] = \int f(x)p(\dd x)
\end{align*}
for the Lebesgue integral or the expected value of $f$ under $p$. As before, we use (with a slight abuse of notation) the same notation $p(x)$ to denote the density of the measure $p(\dd x)$.

\begin{table}[t]
\centering
\begin{adjustbox}{width=\textwidth}
\renewcommand{\arraystretch}{1.5}
\begin{tabular}{lcccc}
\toprule
Name & \highlight{Flow} & \highlight{Diffusion} &
\highlight{Jump process} &  \begin{tabular}{@{}c@{}}\highlight{Continous-time} \\ \highlight{Markov chain}
\end{tabular}
\\
\toprule
Space $\gS$ & $\gS=\mathbb{R}^d$ &  $\gS=\mathbb{R}^d$ & $\gS$ arbitrary & $|\gS|<\infty$
\\
\midrule
Parameters & \begin{tabular}{@{}c@{}}
Velocity field: \\
$u_t(x)\in\mathbb{R}^d$
\end{tabular}
& 
\begin{tabular}{@{}c@{}}
Diffusion coefficient:\\
$\sigma_t(x)\in\mathbb{R}^{d\times d}$ (p.s.d.)
\end{tabular}
& 
\begin{tabular}{@{}c@{}} 
Jump measure $Q_t(\dd y,x)$\\
\end{tabular} 
& 
\begin{tabular}{@{}c@{}}$u_t\in\mathbb{R}^{\gS \times \gS}$, $1^Tu_t=0$\\
$u_{t}(x';x)\geq 0$ ($x'\neq x$)
\end{tabular}
\\
\midrule
\begin{tabular}{@{}c@{}}
Sampling 
\end{tabular}
& $X_{t+h} = X_{t}+hu_t(X_t)$
& 
\begin{tabular}{@{}c@{}}
$X_{t+h} = X_{t}+\sqrt{h}\sigma_t(X_t)\epsilon_t$\\
$\epsilon_t\sim\mathcal{N}(0,I)$
\end{tabular} & \begin{tabular}{@{}c@{}}
    $X_{t+h} = X_{t}$ with prob. $1-h\int Q_t(\dd y,x)$\\
    $X_{t+h} \sim \frac{Q_t(\dd y,x)}{\int Q_t(\dd y,x)}$ with prob. $h\int Q_t(\dd y,x)$
\end{tabular}
& 
$X_{t+h} \sim(I+h u_{t})(\cdot;X_t)$ 
\\
\midrule
Generator $\mathcal{L}_t$& $\nabla f^Tu_t$ & $\frac{1}{2}\nabla^2 f \cdot \sigma_t^2$ & 
$\int [f(y)-f(x)]Q_t(\dd y,x)$
& $f^Tu_t^T$
\\
\midrule
\begin{tabular}{@{}c@{}}
KFE \\
(Adjoint)
\end{tabular}
&
\begin{tabular}{@{}c@{}}
Continuity Equation: \\
$\partial_tp_t=-\divv(p_tu_t)$
\end{tabular} 
&
\begin{tabular}{@{}c@{}}
Fokker-Planck Equation: \\
$\partial_tp_t=\frac{1}{2}\nabla^2\cdot(p_t\sigma^2_t)$
\end{tabular} 
& 
\begin{tabular}{@{}c@{}}
Jump Continuity Equation:
$\partial_tp_t(x) =$\\
$\int Q_t(x,y)p_t(y)-Q_t(y,x)p_t(x)\nu(\dd y)$
\end{tabular}
&
\begin{tabular}{@{}c@{}}
Mass preservation:\\
$
\partial_tp_{t}
=u_tp_t$
\end{tabular}
\\
\midrule
\begin{tabular}{@{}c@{}}
Marginal
\end{tabular} & $\mathbb{E}_{Z\sim p_{Z|t}(\cdot|x)}[u_t(x|Z)]$ & $\mathbb{E}_{Z\sim p_{Z|t}(\cdot|x)}[\sigma_t^2(x|Z)]$
&
$\mathbb{E}_{Z\sim p_{Z|t}(\cdot|x)}[Q_t(y,x|Z)]$
& $\mathbb{E}_{Z\sim p_{Z|t}(\cdot|x)}[u_t(y;x|Z)]$
\\
\midrule
\begin{tabular}{@{}c@{}}
CGM Loss \\
(Example)
\end{tabular}
& $\|u_t(X|Z)-u_t^\theta(X)\|^2$
& $\|\sigma_t^2(X|Z)-(\sigma_t^{\theta})^2(X)\|_{2}^2$
& \begin{tabular}{@{}c@{}}
$(\int Q_{t}^\theta(y;X)\nu(\dd y)$\\
$-Q_t(y,X|Z)\log Q_{t}^\theta(y;X)\nu(\dd y))$  
\end{tabular}
& 
\begin{tabular}{@{}c@{}}
$(\sum\limits_{y\neq x}u_{t}^\theta(y;X)$\\
$-u_t(y;X|Z)\log u_{t}^\theta(y;X))$
\end{tabular}
\\
\bottomrule
\end{tabular}
\end{adjustbox}
\caption{\label{table:Markov_overview} Some examples of CTMP generative models and how they can be learnt with Generator Matching. This list is not exhaustive. Derivations are in \cref{sec:fellerflow}. For diffusion, we assume zero drift as this is covered by the ``Flow'' column. KFE is listed in its adjoint version, \ie, assumes jump kernel $Q_t(y,x)$ and density $p_t(x)$ exists with respect to reference measure $\nu$. ``p.s.d.'': Positive semi-definite.}
\end{table}

\subsection{The CTMP generative model}

Similarly, our explicit goal is build a model that works for arbitrary evolution process - regardless of whether we use a flow, diffusion, a CTMC, a combination, or something else. Therefore, we define a evolution process in this section that is general but satisfies the necessary regularity assumptions to build a generative model. For $t\in [0,1]$, let $X_t\in \gS$ be a random variable. We call $(X_t)_{0\leq t\leq 1}$ a \highlight{Continuous-time Markov process (CTMP)} if it fulfills the following condition:
\begin{myframe}
\begin{equation}
   \mathbb{P}[X_{t_{n+1}}\in A|X_{t_1},X_{t_2},\dots, X_{t_n}]= \mathbb{P}[X_{t_{n+1}}\in A|X_{t_n}]
   \quad (0\leq t_1<\dots<t_{n+1}\leq 1, A\subseteq \gS)
\end{equation}
\end{myframe}
Informally, the above condition says that the process has no memory. If we know the present, knowing the past will not influence our prediction of the future.
In \cref{table:Markov_overview}, we give an overview over important classes of Markov processes. For example, a flow on a manifold is a Markov process with deterministic transitions, a diffusion is a Markov process with transitions driven by Brownian motion, and CTMCs are Markov processes determined by rates (we will explain this in detail in \cref{subsub:example_generators}). Each Markov process has a \highlight{transition kernel} $(\tkernel_{t+h|t})_{0\leq t<t+h\leq 1}$ that assigns every $x\in \gS$ a probability distribution $\tkernel_{t+h|t}(\cdot|x)$
such that 
\begin{align}
\mathbb{P}[X_{t+h}\in A|X_{t}=x]=\tkernel_{t+h|t}(A|x)\quad \text{ for all }t,h\geq 0,A\subset \gS\text{ measurable}
\end{align}
Due to the Markov assumption, a Markov process is uniquely determined by the transition kernel and the distribution of $X_0$. Conversely, any transition kernel and initial distribution defines a Markov process. Therefore, there is a 1:1 correspondence.

Our next goal is to define a corresponding  generalization of a velocity field for CTMPs. Informally, it would be the 1st-order derivative of the transition kernel in $t$:
\begin{equation}\label{e:taylor_informal}
\mathcal{L}_t\defe\frac{\dd}{\dd h}\Big\vert_{h=0}\tkernel_{t+h|t}
\end{equation}
We call the 1st-order derivative $\mathcal{L}_t$ the \highlight{generator} of $\tkernel_{t+h|t}$ \citep{ethier2009markov, ruschendorf2016comparison}. Similar to derivatives, generators are first-order \emph{linear} approximations and easier to parameterize than $\tkernel_{t+h|t}$. As we will see, diffusion, flows, and other generative models can all be seen as algorithms to learn the generator of a Markov process (see \cref{table:Markov_overview}). This leads to the general form of the \highlight{CTMP generative model} given by
\begin{myframe}
\begin{equation}
    \text{\highlight{CTMP model (informal):}}\quad p_{t+h|t}(\cdot|x) \defe \delta_{x} + h \mathcal{L}_t(x) + o(h), \text{ and } 
    X_0\sim p.\label{e:ctmp_model}
    \end{equation}
\end{myframe}
However, as a transition kernel $\tkernel_{t+h|t}$ is not a real function, equation \ref{e:taylor_informal} and \ref{e:ctmp_model} are only heuristic and not well-defined yet. Therefore the first goal of this section is to provide a \emph{formal} definition of the generator and the CTMP generative model.

\subsubsection{Formal definition of generator}
\label{subsec:formal_definition_generator}
The first problem in \cref{e:taylor_informal} is that derivatives are usually defined with respect to functions mapping to vector spaces but $p_{t+h|t}$ maps to a distribution. However, this can be alleviated by using \highlight{test functions}.  Test functions are a way to ``probe'' a probability distribution. They serve as a theoretical tool to handle distributions as if they were real-valued functions.  Specifically, a set of test functions is a family $\mathcal{T}$ of bounded, measurable functions $f:S\to\mathbb{R}$ that \emph{characterize} probability distributions fully, \ie, for two probability distributions $\mu_1,\mu_2$ on $\gS$ it holds 
\begin{align}
\label{eq:test_function_requirement}
\mu_1=\mu_2\quad \Leftrightarrow \quad \mathbb{E}_{X\sim \mu_1}[f(X)]=\mathbb{E}_{X\sim \mu_2}[f(X)]\quad\text{for all }f\in\mathcal{T}
\end{align}
Generally speaking, one chooses $\mathcal{T}$ to be as ``nice'' (or regular) as possible. For example, if $\gS=\mathbb{R}^d$, the space $\mathcal{T}=C_c^{\infty}(\Real^d)$ of infinitely differentiable functions with compact support fulfills that property. For $\gS$ discrete, $\mathcal{T}=\mathbb{R}^{\gS}$ simply consists of all functions (which are just vectors in this case). Let $X_t\sim p_t$. We define the \highlight{marginal action} and \highlight{transition action} as
\begin{align}
\label{eq:marginal_action}
    \action{p_t}{f} &\defe \int f(x)p_t(\dd x) = \E_{X\sim p_t}\brac{f(X)}\\ %
    \action{\tkernel_{t+h|t}}{f}(x) &\defe\ip{\tkernel_{t+h|t}(\cdot|x),f}= \E\brac{f(X_{t+h})|X_t=x} %
\end{align}
where the marginal action maps each test function $f$ to a scalar $\ip{p_t,f}\in\mathbb{R}$, while the transition action maps a real-valued function $x\mapsto f(x)$ to a another real-valued function $x\mapsto \ip{\tkernel_{t+h|t},f}(x)$. The tower property implies that $\ip{p_t,\ip{\tkernel_{t+h|t},f}}=\ip{p_{t+h},f}$. We note that the above is only a ``symbolic'' dot product but becomes a ``proper'' dot product if a density $p_t(x)$ exists, \ie, $\ip{p_t,f}=\int f(x)p_t(x)\nu(\dd x)$.

The second step to formally define a derivative as in \cref{e:taylor_informal} is that we need to impose some notion of ``smoothness'' on a Markov process that we define now. Let $C_0(\gS)$ be the space of continuous functions $f:\gS\to\mathbb{R}$ that vanish at infinity, \ie, for all $\epsilon>0$ there exists a compact set $K\subset \gS$ such that $|f(x)|<\epsilon$ for all $x\in\gS\setminus K$. We use the supremum norm $\|\cdot\|_{\infty}$ on $C_0(\gS)$. A CTMP $X_t$ is called a \highlight{Feller process} if it fulfils the following two conditions \citep{feller1955second,ruschendorf2016comparison}:
\begin{enumerate}
    \item \highlight{Strong continuity: }The action of $\tkernel_{t+h|t}$ is continuous in time:
\begin{align*}
\lim\limits_{h'\to h,t'\to t}\|\action{p_{t'+h'|t'}}{f}-\action{\tkernel_{t+h|t}}{f}\|_{\infty} =0\quad \text{ for all }h,t\geq 0, f\in C_0(\gS)
\end{align*}
\item \highlight{No return from infinity: }The action of $\tkernel_{t+h|t}$ preserves functions that vanish at infinity:
\begin{align*}
\action{\tkernel_{t+h|t}}{f}\in C_0(\gS)\quad\text{for all }h,t\geq 0, f\in C_0(\gS)
\end{align*}
\end{enumerate}
\begin{myframe}
\begin{assumption}\label{as:feller}
The CTMP $(X_t)_{0\leq t\leq 1}$ is a Feller process.
\end{assumption}    
\end{myframe}
This is a reasonable assumption given that we want to use $X_t$ in a machine learning model: We define probability paths where the distribution of the generative process $X_t$ vary smoothly, and all our data usually lies in some bounded (compact) set. 

Let us now revisit \eqref{e:taylor_informal} and try define the derivative of $\tkernel_{t+h|t}$. With the test function perspective in mind, we can take derivatives of $\action{\tkernel_{t+h|t}}{f}(x)$ per $x\in \gS$ and define
\begin{align}\label{e:generator}
    \frac{\dd}{\dd h}\Big\vert_{h=0}\action{\tkernel_{t+h|t}}{f}(x)
    &= \lim_{h\too 0} \frac{\action{\tkernel_{t+h|t}}{f}(x) - f(x)}{h} \defe [\gL_t f](x).
\end{align}
We call this action the \highlight{generator} $\gL_t$ and define it for all $f$ for which the above limit exists uniformly, \ie, in the norm $\|\cdot\|_{\infty}$. Intuitively, a generator is defined as an \emph{operator of test functions}. In \cref{table:Markov_overview}, there are several examples of generators that we derive in \cref{subsub:example_generators}. There is a 1:1 correspondence between a generator and a Feller process 
\citep{rogers2000diffusions, ethier2009markov, pazy2012semigroups} - in the same way as there is a correspondence between a flow and a vector field (see \cref{thm:ode_existence_and_uniqueness}). This will later allows us to parameterize a Feller process via a generator in a neural network.

\pagebreak
With this definition, the \highlight{CTMP model} in \eqref{e:ctmp_model} has the, now well-defined, form as 
\begin{myframe}
\begin{equation}
\label{eq:ctmp_model_formal}
\action{\tkernel_{t+h|t}}{f} = f + h \gL_t f + o(h)\text{ (for all }f\in\mathcal{T})\quad \text{ and } 
    X_0\sim p
\end{equation}
\end{myframe}
where $o(h)$ describes an error terms $E(h)\in C_0(\gS)$ such that $\lim\limits_{h\to 0}\frac{1}{h}\|E(h)\|_\infty=0$.  Similarly to \cref{def:generates} for the case of flows and to \cref{def:generates_ctmc} for the case of CTMCs, we say that
\par %
\begin{myframe}
\begin{equation}\label{def:generates_ctmp}
    \mathcal{L}_t \text{ \highlight{generates} } p_t \text{ if there exists a } \tkernel_{t+h|t} \text{ satisfying \eqref{eq:ctmp_model_formal}}
 \text{ with CTMP $X_t$ such that }X_t\sim p_t
\end{equation}
\end{myframe}
In other words, a generator $\mathcal{L}_t$ generates the probability path $p_t$ if a Markov process that is (1) initialized with $p=p_0$ and (2) simulated with $\mathcal{L}_t$ has marginals $p_t$.

\subsubsection{Examples of CTMP models}
\label{subsub:example_generators}
We go through several examples to illustrate how to compute a generator of a Markov process. The results from this section are summarized in \cref{table:Markov_overview}.

\paragraph{Flows.} Let $\gS=\mathbb{R}^d$ and $u:[0,1]\times \mathbb{R}^d\to \mathbb{R}^d, (t,x)\mapsto u_t(x)$ be a time-dependent velocity field defining a flow $\psi_t$ (see \cref{s:flow_models}). Let $\mathcal{T}=C_{c}^\infty(\mathbb{R}^d)$ be the space of infinitely differentiable and smooth functions with compact support. Then we can compute the generator via
\begin{align}
    [\gL_t f](x) 
    &=\lim_{h\too 0} \frac{\E\brac{f(X_{t+h})|X_t=x}-f(x)}{h}\\
    &\overset{(i)}{=}\lim_{h\too 0} \frac{\E\brac{f(X_t+hu_t(X_t)+o(h))|X_t=x}-f(x)}{h}\\
    &\overset{(ii)}{=}\lim_{h\too 0} \frac{\E\brac{f(X_t)+h\nabla f(X_t)^Tu_t(X_t)+o(h)|X_t=x}-f(x)}{h}\\
    &= \lim_{h\too 0} \frac{f(x)+h\nabla f(x)^T u_t(x) + o(h)-f(x)}{h}\\
    \label{eq:generator_flow}\\
    &= \nabla f(x)^T u_t(x),
\end{align}
where (i) follows from a Euler approximation of the flow and (ii) follows from a first-order Taylor approximation of $f$ around $X_t$. Therefore, the \highlight{flow generator} is given by
\begin{myframe}
\begin{equation}
\label{eq:flow_generator}
  \mathcal{L}_t f(x) =\nabla f(x)^T u_t(x).
\end{equation}
\end{myframe}

\paragraph{Diffusion.} Let $\gS=\mathbb{R}^d$ and $\sigma_t:[0,1]\times \mathbb{R}^d\to \mathbb{R}^{d\times d}, (t,x)\mapsto \sigma_t(x)$  be a time-dependent function mapping to symmetric positive semi-definite matrices $\sigma_t$ in a continuous fashion. A diffusion process with diffusion coefficient $\sigma_t$ is defined via the SDE $dX_t=\sigma_t(X_t)dW_t$ for a Wiener process $W_t$ \citep{oksendal2003stochastic}. This process can be approximated via the infinitesimal sampling procedure: 
\begin{align}
    X_{t+h} = X_{t} + \sqrt{h }\sigma_t(X_t)\epsilon_t,\quad \epsilon_t\sim\mathcal{N}(0,I)
\end{align}
Let again be $\mathcal{T}=C_{c}^\infty(\mathbb{R}^d)$. Then we can compute the generator via
\begin{align}
    [\gL_t f](x) &=\lim_{h\too 0} \frac{\E\brac{f(X_{t} + \sqrt{h} \sigma_t(X_t)\epsilon_t+o(h))|X_t=x}-f(x)}{h}\\
    &\overset{(i)}{=}\lim_{h\too 0} \frac{\mathbb{E}_{\epsilon_t}[f(x)+\nabla f(x)^T\sqrt{h}\sigma_t(x)\epsilon_t+\frac{1}{2}h[\sigma_t(x)\epsilon_t]^T\nabla^2f(x)[\sigma_t(x)\epsilon_t]+o(h)-f(x)]}{h}\\
    &=\lim_{h\too 0} \frac{\nabla f(x)^T\sqrt{h}\sigma_t(x)\mathbb{E}_{\epsilon_t}[\epsilon_t]+\mathbb{E}_{\epsilon_t}[\frac{1}{2}h[\sigma_t(x)\epsilon_t]^T\nabla^2f(x)[\sigma_t(x)\epsilon_t]]}{h}\\
    &=\frac{1}{2}\mathbb{E}_{\epsilon_t}[\epsilon_t^T[\sigma_t(x)]^T\nabla^2f(x)[\sigma_t(x)]\epsilon_t]]\\
    &\overset{(ii)}{=}\frac{1}{2}\text{tr}\left(\sigma_t(x)^T\nabla^2f(x)\sigma_t(x)\right)\\
    &\overset{(iii)}{=}\frac{1}{2}\text{tr}(\sigma_t(x)\sigma_t(x)^T\nabla^2f(x))\\
    &\overset{(iv)}{=}\frac{1}{2}\sigma_t^2(x)\cdot \nabla^2f(x)
\end{align}
where in (i) we use a 2nd order Taylor approximation (2nd order because $\mathbb{E}[\|\sqrt{h}\epsilon_t\|^2]\propto h$), in (ii) the identity $\text{tr}(A)=\mathbb{E}_{\epsilon_t}[\epsilon_t^TA\epsilon_t]$ for $A\in\mathbb{R}^{d\times d}$, in (iii) the cyclic property of the trace, and in (iv) the symmetry of $\sigma_t$. Further, we use $A\cdot B\defe\text{tr}(A^TB)$ to denote the matrix inner product for matrices $A,B\in\mathbb{R}^{d\times d}$. Therefore, the \highlight{diffusion generator} is given by
\begin{myframe}
\begin{equation}
\label{eq:diffusion_generator}
  \mathcal{L}_t f(x) =\frac{1}{2}\sigma_t^2(x)\cdot \nabla^2f(x).
\end{equation}
\end{myframe}

\paragraph{Jumps.} Next, let $\gS$ be arbitrary and let us consider a \highlight{jump process}. A jump process is defined by a time-dependent kernel $Q_t(\dd y,x)$, \ie, for every $0\leq t \leq 1$ and every $x\in \gS$, $Q_t(\dd y,x)$ is a positive measure over $S\setminus\{x\}$. The idea of a jump process is that the total volume assigned to $S\setminus\{x\}$
\begin{align}
    \lambda_t(x) = \int Q_t(\dd y,x)
\end{align}
gives the \highlight{jump intensity}, \ie, the infinitesimal likelihood of jumping. Further, if $\lambda_t(x)>0$, we can assign a \highlight{jump distribution} by normalizing $Q_t$ to a probability kernel
\begin{align}
\label{eq:jump_kernel_decomposition}
    J_t(\dd y,x) = \frac{Q_t(\dd y,x)}{\lambda_t(x)}.
\end{align}
A jump process can be approximated via the infinitesimal sampling procedure as follows:
\begin{align}
    X_{t+h} = 
\begin{cases}
        X_t & \text{with probability }1-h\lambda_t(X_t)+o(h) \\
        \sim J_t(\dd y,X_t) & \text{with probability }h\lambda_t(X_t)+o(h)
\end{cases}
\end{align}
For a rigorous treatment of jump processes, see for example \citep{davis1984piecewise}. The generator is then given by
\begin{align}
\mathcal{L}_tf(x)&=\lim\limits_{h\to 0}\frac{\mathbb{E}[f(X_{t+h})-f(X_t)|X_t=x]}{h}\\
&=\lim\limits_{h\to 0}\frac{\mathbb{E}[f(X_{t+h})-f(X_t)|X_t=x,\text{Jump in }[t,t+h)]\mathbb{P}[\text{Jump in }[t,t+h)|X_t=x]}{h}\\
&+\lim\limits_{h\to 0}\underbrace{\frac{\mathbb{E}[f(X_{t+h})-f(X_t)|X_t=x,\text{No jump in }[t,t+h)]\mathbb{P}[\text{No jump in }[t,t+h)|X_t=x]}{h}}_{=0}\\
&=\lim\limits_{h\to 0}\frac{\mathbb{E}_{Y\sim J_t(\dd y,x)}\left[f(Y)-f(x)\right]h\lambda_t(x)}{h}\\
&=\mathbb{E}_{Y\sim J_t(\dd y,x)}\left[f(Y)-f(x)\right]\lambda_t(x)\\
&=\int (f(y)-f(x))Q_t(\dd y,x)
\end{align}
where we have used that if $X_t$ does not jump in $[t,t+h]$, then $X_{t+h}=X_t$. Therefore, the \highlight{jump generator} is given by
\begin{myframe}
\begin{equation}
\label{eq:jump_generator}
  \mathcal{L}_t f(x) =\int (f(y)-f(x))Q_t(\dd y,x)=\lambda_t(x)\mathbb{E}_{Y\sim J_t(\dd y,x)}[f(Y)-f(x)].
\end{equation}
\end{myframe}

\paragraph{Continuous-time Markov chain (CTMC).} Let us consider a continuous-time Markov chain $X_t$ on a discrete space $\gS$ with $|\gS|<\infty$.  In fact, this is simply a jump process on a discrete state space with a specific parameterization.  To see this, consider a vanilla jump kernel on a discrete state space $\gS$ given by a matrix $Q_t\in \mathbb{R}^{\gS\times \gS}_{\geq 0}$ and using \cref{eq:jump_generator}, the generator is given by 
\begin{align}
\label{eq:generator_discrete_jump}
    \mathcal{L}_t f(x)&=\sum\limits_{y\in \gS}[f(y)-f(x)]Q_t(y,x)=\sum\limits_{y\neq x}[f(y)-f(x)]Q_t(y,x)\quad \text{for all }x\in\gS, f\in\mathbb{R}^{\gS}
\end{align}
\ie, the value of $Q_t(x,x)$ does not matter and is underdetermined. Therefore, a natural convention is to reparameterize the jump kernel on discrete state spaces by rates:  
\begin{align*}
    u_t(y,x)=\begin{cases} Q_t(y,x) & \text{ if }y\neq x\\
    -\sum\limits_{z\neq x}Q_t(z;x) & \text{ if }y=x
\end{cases}
\end{align*}
With this, we recover the rates $u_t(y,x)$ from \cref{sec:ctmcm} fulfilling the rate conditions in \ref{e:rate_conds} by construction. Therefore, this shows that a jump model on a discrete space coincides with the CTMC model (\cref{sec:ctmcm}). Applying this on \cref{eq:generator_discrete_jump},  we get that the \highlight{CTMC generator} is given by
\begin{myframe}
\begin{equation}
\label{eq:ctmc_generator}
  \mathcal{L}_t f(x) =\sum\limits_{y\in \gS}f(y)u_t(y,x)=f^Tu_t
\end{equation}
\end{myframe}
where we consider $f=(f(x))_{x\in \gS}$ as a column vector and $u_t\in\mathbb{R}^{\gS \times \gS}$ as a matrix. Therefore, the generator function is simply vector multiplication from the left.

\paragraph{Flows on manifolds.} Next, we consider flows on Riemannian manifolds $\gS=\gM$ as in \cref{s:fm_non_euclidean}. A flow $\psi:[0,1]\times \gM\too\gM$ is defined via a vector field $u:[0,1]\times \mathcal{M}\to T\mathcal{M}$ via the ODE in \eqref{e:flow}. 
Let us denote the transition from time $s$ to $t$ via $\psi_{t|s}(x)=\psi_t(\psi_s^{-1}(x))$ (as in \eqref{e:flow_is_markov}). Then, 
for a smooth function $f:\mathcal{M}\to\mathbb{R}$ we have that the \highlight{Riemannian flow generator} is given via
\begin{myframe}
\begin{align}
\label{eq:generator_manifold_flows}
    \mathcal{L}_tf(x) = \lim\limits_{h\to 0}\frac{f(\psi_{t+h|t}(x))-f(x)}{h}=\ip{\nabla f(x), \frac{\dd}{\dd  h}\Big\vert_{h=0}\psi_{t+h|t}(x)}_{g}=\ip{\nabla f(x), u_t(x)}_{g}
\end{align}
\end{myframe}
where $\ip{\cdot,\cdot}_{g}$ describes the dot product defining the Riemannian metric $g$ and $\nabla f$ describes the gradient of $f$ with respect to $g$. In fact, the generator  coincides with the \emph{Lie derivative} of a function \citep{jost2008riemannian}, a fundamental concept in differential geometry.

\subsection{Probability paths and Kolmogorov Equation}

For Flow Matching on $\gS=\mathbb{R}^d$, the Continuity Equation (see \ref{e:continuity}) is the central mathematical equation that allows us to construct velocity fields that generate a desired probability path (see \cref{ss:continuity_equation}). In this section, we derive a corresponding - more general - equation for CTMPs. Let $X_t$ be a CTMP with generator $\gL_t$ and let $X_t\sim p_t$, then we know that:
\begin{align*}
    \frac{\dd}{\dd t}\action{p_t}{f}
    &=\frac{\dd}{\dd h}\Big\vert_{h=0}\action{p_{t+h}}{f}
        =\frac{\dd}{\dd h}\Big\vert_{h=0}\action{p_t}{\action{\tkernel_{t+h|t}}{f}}
    =\action{p_t}{\frac{\dd}{\dd h}\Big\vert_{h=0}\action{\tkernel_{t+h|t}}{f}}
= \action{p_t}{\gL_t f}
\end{align*}
where we used that the $\action{p_t}{\cdot}$ operation is linear to swap the derivative, and that by the tower property $\action{p_{t}}{\action{\tkernel_{t+h|t}}{ f}}=\action{p_{t+h}}{f}$. This shows that given a generator $\gL_t$ of a Markov process $X_t$ we can recover its marginal probabilities via their infinitesimal change, \ie, we arrive at the \highlight{Kolmogorov Forward Equation (KFE)}

\begin{myframe}
\begin{equation}
\label{eq:weak_kfe}
\frac{\dd}{\dd t}\action{p_t}{f}= \action{p_t}{\gL_t f}\text{ for all }f\in\mathcal{T}
\end{equation}
\end{myframe}
The version of the KFE in \cref{eq:weak_kfe} determines the evolution of expectations of test functions $f$. This is necessary if we use probability distributions that do not have a density. If a density exists, a more familiar version of the KFE can be used that directly prescribes the change of the probability densities. To present it, we introduce the \highlight{adjoint generator} $\gL_t^*$, which acts on probability densities $p_t(x)$ with respect to a reference measure $\nu$, namely $\gL_t^*p_t(x)$ is (implicitly) defined by the identity
\begin{align}\label{e:adjoint}
\int p_t(x)\mathcal{L}_tf(x)\nu(\dd x)&=\int \mathcal{L}_t^*p_t(x)f(x)\nu(\dd x)
 \quad &&\forall f\in \gT
\end{align}
Further, we need to assume that $p_t$ is differentiable in $t$. Now, \eqref{e:adjoint} applied to the KFE \eqref{eq:weak_kfe} we get
\begin{align}
\int \frac{\dd}{\dd t} p_t(x) f(x)\nu(\dd x)&=\frac{\dd}{\dd t} \int p_t(x)f(x)\nu(\dd x)\\
&=\frac{\dd}{\dd t}\action{p_t}{f}\\
&=\action{p_t}{\mathcal{L}_tf}\\
&=\int p_t(x)\mathcal{L}_tf(x)\nu(\dd x)\\
&=\int \mathcal{L}_t^*p_t(x)f(x)\nu(\dd x)
\end{align}
As this holds for all test functions $f$, we can conclude using \cref{eq:test_function_requirement} that this is equivalent to the \highlight{adjoint KFE}

\begin{myframe}
\begin{equation}\label{e:adjoint_kfe}
    \frac{\dd}{\dd t} p_t(x) = \gL_t^* p_t(x)\text{ 
 for all }x\in \gS %
\end{equation}
\end{myframe}
As we will derive in the following examples,  the adjoint KFE generalizes many famous equations used to develop generative models such as the Continuity Equation or the Fokker-Planck Equation \citep{song2021sde, lipman2022flow} (see \cref{table:Markov_overview}). Whenever a probability density exists, we use the adjoint KFE - to avoid using test functions and work with probability densities directly. We summarize our findings in the following
\begin{myframe}
    \begin{theorem}[General Mass Conservation]\label{thm:general_mass_conservation}
    Let $\mathcal{L}_t$ be a generator of $(X_t)_{0\leq t\leq 1}$. Informally, the following conditions are equivalent:
    \begin{enumerate}
        \item $p_t,\mathcal{L}_t$ satisfies the KFE \eqref{eq:weak_kfe}.
        \item $\frac{\dd p_t}{\dd \nu}(x),\mathcal{L}_t$ satisfy the adjoint KFE \eqref{e:adjoint_kfe}.
        \item $\mathcal{L}_t$ generates $p_t$ in the sense of equation \eqref{def:generates_ctmp}.
    \end{enumerate}
    Formally, (1) and (2) are equivalent whenever $\frac{\dd p_t}{\dd \nu}$ exists and is continuously differentiable in $t$. Further, (3) implies (1) for arbitrary state spaces. There are weak regularity assumptions that ensure that (1) implies (3) (see \cref{appendix:regularity_assumptions_list} for a list). In this work, we assume that these hold, \ie, \textbf{we assume that (3) implies (1)}.
    \end{theorem}
\end{myframe}
To the best of our knowledge, there is no known result for abstract general state spaces that ensures that in \cref{thm:general_mass_conservation} condition (3) implies (1). This is why we simply assume it here. For the machine learning researcher, this assumption holds for any state space of interest and should therefore be of no concern (see \cref{appendix:regularity_assumptions_list}). 

\pagebreak
\subsubsection{Examples of KFEs}
\label{ss:kfe_examples}

\paragraph{Adjoint KFE for Flows.} Let us set $\gS=\mathbb{R}^d$ and assume that $p_t$ has a density $p_t(x)$ with respect to the Lebesgue measure that is bounded and continuously differentiable. Then we can compute the adjoint generator $\mathcal{L}_t^*$ via
\begin{align}
\action{p_t}{\mathcal{L}_t f}
=&\mathbb{E}_{x\sim p_t}[\mathcal{L}_t f(x)]\\
=&\int \mathcal{L}_t f(x) p_t(x)\dd x\\
\overset{(i)}{=}&\int \nabla f(x)^T u_t(x) p_t(x)\dd x\\
\overset{(ii)}{=}&\int f(x) \underbrace{[-\divv(p_tu_t)(x)]}_{=:\mathcal{L}_t^*p_t(x)}\dd x\\
=&\int f(x) \gL_t^*p_t(x)\dd x
\end{align}
where (i) follows by \cref{eq:generator_flow} and (ii) by integration by parts. The above derivation shows that the adjoint generator is given by $\gL_t^*p_t=-\divv(p_tu_t)(x)$ (because it fulfils the condition in \cref{e:adjoint}). Using the adjoint KFE, we recover the \highlight{Continuity Equation} (see \cref{e:continuity})
\begin{myframe}
\begin{equation}
\frac{\dd}{\dd t} p_t(x) = -\divv(p_t u_t) (x),
\end{equation}
\end{myframe}
an equation that we extensively studied  in \cref{ss:flows_and_velocities}.

\paragraph{Adjoint for diffusion. }Let's set $\gS=\mathbb{R}^d$  and assume that $p_t$ has a density $p_t(x)$ with respect to the Lebesgue measure that is bounded and continuously differentiable. We can compute the adjoint generator $\mathcal{L}_t^*$ via
\begin{align}
\action{p_t}{\mathcal{L}_t f}
=&\mathbb{E}_{x\sim p_t}[\mathcal{L}_t f(x)]\\
=&\int \mathcal{L}_t f(x) p_t(x)\dd x\\
\overset{(i)}{=}&\frac{1}{2}\int \sigma_t^2(x)\cdot\nabla^2f(x) p_t(x)\dd x\\
\overset{(ii)}{=}&\int f(x)\underbrace{\frac{1}{2}\nabla^2\cdot(p_t\sigma_t^2)(x)}_{=:\mathcal{L}_t^*p_t(x)}\dd x\\
=&\int f(x)\mathcal{L}_t^*p_t(x)\dd x
\end{align}
by (i) follows by \cref{eq:diffusion_generator} and (ii) follows by applying integration by parts twice. The above derivation shows that the adjoint generator is given by $\gL_t^*p_t=\frac{1}{2}\nabla^2\cdot(p_t\sigma_t^2)(x)$ (because it fulfils the condition in \cref{e:adjoint}). The adjoint KFE then recovers the well-known \highlight{Fokker-Planck equation}
\begin{myframe}
\begin{equation}\label{e:fokker_planck}
\frac{\dd}{\dd t} p_t(x) =\frac{1}{2}\nabla^2\cdot(p_t\sigma_t^2)(x)
\end{equation}
\end{myframe}

\paragraph{Adjoint KFE for jumps.} Let's assume that $p_t$ has a density $p_t(x)$ with respect to the Lebesgue measure that is bounded and continuously differentiable. Let's assume that the jump measures $Q_t(\dd y,x)$ is given via a kernel $Q_t:\gS\times \gS\to\mathbb{R}_{\geq 0}, (y,x)\mapsto Q_t(y,x)$ such that 
\begin{align}
    \int f(y)Q_t(\dd y,x) = \int f(y) Q_t(y,x)\nu(\dd y)\text{ for all integrable }f:\gS\to\mathbb{R}.
\end{align}
Then we can derive the adjoint generator as follows:
\begin{align}
    \action{p_t}{\mathcal{L}_tf}(x) =&\int \int (f(y)-f(x))Q_t(y,x)\nu(\dd y)p_t(x)\nu(\dd x)\\
    =&\int \int f(y)Q_t(y,x)p_t(x)\nu(\dd y)\nu(\dd x)-\int \int f(x)Q_t(y,x)p_t(x)\nu(\dd y)\nu(\dd x)\\
\overset{(i)}{=}&\int \int f(x)Q_t(x,y)p_t(y)\nu(\dd y)\nu(\dd x)-\int \int f(x)Q_t(y,x)p_t(x)\nu(\dd y)\nu(\dd x)\\
=&\int f(x)\underbrace{\left[\int Q_t(x,y)p_t(y)-Q_t(y,x)p_t(x)\nu(\dd y)\right]}_{=:\mathcal{L}_t^*p_t}\nu(\dd x)\\
=&\int f(x)\mathcal{L}_t^*p_t(x)\nu(\dd x)
\end{align}
where in $(i)$ we simply swap the variables $x$ and $y$. The above derivation shows that $\mathcal{L}_t^*$ as defined above fulfils the condition in \cref{e:adjoint} and indeed describes the adjoint generator for jumps. With this, the adjoint KFE becomes the \highlight{Jump Continuity equation}:

\begin{myframe}
\begin{equation}
\label{eq:jump_continuity}
    \frac{\dd}{\dd t} p_t(x)=\int \brac{Q_t(x,y)p_t(y)-Q_t(y,x)p_t(x)}\nu(\dd y)=\int \lambda_t(y)J_t(x,y)p_t(y)\nu(\dd y)-\lambda_t(x)p_t(x)
\end{equation}
\end{myframe}
where we use the decomposition $Q_t(y,x)=\lambda_t(x)J_t(y,x)$ into a jump intensity $\lambda_t$ and a jump distribution $J_t$ (see \cref{eq:jump_kernel_decomposition}).

\paragraph{Adjoint KFE for CTMCs.} For $\gS$ discrete and generator given by $f^Tu_t$ as in \cref{eq:ctmc_generator}, we get that
\begin{align*}
\action{p_t}{\mathcal{L}_tf}
=\int p_t(x)\mathcal{L}_tf(x)\nu(\dd x)
&=\sum\limits_{x\in \gS}p_t(x)\sum\limits_{y\in \gS}u_t(y,x)f(y)\\
&=\sum\limits_{y\in \gS}\underbrace{[\sum\limits_{x\in \gS}p_t(x)u_t(y,x)]}_{=:\mathcal{L}_t^*p_t(x)}f(y)\\
&=\int \mathcal{L}_t^*p_t(y)f(y)\nu(\dd y)
\end{align*}
where $\nu$ here just denotes the counting measure. Therefore, the adjoint is KFE simply given by 

\begin{myframe}
\begin{equation}
   \frac{\dd}{\dd t} p_t(x) = \sum\limits_{y\in \gS}u_t(x,y)p_t(y)
\end{equation}
\end{myframe}
This recovers the KFE for CTMCs as derived in \cref{e:kolmogorov} (with $x$ and $y$ switched to keep consistency with the derivations in this section). 

\pagebreak
\subsection{Universal representation theorem}
Generators allow us to characterize the space of possible Markov processes. Specifically, the following result allows us to not only characterize a wide class of CTMP generative models but to characterize the design space \emph{exhaustively} for $\gS=\mathbb{R}^d$ or $\gS$ discrete.
\begin{theorem}[Universal characterization of generators] 
\label{theorem:representation} Under weak regularity assumptions, the generators of a Feller processes $X_t$ ($0\leq t\leq 1$) take the form:
\begin{enumerate}
\vspace{-0.5em}
\item \textbf{Discrete $|\gS|<\infty$:} The generator is given by a rate transition matrix $u_t$ and the Markov process corresponds to a continuous-time Markov chain (CTMC). 
\vspace{-0.5em}
\item \textbf{Euclidean space  $\gS=\mathbb{R}^d$:} The generator has a representation as a sum of components described in \cref{table:Markov_overview}, \ie,
\begin{align}
\label{eq:universal_representation}
    \mathcal{L}_tf(x) = \underbrace{\vphantom{\frac{1}{2}}\nabla f(x)^T u_t(x)}_{\text{flow}}+\underbrace{\frac{1}{2}\nabla^2 f(x) \cdot \sigma^2_t(x)}_{\text{diffusion}}+\underbrace{\int \brac{f(y)-f(x)}Q_t(\dd y,x)}_{\text{jump}}
\end{align}
where $u:[0,1]\times\Real^d\too\Real^d$ is a velocity field, $\sigma:[0,1]\times\Real^d\too S^{++}_d$ the diffusion coefficient ($S^{++}_d$ denotes the positive semi-definite matrices), and $Q_t(dy;x)$ a jump measure; $\nabla^2 f(x)$ describes the Hessian of $f$ and $\nabla^2 f(x) \cdot \sigma^2_t(x)$ describes the Frobenius inner product.
\end{enumerate}
\end{theorem}
The proof adapts a known result in the mathematical literature \citep{courrege1965forme, von1965fast} and can be found in \citep{holderrieth2024gm}.

\newpage
\section{Generator Matching}\label{sec:generatormatching}
In this section, we describe \emph{Generator Matching} (GM) \citep{holderrieth2024gm}, a generative modeling framework for (1) arbitrary data modalities and (2) general Markov processes. GM unifies the vast majority of  generative models developed in recent years, including diffusion models, ``discrete diffusion'' models, and the FM variants described in previous sections. To introduce GM, we defined the CTMP generative model in \cref{sec:fellerflow} that is constructed via a \emph{generator} of a Markov process. GM describes a scalable algorithm to train generators - giving the method its name. Beyond providing a unifying framework, GM gives rise to a variety of new models, allows us to combine models of different classes, as well as allows to build models for arbitrary modalities including models across multiple data modalities.

\subsection{Data and coupling}
As before, our goal is to transfer samples $X_0\sim p$ from a distribution $p$ to samples $X_1\sim q$ from a target distribution $q$, where $X_0,X_1\in\gS$ are two RVs each taking values in the state space $\gS$.
Source and target samples can be related by means of the independent coupling $(X_0,X_1)\sim p\otimes q$ (product distribution), or associated by means of a general PMF coupling $\pi_{0,1}$, \ie, distribution over $\gS\times\gS$ with marginal $\pi_{0}=p$ and $\pi_{1}=q$.
The only difference to before is that $\gS$ is a general state space and that $p,q$ can be arbitrary probability measures.

\subsection{General probability paths}

The next step in the GM recipe is, as before, to prescribe a probability path $p_t$ interpolating $p$ and $q$.
Following~\cref{s:general_conditioning_and_main_theorem}, we use a \highlight{conditional probability path} $p_{t|Z}(\dd x|z)$, \ie, a set of time-varying probability measures dependent on a latent state $z\in\gZ$. Given a distribution $p_{Z}$ over $\gZ$, we consider the corresponding \highlight{marginal probability path} $p_t(\dd x)$ defined via the hierarchical sampling procedure:
\begin{align*}
Z\sim p_{Z}, X_t\sim p_{t|Z}(\dd x|z) \quad \Rightarrow \quad X_t\sim p_t(\dd x)
\end{align*}
\ie, we obtain a sample from $p_t$ by first sampling $Z$ from $p_{Z}$ and then sampling $X_t$ from $p_{t|Z}(\dd x|z)$. As before, the marginal probability path is constructed to satisfy the boundary constraints $p_0=p$ and $p_1=q$.

We have seen already two common constructions for $\gZ = \gS$ and $p_{Z}=q$: First, affine conditional flows for $\gS=\mathbb{R}^d$ (as used in continuous FM; \cref{s:fm_continuous}) defined via
\begin{align}
Z\sim q, X_0\sim p, X_t=\sigma_t X_0+\alpha_t Z\quad \Rightarrow \quad X_t\sim p_t(\dd x)
\end{align}
where $\alpha_t,\sigma_t\in \mathbb{R}_{\geq 0}$ are differentiable functions satisfying $\alpha_0=\sigma_1=0$ and $\alpha_1=\sigma_0=1$. Second, for arbitrary $\gS$, we can use mixtures as used in discrete FM for discrete state spaces (\cref{e:mixture_cond}):
\begin{align}
\label{eq:general_definition_mixture}
Z\sim q, X_0\sim p, X_t \sim \begin{cases}
        Z & \text{with prob } \kappa_t\\
        X_0 & \text{with prob } (1-\kappa_t)
    \end{cases}\quad \Rightarrow \quad X_t\sim p_t(\dd x)
\end{align}
where $\kappa_t\in \mathbb{R}_{\geq 0}$ is a differentiable functions satisfying $\kappa_0=0$ and $\kappa_1=1$ and $0\leq \kappa_t\leq 1$. One can easily see that the affine conditional and mixture probability paths interpolate $p$ and $q$, \ie, $p_0=p$ and $p_1=q$.

\subsection{Parameterizing a generator via a neural network}
Given a probability path $p_t$, our goal is construct a CTMP model specified by a generator $\mathcal{L}_t$ that generates this probability path (see \cref{def:generates_ctmp}). To train a neural network for that, we first need to explain how to parameterize a generator $\mathcal{L}_t$ with a neural network $\mathcal{L}_t^\theta$ with parameters $\theta$. We will do this in this section.

Let $\gT$ again be a family of test functions (see \cref{subsec:formal_definition_generator}). A \highlight{linear parameterization} of $\mathcal{L}_t$ is defined as follows: for every $x\in\gS$ there is (1) a convex closed set $\Omega_{x}\subset V_{x}$ that is a subset of a vector space $V_{x}$ with an inner product $\ip{\cdot,\cdot}_{x}$ and (2) a linear operator $\mathcal{K}:\mathcal{T}\to C(\gS;V_x)$ such that every considered generator $\mathcal{L}_t$ can be written as 
\begin{myframe}
\begin{equation}
\label{eq:linear_parameterization}
\mathcal{L}_tf(x)=\ip{\mathcal{K}f(x),F_t(x)}_{x}
\end{equation}
\end{myframe}
for a function $F_t$ such that $F_t(x)\in\Omega_{x}$ for every $x\in\gS$. Crucially, the operator $\mathcal{K}$ cannot depend on $\mathcal{L}_t$, \ie, only $F_t$ has to be learned. This leads to the

\begin{myframe}
\begin{equation}
\label{def:generator_neural_network_parameterization}
\text{\highlight{Parameterized generator}: }
\mathcal{L}_t^\theta f(x)=\ip{\mathcal{K}f(x),F_t^\theta(x)}_{x}\text{ with neural network }F_t^\theta\text{ and parameters }\theta,
\end{equation}
\end{myframe} 
where again $F_t^\theta$ maps an element $x\in\gS$ to $F_t^\theta(x)\in\Omega_{x}$. We list several examples to make this definition more concrete.

\paragraph{Linear parameterization of flows.} Let $\gS=\mathbb{R}^d$ and $\Omega_{x}=\mathbb{R}^d=V_{x}$. Let's consider all flows, \ie, the family of generators is given by (see \cref{eq:generator_flow}):
\begin{align}
    \mathcal{L}_tf=\nabla f^Tu_t,\quad u_t:\mathbb{R}^d\to\mathbb{R}^d.
\end{align}
Setting $\mathcal{K}f=\nabla f$ and $F_t=u_t$ we recover the shape of \cref{eq:linear_parameterization}. This gives a natural linear parameterization of flow generators via their vector fields.

\paragraph{Linear parameterization of diffusion.} Let $\gS=\mathbb{R}^d$ and $\Omega_{x}=S_d^{++}\subset \mathbb{R}^{d\times d}=V_{x}$, where $S_d^{++}$ denotes the set of all positive semi-definite matrices. Then a diffusion generator is given by (see \cref{eq:diffusion_generator}):
\begin{align}
\label{e:linear_parameterization_diffusion}
    \mathcal{L}_tf=\nabla^2 f\cdot \sigma_t^2,\quad \sigma_t:\mathbb{R}^d\to S_d^{++}
\end{align}
Setting $\mathcal{K}f=\nabla^2 f$ and $F_t=\sigma_t^2$ we recover the shape of \cref{eq:linear_parameterization}. This gives a natural linear parameterization of diffusion generators.

\paragraph{Linear parameterization of jumps.} Let $\Omega_{x}=\{a:\gS\setminus\{x\}\to \mathbb{R}_{\geq 0}\, |\,  a\text{ integrable}\}\subset L^2(\gS\setminus\{x\})=V_{x}$ with dot product $\ip{a,b}_{x}=\int\limits_{\gS\setminus\{x\}}a(x)b(x)\nu(dx)$. Then the jump generator is given by (see \cref{eq:jump_generator}): 
\begin{align}
\label{e:linear_parameterization_jump}
    \mathcal{L}_tf(x)=\int [f(y)-f(x)] Q_t(y,x)\nu(\dd y)
=\ip{\mathcal{K}f(x),Q_t(\cdot;x)}_{x}
\end{align}
where we set $\mathcal{K}f(x)$ as the function $y\mapsto f(y)-f(x)$. Setting $F_t=Q_t$ we recover the shape of \cref{eq:linear_parameterization} - giving a linear parameterization of jump generators. We note that the above only parameterizes jumps with a jump kernel $Q_t(y,x)$, which does not necessarily include all jump measures.

\paragraph{Linear parameterization of CTMCs.}Let $\gS$ be discrete and $u_t\in \mathbb{R}^{\gS \times \gS}$ be a rate matrix of a continuous-time Markov chain. As for discrete FM (see \cref{eq:dfm_omega_x}), we define
\begin{equation}
    \Omega_x = \set{ v\in \Real^{\gS} \ \Bigg\vert \  v(y)\geq 0 \ \forall  y\ne x, \text{ and } v(x)=-\sum_{y\ne x} v(y) } \subset V_{x}=\Real^\gS.
\end{equation}
Then by \cref{eq:ctmc_generator}, the generator is given for $f\in \mathbb{R}^{\gS}$ by
\begin{align}
\mathcal{L}_tf(x)=f^Tu_t(\cdot,x)=\ip{f,u_t(\cdot,x)}_{x}
\end{align}
where $V_{x}=\mathbb{R}^{S}$ and $\mathcal{K}f=f$ and $\ip{\cdot,\cdot}_{x}$ is the standard Euclidean dot product. With this, we recover the shape of \cref{eq:linear_parameterization}. Therefore, this gives a natural linear parameterization of CTMCs via their rates $u_t$.

\paragraph{Linear parameterization of flows on manifolds.} Let $\gS=\gM$ be a Riemannian manifold and as in \cref{s:fm_non_euclidean}, let us consider flows on Riemannian manifolds. By \cref{eq:generator_manifold_flows}, the generator is given by
\begin{align}
\label{eq:linear_param_generator_manifold}
\mathcal{L}_tf(x)=\ip{\nabla f(x), u_t(x)}_{g}
\end{align}
with $u_t$ being a time-dependent smooth vector field $u_t :[0,1]\times \gM\too T\gM$, and $u_t(x)\in T_x\gM$ for all $x\in\gM$. Setting $\Omega_{x}=V_{x}=T_x\gM$ and $\mathcal{K}=\nabla f$ the gradient operator,  we recover the shape of \cref{eq:linear_parameterization}. Therefore, this gives a natural linear parameterization of Riemannian flow generators.

\subsection{Marginal and conditional generators}
In this section, we show how to find generators for marginal probability paths. The recipe is as follows: We can find generators for conditional probability paths $p_{t|Z}(\dd x|z)$, often analytically, and use these to construct generators for the marginal path. Specifically, let us assume that for every $z\in \gZ$ we found a (conditional) generator $\gL^z$ that generates $p_{t|Z}(\dd x|z)$, \ie, by \cref{thm:general_mass_conservation} this equivalent to the KFE (\cref{eq:weak_kfe}):
\begin{align}
\frac{\dd}{\dd t} \action{p_{t|Z}(\cdot|z)}{f} = \action{p_{t|Z}(\cdot|z)}{\gL_t^z f}\quad \text{ for all }f\in\mathcal{T}.
\end{align}
Further, let us assume that we found a linear parameterization (see \cref{eq:linear_parameterization}) as follows: 
\begin{align}
\label{eq:cond_linear_parameterization}
    \mathcal{L}_t^z f(x) = 
\ip{\mathcal{K}f(x),F_t(x|z)}_{x}\quad z\in \gZ
\end{align}
for functions $F_t(x|z)\in \Omega_{x}\subset V_{x}$. For example, $F_t(x|z)$ could be conditional velocity field in continuous FM (see \cref{sec:Deriving generating velocity fields}) or the conditional rates in discrete FM (see \cref{e:u_t_discrete}). This allows us to find a formula for a generator that generates the marginal path:
\begin{myframe}
\begin{theorem}[General Marginalization Trick]
\label{thm:marginal_generator}
The marginal probability path $(p_t)_{0\leq t \leq 1}$ is generated by a Markov process $X_t$ with generator
\begin{align}
\label{eq:marginal_generator_equation}
\mathcal{L}_tf(x) &=\mathbb{E}_{Z\sim p_{Z|t}(\cdot|x)}[\mathcal{L}_t^Zf(x)]
\end{align}
where $p_{Z|t}(\dd z|x)$ is the posterior distribution (\ie, the conditional distribution of $z$ given  $x$). The generator $\mathcal{L}_t$ has a linear parameterization given by
\begin{align}
\label{eq:marginal_generator_equation_linear_parameterization}
F_t(x)=\mathbb{E}_{Z\sim p_{Z|t}(\cdot|x)}[F_t(x|Z)].
\end{align}
\end{theorem}    
\end{myframe}
The above theorem gives us our training target: to approximate $\mathcal{L}_t$ in \cref{eq:marginal_generator_equation} with a neural network. The marginalization tricks seen in previous chapters (\cref{thm:fm_main}, \cref{thm:rfm_main}, \cref{thm:dfm_main}) are special cases of this theorem. We give a proof here and then show a few examples of novel instantiations. 
\begin{proof} 
To prove that $\mathcal{L}_t$ generates $p_t$, we need to show by \cref{thm:general_mass_conservation} that the KFE is fulfilled. Let $\tkernel_{t+h|t}(\cdot|x,z)$ be the transition kernel of $\mathcal{L}_t^z$. Then:
\begin{align*}
  \frac{\dd}{\dd t}\action{p_t}{f}&=
  \lim\limits_{h\to 0}\frac{1}{h}\left(\action{p_{t+h}}{f}-\action{p_t}{f}\right)\\
  &=\lim\limits_{h\to 0}\frac{1}{h}(\mathbb{E}_{Z\sim p_{Z}, X'\sim p_{t+h|Z}(\cdot|Z)}(f(X'))-\mathbb{E}_{Z\sim p_{Z}, X\sim p_{t|Z}(\cdot|z)}(f(X)))\\
  &=\lim\limits_{h\to 0}\frac{1}{h}(\mathbb{E}_{Z\sim p_{Z}, X\sim p_{t|Z}(\cdot|Z), X'\sim \tkernel_{t+h|t}(\cdot|X,Z)}(f(X'))-\mathbb{E}_{Z\sim p_{Z}, X\sim p_{t|Z}(\cdot|z)}(f(X)))\\
  &=\lim\limits_{h\to 0}\frac{1}{h}(\mathbb{E}_{Z\sim p_{Z}, X\sim p_{t|Z}(\cdot|Z), X'\sim \tkernel_{t+h|t}(\cdot|X,Z)}(f(X')-f(X)))\\
&=\lim\limits_{h\to 0}\frac{1}{h}\mathbb{E}_{X\sim p_t}\left(\mathbb{E}_{Z\sim p_{t|Z}(\cdot|X)}\left(\left(\mathbb{E}(_{X'\sim \tkernel_{t+h|t}(\cdot|X,Z)}(f(X'))-f(X)\right)\right)\right)\\
&=\mathbb{E}_{X\sim p_t}\left(\mathbb{E}_{Z\sim p_{Z|t}(\cdot|X)}\left(\lim\limits_{h\to 0}\frac{1}{h}\left(\mathbb{E}_{X'\sim \tkernel_{t+h|t}(\cdot|X,Z)}(f(X'))-f(X)\right)\right)\right)\\
  &=\mathbb{E}_{X\sim p_t}(\underbrace{\mathbb{E}_{Z\sim p_{Z|t}(\cdot|X)}(\mathcal{L}_t^zf(X))}_{=:\mathcal{L}_tf(X)})\\
  &=\action{p_t}{\mathcal{L}_tf}
\end{align*}
The proof for the form of $F_t$ follows then by
\begin{align*}
\mathbb{E}_{Z\sim p_{Z|t}(\cdot|X)}(\mathcal{L}_t^zf(x))
\overset{\eqref{eq:cond_linear_parameterization}}{=}\mathbb{E}_{Z\sim p_{Z|t}(\cdot|X)}(\ip{\mathcal{K}f(x),F_t(x|z)}_{x})=&\ip{\mathcal{K}f(x),\mathbb{E}_{Z\sim p_{Z|t}(\cdot|X)}(F_t(x|z))}_{x}\\
=&\ip{\mathcal{K}f(x),F_t(x)}_{x}
\end{align*}
where we use the linearity of the dot product to swap it with the expected value. This shows that $F_t$ is a linear parameterization (see \cref{eq:linear_parameterization}) of the marginal generator.
\end{proof}

\paragraph{Example - Jumps.} Let $\gS$ be arbitrary and $Q_t(y,x|z)$  be a conditional jump kernel on $\gS$ for $y,x\in \gS, z\in\gZ$ generating the conditional probability path $p_{t|Z}(\dd x|z)$. Using the linear parameterization of the jump kernel (see \cref{e:linear_parameterization_jump}), we get that the \highlight{marginal jump kernel}
\begin{align*}
Q_t(y,x)=\mathbb{E}_{Z\sim p_{Z|t}(\cdot|x)}[Q_t(y,x|z)].
\end{align*}
generates the marginal probability $p_t(\dd x)$.

\paragraph{Example - Marginal diffusion coefficient.} Let $\gS=\R^d$ and $\sigma_t^2(x|z)$ be a diffusion coefficient generating the conditional probability path $p_{t|Z}(\dd x|z)$. Using the linear parameterization of the diffusion coefficient (see \cref{e:linear_parameterization_diffusion}), we get that the \highlight{marginal diffusion coefficient}
\begin{align*}
\sigma_t^2(x)=\mathbb{E}_{Z\sim p_{Z|t}(\cdot|x)}[\sigma_t^2(x|Z)]
\end{align*} 
generates the marginal probability path $p_t(\dd x)$.

\subsection{Generator Matching loss}
Our next goal is to develop a training objective for learning a CTMP model. Let us assume that we have a neural network $F_t^\theta$ that gives us a generator parameterization $\mathcal{L}_t^\theta$ as in \cref{def:generator_neural_network_parameterization}. As derived in \cref{thm:marginal_generator}, our goal is to approximate the true marginal linear parameterization $F_t$ given by \cref{eq:marginal_generator_equation_linear_parameterization}. 

As before, let us assume that for every $x\in\gS$ we have a Bregman divergence $D_{x}:\Omega_{x}\times\Omega_{x}\to\mathbb{R}$ defined via
\begin{align}
\label{eq:general_bregman_divergences}
    D_{x}(a,b)=\Phi_{x}(a) - [\Phi_{x}(b) +\ip{a-b, \nabla \Phi_{x}(b)}],\quad a,b\in \Omega_x
\end{align}
for a strictly convex function $\Phi_{x}:\Omega_{x}\to\R$ (see \cref{fig:bregman}). The \highlight{Generator Matching loss} to train the CTMP model is defined as
\begin{equation}\label{e:gm_loss}
    \gL_{\GM}(\theta) = \E_{t,X_t\sim p_t} D_{X_t}(F_t(X_t), F_t^\theta(X_t)),
\end{equation}
for $t\sim U[0,1]$. Unfortunately, the above training objective is intractable as we do not know the marginal generator $\mathcal{L}_t$ and also not the parameterization $F_t$ thereof (we only know the intractable formula in \cref{eq:marginal_generator_equation_linear_parameterization}). Hence, we introduce the \highlight{Conditional Generator Matching loss} as a tractable alternative that takes the form 
\begin{equation}\label{e:cgm_loss}
   \gL_{\CGM}(\theta) = \E_{t,Z, X_t\sim p_{t|Z}} D_{X_t}(F_t(X_t|Z), F_t^\theta(X_t)).
\end{equation}
This objective is tractable as we can derive $F_t(x|z)$ analytically in many cases (see \cref{subsec:find_cond_generators}). As the next theorem shows, the two losses  \eqref{e:gm_loss} and \eqref{e:cgm_loss} both provide the same learning gradients.

\begin{myframe}
\begin{theorem}\label{thm:cgm}
The gradients of the Generator Matching loss and the Conditional Generator Matching loss coincide:
\begin{equation}
        \nabla_\theta \gL_{\GM}(\theta) = \nabla_\theta \gL_{\CGM}(\theta).
    \end{equation}
    In particular, the minimizer of the Conditional Generator Matching loss is the linear parameterization of the marginal generator (\cref{eq:marginal_generator_equation_linear_parameterization}):
    \begin{equation}
        F_t^\theta(x) = \mathbb{E}_{Z\sim p_{Z|t}(\cdot|x)}[F_t(x|Z)].
    \end{equation}
Furthermore, for these properties to hold, $D_{x}$ must \highlight{necessarily} be a Bregman divergence.
\end{theorem}
\end{myframe}
The above theorem generalizes \cref{thm:cfm}, \cref{thm:rcfm}, and \cref{thm:cdfm}  derived in previous sections to general CTMP models. It allows us to easily train any CTMP model parameterized by a neural network $F_t^\theta$ in a scalable fashion by minimizing the Conditional Generator Matching loss. In addition, it universally characterizes the space of loss functions. The proof of \cref{thm:cgm} is identical as the proof of \cref{thm:cfm} with $u_t$ replaced by $F_t$. For the proof of the necessity of $D$ being a Bregman divergence, we refer to \citep{holderrieth2024gm}.

\paragraph{Example - Training a diffusion coefficient.} We illustrate how \cref{thm:cgm} allows us to train a diffusion coefficient of an SDE. Let $\gS=\R^d$ and $\sigma_t^2(x|z)$ be a diffusion coefficient generating the conditional probability path $p_{t|Z}(\dd x|z)$. We can parameterize the diffusion coefficient in a neural network $(\sigma_t^2)^\theta(x)\in\mathbb{R}^{d\times d}$. The Conditional Generator Matching loss then becomes
\begin{align*}
\gL_{\CGM}(\theta) = \E_{t,Z, X_t\sim p_{t|Z}} \|\sigma_t^2(X_t|Z)- (\sigma_t^2)^\theta(X_t)\|^2
\end{align*} 
where we used the mean squared error as a Bregman divergence (many other options are possible). In \citep{holderrieth2024gm}, examples are shown of models trained in this manner.

\subsection{Finding conditional generators as solutions to the KFE}
\label{subsec:find_cond_generators}
To enable scalable training with the Conditional Generator Matching loss (see \cref{thm:cgm}), we need to be able to find a conditional generator $\mathcal{L}_t^z$ that solves the KFE
\begin{align}
\label{eq:weak_conditional_kfe}
\frac{\dd}{\dd t} \action{p_{t|Z}(\cdot|z)}{f} = \action{p_{t|Z}(\cdot|z)}{\gL_t^z f}\quad \text{ for all }f\in\mathcal{T},z\in\gZ.
\end{align}
If $p_{t|Z}(\dd x|z)$ admits a density $p_{t|Z}(x|z)$ with respect to $\nu$. In this case, we can equivalently solve the adjoint KFE 
\begin{align}
\label{eq:adjoint_conditional_kfe}
\frac{\dd}{\dd t}  p_{t|Z}(x|z) = [(\mathcal{L}_t^z)^*p_{t|Z}(\cdot|z)](x)\quad \text{ for all }x\in\gS, z\in\gZ.
\end{align}
In general, \cref{eq:weak_conditional_kfe} and \cref{eq:adjoint_conditional_kfe} are challenging equations to solve analytically and there is no general formula to solve it for arbitrary generators. Therefore, we give 2 examples on how this can be done as illustration. 

We illustrate it with jump models here, as these work for arbitrary state spaces. As explained in \cref{subsub:example_generators}, they are specified by a jump measure $Q_t$ that can be decomposed into
\begin{align}
    Q_t(\dd y,x) = \lambda_t(x)J_t(\dd y,x)\quad \text{for all }x\in\gS\\
    \label{eq:lambda_constraint}
    \lambda_t(x)\geq 0 \quad \text{ for all }x\in\gS\\
\label{eq:J_t_constaint}
    \int J_t(\dd y,x) = 1 \quad \text{ for all }x\in\gS
\end{align}
where $\lambda_t(x)$ describes the jump intensity and $J_t$ describes a probability kernel specifying the jump distribution. Note that we drop the dependency on $z\in \gZ$ to simplify notation (we keep it for $p_{t|Z}(\dd x|z)$ to avoid confusion with the marginal probability path).

\paragraph{Jump models for convex mixtures.} Let us consider the mixture probability path given by (see \cref{e:mixture_cond}):
\begin{align}
\label{eq:mixture_path_example}
    p_{t|Z}(\dd x|z)=\kappa_t\delta_{z}(\dd x)+(1-\kappa_t)p(\dd x),\quad z\in\gS.
\end{align}
Using the form of the generator for jump processes (see \cref{eq:jump_generator}), the KFE becomes:
\begin{align}
\label{eq:jump_kfe_mixture}
\frac{\dd}{\dd t} \action{p_{t|Z}(\dd x|z)}{f} &= \mathbb{E}_{X\sim p_{t|Z}(\cdot|z)}\lambda_t(X)\mathbb{E}_{Y\sim J_t(\dd y,x)}[f(Y)-f(X)]\quad\text{for all }f\in\gT,x\in\gS
\end{align}
for $\lambda_t,J_t$ satisfying the constraints in \cref{eq:lambda_constraint} and \cref{eq:J_t_constaint}. We make the claim that this is satisfied for a jump model with 
\begin{align*}
Q_t(\dd y,x)=\lambda_t(x)J_t(\dd y,x),\quad\lambda_t(x) = \frac{\dot{\kappa}_t}{1-\kappa_t},\quad J_t(\dd y,x)=\delta_{z}(\dd y)
\end{align*}
\ie, the jump intensity is given by $\lambda_t$ and once we decided to jump, we jump straight to $z\in\gS$. To show this, we show that the above jump process fulfils the KFE. We can derive:
\begin{align*}
\mathbb{E}_{X\sim p_{t|Z}(\cdot|z)}\brac{\lambda_t(X)\mathbb{E}_{Y\sim J_t(\cdot,X)}[f(Y)-f(X)]}
&=\frac{\dot{\kappa}_t}{1-\kappa_t}\mathbb{E}_{X\sim p_{t|Z}(\cdot|z)}[f(z)-f(X)]\\
&=\frac{\dot{\kappa}_t}{1-\kappa_t}[f(z)-\mathbb{E}_{X\sim p_{t|Z}(\cdot|z)}[f(X)]]\\
&=\frac{\dot{\kappa}_t}{1-\kappa_t}[f(z)-[\kappa_tf(z) + (1-\kappa_t)\mathbb{E}_{X\sim p}f(X)]\\
&=\dot{\kappa}_tf(z)-\dot{\kappa}_t\mathbb{E}_{X\sim p}f(X)\\
&=\frac{\dd}{\dd t}\left[\kappa_tf(z)+(1-\kappa_t)\mathbb{E}_{x\sim p}[f(x)]\right]\\
&=\frac{\dd}{\dd t}\mathbb{E}_{X\sim p_{t|Z}(\cdot|z)}[f(X)]\\
&=\frac{\dd}{\dd t}\action{p_{t|Z}(\cdot|z)}{f}.
\end{align*}
Therefore, we see that the process fulfills the jump KFE (\cref{eq:jump_kfe_mixture}). Therefore, we have established a jump model. We have seen a special example of that model for discrete state spaces in \cref{e:discrete_mixture_conditional_velocity}. Here, we have shown that one could also build a similar jump model for Euclidean space $\mathbb{R}^d$, for example.

\paragraph{Jump models for arbitrary paths with densities.}Let us assume that we have a probability $p_{t|Z}(\dd x|z)$ that admits a density $p_{t|Z}(x|z)$ with respect to a reference measure $\nu$ on $\gS$ and that is differentiable in $t$ (note that the mixture path in \cref{eq:mixture_path_example} would not fulfill that for $\gS=\mathbb{R}^d$). Further, we restrict ourselves to jump kernels $J_t(y,x)$ that admit a density. With this, the adjoint KFE becomes the Jump Continuity Equation (\cref{eq:jump_continuity}):
\begin{align}
    \frac{\dd}{\dd  t}p_{t|Z}(x|z)=\int \lambda_t(y)J_t(x,y)p_{t|Z}(y|z)dy-p_{t|Z}(x|z)\lambda_t(x)\\
    \Leftrightarrow \quad p_{t|Z}(x|z)\brac{\frac{\dd}{\dd  t}\log p_{t|Z}(x|z)+\lambda_t(x)}=\int \lambda_t(y)J_t(x,y)p_{t|Z}(y|z)dy
\end{align}
Making $J_t(x,y)=J_t(x)$ (``target-state-independent'') and using $\partial_t=\frac{\dd}{\dd t}$, we get that this equivalent to:
\begin{align}
    p_{t|Z}(x|z)\brac{\partial_t\log p_{t|Z}(x|z)+\lambda_t(x)}&=J_t(x)\int \lambda_t(y)p_{t|Z}(y|z)\nu(dy)\\
   \Leftrightarrow \quad \frac{p_{t|Z}(x|z)\brac{\partial_t \log p_{t|Z}(x|z)+\lambda_t(x)}}{\int \lambda_t(y)p_{t|Z}(y|z)\nu(dy)}&=J_t(x)
\end{align}
In order to define a valid jump process, we require $\lambda_t, J_t$ to satisfy $\lambda_t(x)\geq 0$, $J_t(x)\geq 0$. Therefore, we get:
\begin{align}
    \lambda_t(x) \geq& 0, \ J_t(x)\geq 0\quad 
    \Leftrightarrow \quad \lambda_t(x)\geq \brac{-\partial_t\log p_t(x|z)}_{+}
\end{align}
where $[x]_{+}=\max(x,0)$ describes the ReLU operation. Further, we require $J_t$ to define a valid jump distribution, \ie, integrate to $1$. This can be seen to hold:
\begin{align*}
    1 &= \int J_t(x) dx \\
    \Leftrightarrow \quad \int \lambda_t(x)p_{t|Z}(x|z)\nu(\dd x) &= \int p_{t|Z}(x|z)\brac{\partial_t\log p_{t|Z}(x|z)+\lambda_t(x)}\nu(\dd x)\\
    \Leftrightarrow \quad 0 &= \int \partial_t p_{t|Z}(x|z)\nu(dx)\\
    \Leftrightarrow \quad 0 &= \partial_t \int  p_{t|Z}(x|z)\nu(dx)\\
        \Leftrightarrow \quad 0 &= 0
\end{align*}
\ie, $J_t$ indeed integrates to $1$. Choosing the minimal $\lambda_t(x)$, we get that a jump model defined via
\begin{align*}
    \lambda_t(x) &= \brac{-\partial_t \log p_{t|Z}(x|z)}_{+}, \\
    J_t(x)&= \frac{p_{t|Z}(x|z)[\partial_t\log p_{t|Z}(x|z)]_{+}}{\int p_{t|Z}(y|z)[\partial_t \log p_{t|Z}(y|z)]_{+}\nu(\dd y)}
    =\frac{[\partial_t  p_{t|Z}(x|z)]_{+}}{\int[\partial_t p_{t|Z}(y|z)]_{+}\nu(\dd y)}
\end{align*}
is a solution to the jump continuity equation and therefore generates the conditional probability path $p_{t|Z}(x|z)$. At first, it seems dissatisfying that jump distribution is independent of the location. However, if we extend this model to multiple dimensions, the jump distribution will depend on the location and leads to a powerful generation model \citep{holderrieth2024gm}.

\subsection{Combining Models}
\label{subsec:gm_combining_models}
In this section, we explain how GM allows us to combine generative models for the same state space $\gS$ in different ways.  The underlying principle is simple: the generator is a linear operator and the KFE $\partial_t\action{p_t}{f}= \action{p_t}{\mathcal{L}_t f}$ is a linear equation - so essentially, we can combine solutions for this equation like we do for matrix equations in linear algebra. Specifically, let $\mathcal{L}_t,\mathcal{L}_t'$ be two generators of two Markov processes that solve the KFE for a probability path $p_t$. Then for $\alpha^1_t,\alpha^2_t\in \mathbb{R}$ with $\alpha^1_t+\alpha^2_t=1$ it holds that:
\begin{align}
&\action{p_t}{(\alpha^1_t\mathcal{L}_t+\alpha^2_t\mathcal{L}_t')f}\\
&= \alpha^1_t \action{p_t}{\mathcal{L}_tf}+\alpha^2_t\action{p_t}{\mathcal{L}_t'f} \\
&= \alpha^1_t\partial_t\action{p_t}{f} + \alpha^2_t\partial_t\action{p_t}{f} \\
&= (\alpha^1_t+\alpha^2_t)\partial_t\action{p_t}{f}\\
&=\partial_t\action{p_t}{f},
\end{align}
\ie, $\alpha_t^1\mathcal{L}_t+\alpha^2_t\mathcal{L}_t'$ is again a solution of the KFE. A small but important detail is whether $\alpha_t^{1},\alpha_t^2$ are positive or negative and whether $\mathcal{L}_t,\mathcal{L}_t'$ correspond to Markov processes in forward or backward time. This leads to the following
\begin{myframe}
\begin{proposition}[Combining models] 
\label{prop:markov_superposition}
Let $p_t$ be a marginal probability path, then the following generators solve the KFE for $p_t$ and consequently define a generative model with $p_t$ as marginal:
\begin{enumerate}
\item \highlight{Markov superposition: } $\alpha_t^1\mathcal{L}_t+\alpha^2_t\mathcal{L}_t'$, where $\mathcal{L}_t,\mathcal{L}_t'$ are two generators of Markov processes solving the KFE for $p_t$, and $\alpha_t^{1},\alpha_t^{2}\geq 0$ satisfy $\alpha^1_t+\alpha^2_t=1$.
\item \highlight{Divergence-free components:} $\mathcal{L}_t+\beta_t\mathcal{L}_t^{\divv}$, where $\mathcal{L}_t^{\divv}$ is a generator such that $\action{p_t}{\mathcal{L}_t^{\divv}f}=0$ for all $f\in \gT$, and $\beta_t\geq 0$. We call such $\mathcal{L}_t^{\divv}$ \highlight{divergence-free}. \vspace{-0.3em}
\item \highlight{Predictor-corrector: }$\alpha^1_t\mathcal{L}_t+\alpha^2_t\bar{\mathcal{L}}_t$, where $\mathcal{L}_t$ is a generator solving the KFE for $p_t$ in forward-time and $\bar{\mathcal{L}}_t$ is a generator solving the KFE in backward time, and  $\alpha_t^{1},\alpha_t^{2}\geq 0$ with $\alpha^1_t-\alpha^2_t=1$. 
\end{enumerate}
\end{proposition}
\end{myframe}
We give examples for a Markov superposition and a divergenc-free component here to illustrate \cref{prop:markov_superposition}. An example of the power of a predictor-corrector scheme can be found in \citep{gat2024discrete}.

\paragraph{Markov superposition example - combining jump and flow.} Markov superpositions can be used to combine generative models of different classes. These can be 2 networks trained separately or we can train two GM models in one network simultaneously \citep{holderrieth2024gm}. We illustrate this here by combining a jump model and a flow model on $\gS=\mathbb{R}^d$. Let us assume that we have two models where each of them generates the probability path $p_t$: (1) a flow model $u_t$ and (2) a jump model with jump intensity $\lambda_t$ and jump distribution $J_t$. By \cref{prop:markov_superposition}, for $\alpha_t^1,\alpha_t^2\geq 0$ with $\alpha_t^1+\alpha_t^2=1$, it holds that the following generator defines a valid GM model generating $p_t$:
\begin{align*}
    \mathcal{L}_tf(x) &= \alpha_t^1 \mathcal{L}_t^{\text{jump}}f(x)+\alpha_t^2\mathcal{L}_t^{\text{flow}}f(x)\\
&=(\alpha_t^1\lambda_t(x))\mathbb{E}_{Y\sim J_t(\cdot,x)}[f(Y)-f(x)]+ \nabla f^T(x)(\alpha_t^2u_t(x))
\end{align*}
where we have used \cref{eq:flow_generator} and \cref{eq:jump_generator}. In fact, the above generator describes a \highlight{piecewise-deterministic Markov process}, a combined ODE and jump model \citep{davis1984piecewise}. As the equation above shows, we have to scale the jump intensity by $\alpha_t^1$ and the vector field by $\alpha_t^2$. We can sample from the resulting GM model with the following sampling procedure:
\begin{align*}
    X_0 \sim p_0&=p\\
    X_{t+h} &= \begin{cases}
        \sim J_t(\dd y,X_t) & \text{ with probability }h\alpha_t^1\lambda_t(X_t)\\
        X_{t} + h \alpha_t^2 u_t(X_t) & \text{ with probability }1-h\alpha_t^1\lambda_t(X_t)
    \end{cases} 
\end{align*}
In \citep{holderrieth2024gm}, several examples of Markov superpositions of jump and flow are given and shown to lead to performance improvements.

\paragraph{Divergence-free example - MCMC algorithms.}To find divergence-free components, one can use existing Markov-Chain Monte-Carlo (MCMC) algorithms - all of these algorithms describe a general recipe to find a divergence-free component. We illustrate this with 2 famous examples. Let us assume that we are given a general probability path $p_t$ with density $p_t(x)$. Then for a generator $\mathcal{L}_t^{\divv}$  to be divergence-free is equivalent to that its adjoint maps $p_t$ to zero:
\begin{align}
\label{eq:divfree_condition}
\action{p_t}{\mathcal{L}_t^{\divv}f}=0\text{ for all }f\in\gT \quad \Leftrightarrow \quad [\mathcal{L}_t^{\divv}]^*p_t(x)=0\text{ for all }x\in \gS
\end{align}
First, let us consider $\gS=\mathbb{R}^d$. \highlight{Langevin dynamics} correspond to an SDE with velocity field $\frac{1}{2}\beta_t^2\nabla\log p_t(x)$ and diffusion coefficient $\beta_t$, \ie, the dynamics are given via
\begin{align}
\label{eq:langevin_generator}
dX_t &=\frac{1}{2}\beta_t^2\nabla\log p_t(x)dt + \beta_t dW_t
\end{align}
The adjoint generator of this SDE is given by 
\begin{align*}
[\mathcal{L}_t^{\divv}]^*p_t \overset{(i)}{=}& -\divv(p_t\frac{1}{2}\beta_t^2\nabla\log p_t)(x)+\frac{1}{2}\beta_t^2\Delta p_t(x)\\
    \overset{(ii)}{=}&-\frac{1}{2}\divv(\beta_t^2\nabla p_t)(x)+\frac{1}{2}\beta_t^2\Delta p_t(x)\\
    \overset{(iii)}{=}&-\frac{1}{2}\beta_t^2\Delta p_t(x)+\frac{1}{2}\beta_t^2\Delta p_t(x)=0
\end{align*}
where (i) holds by shape of flow and diffusion adjoint derived in \cref{ss:kfe_examples}, (ii) holds because $\nabla \log p_t = \nabla p_t/p_t$, and (iii) holds by the identity $\divv\nabla =\Delta$. The above shows that the generator of Langevin dynamics fulfils \cref{eq:divfree_condition} and is therefore divergence-free in sense of \cref{prop:markov_superposition}. This fact is widely applied in statistical physics and Markov chain Monte Carlo  \citep{roberts1996exponential}. \Cref{prop:markov_superposition} shows that we can add these dynamics for arbitrary $\beta_t\geq 0$ to any generative model. In \cref{s:relation_to_diffusion_models}, we use this to derive stochastic sampling for diffusion models. 

Second, let $\gS$ be a general state space again. The \highlight{Metropolis-Hastings} algorithm \citep{hastings1970monte} wdescribes the construction of a jump process with jump kernel $Q_t(y,x)$ that satisfies the \highlight{detailed balance condition}:
\begin{align*}
 Q_t(y,x)p_t(x) &= Q_t(x,y)p_t(y)\quad \text{ for all }x,y\in\gS \\
 \Rightarrow \quad [\mathcal{L}_t^{\text{div}}]^*p_t(x) \overset{(i)}{=}& \int Q_t(y,x)p_t(x) -Q_t(x,y)p_t(y) = 0
\end{align*}
where in (i) we used \cref{eq:jump_continuity}. This shows that \cref{eq:divfree_condition} is fulfilled and $Q_t$ is divergence-free. \Cref{prop:markov_superposition} shows that one can arbitrarily add such a Metropolis-scheme to any GM model following the probability path $p_t$.

\pagebreak
\subsection{Multimodal models}

We finally comment on how GM enables the construction of generative models over multiple data modalities jointly. For example, this could be a model that generates images and corresponding text descriptions at the same time. Two modalities are represented as two state spaces $\gS_1, \gS_2$ (\eg, $\gS_1$ are images and $\gS_2$ is text) and a multimodal model would be a generative model over the product space $\gS=\gS_1\times \gS_2$. As $\gS$ is just another state space and GM works for arbitrary state spaces, we could simply go about it like constructing any other GM model. However, there is a specific construction of probability paths that allows us to reuse or ``recycle'' GM models built for individual modalities. For example, we could build a joint text-image model by combining a discrete and continuous FM model. This specific construction relies on \highlight{factorized conditional probability paths}. We have seen a simple case of this already in \cref{s:building_paths_with_factorized_velocities} for discrete FM where factorized probability paths lead to factorized velocities. This holds more generally for arbitrary modalities. While the construction is rather simple and intuitive, it is rather technical to express in full generality. We refer to \citep{holderrieth2024gm} for a rigorous treatment. A specific instantiation of this was also realized in \citep{campbell2024generative} for multimodal protein generation. This highlights that GM enables the construction of multimodal models in a principled and rigorous manner.

\clearpage
\section{Relation to Diffusion and other Denoising Models}
\label{s:relation_to_diffusion_models}
In this section, we finally discuss the relation to denoising diffusion models (DDMs) and related models on non-Euclidean spaces. We mainly focus here on the construction of diffusion models by \citep{song2021sde} via SDEs and explain how it can be placed with the FM/GM family of models. At the end of the section, we also discuss models in other modalities that took inspiration from DDMs (``denoising models'') and how they can be framed as a GM model.

\subsection{Time convention}

The first simple difference between denoising diffusion models and flow matching is a difference how time is parameterized. This is just a convention but is important to avoid confusion. Different to FM,  time is inverted in diffusion models and ranges from $0$ to $\infty$. To differentiate the two time parameterizations, let us use $r$ for the time convention of diffusion models and $t$ for the time convention of FM. Then we have:

\begin{myframe}
\begin{align}
&\text{\highlight{FM time} $t$: }\text{Noise}\equiv ``t=0", \text{Data} \equiv ``t=1"\\
&\text{\highlight{Diffusion time} $r$: }\text{Noise}\equiv``r=+\infty", \text{Data} \equiv ``r= 0"\\
&\text{\highlight{Reparameterization}: }r=k(t),\quad t=k^{-1}(r)
\end{align}
\end{myframe}
where $k:(0,1]\to[0,+\infty)$ is some strictly monotonically decreasing mapping with $k(1)=0$ and $\lim_{t\too 0}k(t)=+\infty$.

\subsection{Forward process vs.~probability paths}

The underlying idea of denoising diffusion models is to construct a \emph{forward process} that corrupts the data distribution. We will explain how this corresponds to a specific construction of a probability path as used in FM. The forward process $X_r$ is defined via the SDE
\begin{align}
\label{eq:forward_process}
    \dd X_r = a_r(X_r)\dd r + g_r\dd W_r,\quad X_0\sim q
\end{align}
where $q$ is the data distribution, $W_r$ is a Brownian motion and $a:\R\times \R^d\to\R^d$ a velocity field, also called \emph{drift} in the context of SDEs, and $g:\R\to\R_{\geq 0}$ a diffusion coefficient (see \cref{subsub:example_generators}). Every such SDE defines a conditional probability path and marginal probability path as 
\begin{alignat}{2}
\label{eq:sde_defines_prob_path}
    &\tilde{p}_{r|0}(x|z) = \mathbb{P}[X_t=x|X_0=z],\qquad &&\tilde{p}_{r}(x)=\mathbb{P}[X_t=x]\\
    &p_{t|1}(x|z)=\tilde{p}_{k(t)|0}(x|z),&& p_t(x)=\tilde{p}_{k(t)}(x)
\end{alignat}
where in the second line we reparameterized time into the FM time parameterization. We see that $p_{t|1}(x|z)$ gives a conditional probability path. Further, the forward process is constructed such that for $R\gg 0$, the distribution of $X_R$ is approximately a Gaussian. Therefore, we get that:
\begin{myframe}
Every forward process in diffusion defines a ``valid'' conditional probability path as used in FM, \ie, the corresponding marginal path interpolates between a data distribution $q$ and (roughly) a Gaussian $p$. Specifically:\\
(1) \highlight{Deterministic initialization: }The conditional probability path $p_{t|1}(x|z)$ corresponds to the distribution of the forward process SDE when initialized with $X_0=z$.\\
(2) \highlight{Data initialization: }The marginal probability path $p_{t}(x)$ corresponds to the distribution of the forward process when initialized with $X_0\sim q$ where $q$ is the data distribution.\\
\end{myframe}
An important requirement of diffusion models is that one  can compute the conditional probability $\tilde{p}_{r|0}(x|z)$ in closed form. This enforces working with SDEs that have analytic solution to their respective KFE (\ie, Fokker-Planck equation). Throughout most of the literature, the forward process is therefore assumed to have \highlight{affine drift coefficients} \ie, $a_r(x)=a_rx$ for some continuous function $a:\R\to\R$ \citep{song2021sde,karras2022elucidating}. This assumption allows us to express the conditional distribution of $X_r$ given $X_0=z\in\R^d$ as a Gaussian distribution \citep{sarkka2019applied,song2021sde, karras2022elucidating}:
\begin{alignat}{3}
\label{e:diffusion_affine_gaussian_prob_path}
&\tilde{p}_{r|0}(x|z) = \mathcal{N}\left(\tilde{\alpha}_rz,\tilde{\sigma}_r^2 I\right),\quad
&&\tilde{\alpha}_r=\exp\parr{\int\limits_{0}^{r}a_w \dd w},\quad     &&\tilde{\sigma}_r^2=\tilde{\alpha}_r^2\int\limits_{0}^{r}\frac{g^2_w}{\tilde{\alpha}^2_w}dw\\
\label{e:affine_path_diffusion}
\Rightarrow \quad & p_{t|1}(x|z) = \mathcal{N}\left(\alpha_tz,\sigma_t^2 I\right)\quad &&\alpha_t = \tilde{\alpha}_{k(t)},\quad && \sigma_t = \tilde{\sigma}_{k(t)}
\end{alignat}
Note that we have discussed such probability paths extensively in \cref{subsubsec:gaussian_paths} as \highlight{affine Gaussian probability paths}, \ie, they are constructed via the affine conditional flow (see \cref{sec:affine_conditional_flows}):
\begin{equation}%
    \psi_t(x|x_1) = \alpha_t z + \sigma_t x,\quad z\sim q,\ x\sim \mathcal{N}(\alpha_0, \sigma_0^2 I).
\end{equation}
Therefore, we can see that:
\begin{myframe}
Forward processes with affine drift coefficients correspond to using \highlight{Gaussian probability paths} (see section \eqref{subsubsec:gaussian_paths}) defined by \cref{e:affine_path_diffusion}.
\end{myframe}
Note that for diffusion models in the above time parameterization, there is no finite times $r<+\infty$ at which the marginal $\tilde{p}_r(x)$ is an exact Gaussian. 

\subsection{Training a diffusion model}
We now discuss how we can recover the training algorithm of diffusion models as a special case of  FM training. In \cref{sec:affine_conditional_flows}, we discussed several options of how to parameterize and train a FM model with Gaussian probability paths (see \cref{thm:cm}). One particular option was \highlight{$x_0$-prediction} where the neural network $x_{0|t}^\theta$ is trained to approximate
\begin{align*}
    x_{0|t}^\theta \approx \mathbb{E}[X_0|X_t=x]
\end{align*}
via the following training algorithm
\begin{align*}
  \gL_{\CM}(\theta) &= \E_{t,Z\sim q, X_0\sim p}\|x_{0|t}^\theta(\underbrace{\alpha_t X_0 +\sigma_tZ}_{=X_t})-X_0\|^2 \\
  &\overset{(i)}{=}\E_{t,Z\sim q, X_t\sim p_{t|1}(\cdot|Z)}\sigma_t^2\|s_t^\theta(X_t)-[-\frac{1}{\sigma_t^2}\parr{X_t -  \alpha_t Z}]\|^2\\
  &\overset{(ii)}{=} \E_{t,Z\sim q, X_t\sim p_{t|1}(\cdot|Z)}\sigma_t^2\|s_t^\theta(X_t)-\nabla\log p_{t|1}(X_t|Z)\|^2,
\end{align*}
where in (i) we reparameterized the neural network via $s^\theta_t=-x_{0|t}^\theta/\sigma_t$ and in (ii) we used \cref{e:score_cond}. The above loss is also called the \highlight{Denoising Score Matching} \citep{vincent2011connection} loss and is the fundamental loss function with which diffusion models are trained. By \cref{thm:cm}, we get that the minimizer $\theta^*$ fulfills that
\begin{align}    \label{eq:marginal_score_to_denoiser}
    s_t^{\theta^*}(x)=-\frac{1}{\sigma_t}\mathbb{E}[X_0|X_t=x]\overset{\eqref{e:score_x_0_identity}}{=}\nabla\log p_t(x),
\end{align}
\ie, at minimal loss, $s^\theta_t$ equals the score function $\nabla\log p_t(x)$ of the marginal probability path. Therefore, the network $s^\theta_t$ is also called the score network. We can therefore summarize:
\begin{myframe}
The training algorithm for diffusion models is \highlight{equivalent} to training a specific FM model with $x_0$-prediction. Specifically, in addition to reparameterizing time, it is the same as training a FM model with:\\
(1) \highlight{Probability path: }Using a Gaussian probability path with independent coupling  constructed via an SDE with affine drift coefficients ($\alpha_t,\sigma_t$ defined via \eqref{e:diffusion_affine_gaussian_prob_path}).\\
(2) \highlight{Score parameterization:} Reparameterizing the marginal velocity field via the score function.
\end{myframe}
In \cref{tab:conversion}, we list how one can easily convert a score network to other velocity parameterizations. Therefore, different parameterizations are theoretically equivalent and one can even swap parameterizations post-training (see \cref{s:velocity_param}). Note however, that the score and $x_0$ prediction parameterizations introduce a singularity at time $t=0$ (near noise).

\subsection{Sampling}
Next, we discuss sampling from a diffusion model and how it relates to sampling from FM or GM model.

\paragraph{Deterministic sampling with ODE.} If we consider the diffusion model a FM model, we would sample by sampling from the marginal vector field. In \eqref{e:u_t_score} we expressed the marginal vector field via the score function (for Gaussian paths): 
\begin{align}
    \label{eq:prob_flow_ODE_vector_field}
    u_t(x)
    =&
    \frac{\dot{\alpha}_t}{\alpha_t}x - \brac{\dot{\sigma}_t\sigma_t - \sigma_t^2\frac{\dot{\alpha}_t}{\alpha_t}}\nabla\log p_t(x).
\end{align}
Using the specific form of \cref{e:diffusion_affine_gaussian_prob_path} for $\alpha_t,\sigma_t$, one derive the equivalent identity:
\begin{align}
    u_t(x)&=\dot{k}(t)\left[a_{t}x - \frac{g_{t}^2}{2}\nabla\log p_t(x)\right].
\end{align}
Alternatively, one insert the above $u_t(x)$ into the Continuity Equation to validate this directly. The corresponding ODE to $u_t$ is also called the \highlight{Probability Flow ODE} \citep{song2021sde}: 
\begin{align}
    \dd X_t&=\dot{k}(t)\left[a_{t}X_t - \frac{g_{t}^2}{2}s_t^\theta(X_t)\right]\dd t,
\end{align}
where we set  $s^\theta_t(x)=\nabla \log p_t(x)$ as the learned score function. 
Note that we use here the notation for ODEs that is common for SDEs for a reason that becomes clear next. We also note that in \cref{eq:prob_flow_ODE_vector_field} we add the term $\dot{k}(t)$ compared to \citep{song2021sde} because of the time reparameterization.

\paragraph{Stochastic sampling with SDE.} In \cref{subsec:gm_combining_models}, we have derived that we can add the Langevin dynamics
\begin{align}
\frac{1}{2}\beta_t^2\nabla\log p_t(x)dt + \beta_t dW_t
\end{align}
to any CTMP generative model and we will obtain a generative model following the same probability path. We can apply this to the Probability Flow ODE to get a whole family of SDEs that generate the probability path $p_t$:
\begin{align}
\label{eq:stochastic_sampling}
dX_t &= \left(\dot{k}(t)\alpha_tX_t+\frac{\beta_t^2-\dot{k}(t)g_t^2}{2}\nabla\log p_{t}(X_t)\right)dt+\beta_tdW_t.\end{align}
The above results in \highlight{stochastic sampling} of a diffusion model. In theory, all models  above lead to the same marginals for every $\beta_t\geq 0$. This is a mathematical fact about the ground truth underlying stochastic process. In practice, we need to simulate the SDE
\begin{align}
\label{eq:stochastic_sampling_with_network}
dX_t &= \left(\dot{k}(t)\alpha_tX_t+\frac{\beta_t^2-\dot{k}(t)g_t^2}{2}s_t^\theta(X_t)\right)dt+\beta_tdW_t\end{align}
with a trained network $s_t^\theta$. We have both estimation errors (\ie, imperfect training of $s_t^\theta$) as well as simulation errors (\ie, imperfect sampling of the underlying SDE). Therefore, there is an optimal unknown noise level $\beta_t$ (see \eg, \citep[equation  (2.45)]{albergo2022building}) that can be determined empirically \citep{karras2022elucidating} and theoretically  \citep{ma2024sit}.

Therefore, we get:
\begin{myframe}
\highlight{1. ODE sampling: } For a  Gaussian source, independent coupling, fixing $\alpha_t,\sigma_t$  according to \eqref{e:diffusion_affine_gaussian_prob_path}, and score parameterization, sampling from a diffusion model with the Probability Flow ODE is the same as sampling from a FM model.\\
\highlight{2. SDE sampling: } Under the same conditions, sampling from a diffusion model with stochastic SDE sampling is equivalent to sampling from GM model defined via \cref{eq:stochastic_sampling}.
\end{myframe}
\subsection{The role of time-reversal and the backward process}
To finish our discussion of diffusion models, we discuss the role of time-reversal and the backward process in the context of diffusion models. Given how central the idea of time-reversal is for diffusion models, it might seem surprising to some readers that it is not needed for FM. We explain this, therefore, here in more detail. Specifically, to generate data, diffusion models frame training as learning a \emph{backward process} $\bar{X}_r$ going from $0$ to $R>0$ such that 
\begin{align}
\label{eq:weak_time_reversal}
    \bar{X}_r \overset{d}{=}& X_{R-r} \text{ for all }r\in [0,R]
\end{align}
where $\overset{d}{=}$ denotes equality in distribution. Once we found such a process, we can initialize $\bar{X}_0\overset{d}{=}X_{R}$ with a Gaussian and simulate it to obtain $\bar{X}_{R}\overset{d}{=}X_0\sim q$, \ie, a sample the data distribution $q$. If $X_{r}\sim \tilde{p}_r$ (\ie, it is the marginal probability path as in \cref{eq:sde_defines_prob_path}), then \cref{eq:weak_time_reversal} is equivalent to:
\begin{align*}
\label{eq:prob_path_time_reversal}
\bar{X}_r \sim& \bar{p}_{r}\defe\tilde{p}_{R-r}\text{ for all }r\in [0,R].
\end{align*}
In other words, we want the stochastic process $\bar{X}_r$ to generate the probability path $\bar{p}_r$. But this is exactly what we are trying to do in Generator Matching for general Markov processes (see \cref{def:generates_ctmp}), and in particular in Flow Matching with flows. Therefore, we get:
\begin{myframe}
The following problems are equivalent:\\
(1) \highlight{Time-reverse marginals: }Finding an SDE (resp.~ODE) for the backward process that has the same marginals as the forward process - as done for diffusion models.\\
(2) \highlight{Generate probability path:} Finding an SDE (resp.~ODE) that generates the probability path defined by the forward process.\\
(3) \highlight{Solve KFE:} Finding an SDE (resp.~ODE) that solves the Fokker-Planck Equation (resp.~Continuity Equation).
\end{myframe}
The original description of diffusion models included the ``full'' time-reversal of an SDE \citep{anderson1982reverse}.  This is a notion that is stronger than the one we use, \ie, it requires that the joint distribution across time points are the same
\begin{align}
\mathbb{P}[\bar{X}_{r_1}\in A_1,\dots,\bar{X}_{r_n}\in A_n]=\mathbb{P}[X_{R-r_1}\in A_1,\dots,X_{R-r_n}\in A_n]\\
    \text{ for all }0\leq r_1,\dots, r_n, \text{ and } A_1,\dots,A_n\subset S\text{ measurable}.
\end{align}
As shown in \citet{anderson1982reverse}, one can obtain a time-reversal satisfying the above condition with a backward process for a specific choice of $\beta_t$ in \cref{eq:stochastic_sampling} . However, for the purposes of generative modeling, we often only use the final point $X_1$ of the Markov process (\eg, as a generated image) and discard earlier time points. Therefore, whether a Markov process is a ``true'' time-reversal or only has the same marginals (as in \cref{eq:weak_time_reversal}) does not matter for many applications. A famous example is the probability flow ODE. The probability flow ODE does \emph{not} constitute a time-reversal of a diffusion process in the sense of \citep{anderson1982reverse} but it follows the same marginals. This illustrates that finding ``true'' time-reversals is a harder mathematical problem to solve but (often) not necessary for the purposes of generative modeling. The probability flow ODE is the current state-of-the-art for sampling with a low number of neural network evaluations \citep{karras2022elucidating}. This shows that the ``true'' time-reversal might even give suboptimal results compared to other solutions of the Fokker-Planck equation \citep{ma2024sit}.

\pagebreak
\subsection{Relation to Other Denoising Model}
Inspired by the success of diffusion models, there were significant advances in translating the methods from diffusion models to other state spaces \citep{campbell2022continuous,campbell2024generative,de2022riemannian, huang2022riemannian, benton2022denoising}. Similarly to diffusion models, they construct a forward Markov process that noises data and then time-reverse it (\ie, ``de-noises'' it). We therefore informally refer to such models as ``denoising models''. Naturally, one can ask how these models relate to GM (see \cref{sec:generatormatching}). In brief, the relation of these ``denoising models'' to GM models is the same as the relation between diffusion models and FM: 
\begin{myframe}
Generally speaking, denoising models are Generator Matching models with\\
(1) \highlight{Time convention: }They use the diffusion time convention where time $0$ corresponds to data.\\
(2) \highlight{Probability path construction: }Probability paths are constructed via a ``forward'' or ``noising'' process.\\
(3) \highlight{Solving the KFE: }A particular solution to the KFE is found via a time-reversal of the forward process.
\end{myframe}
We acknowledge that this an \emph{informal} rule and that there might be exceptions to that rule. Therefore, we refer to an extended and detailed discussion of such related work (including a more complete list of references) to \citep{holderrieth2024gm}.

\clearpage
\newpage
\bibliographystyle{assets/plainnat}
\bibliography{paper}

\clearpage
\newpage
\beginappendix

\section{Additional proofs}
\label{a:proofs}

\subsection{Discrete Mass Conservation}
\label{a:discrete_mass_conservation}

\begin{myframe}
   \begin{lemma}[PMF solutions to Kolmogoroc with rate conditions]\label{lem:pmf_solutions}
    Consider a solution $f_t(x)$ to Kolmogorov Equation \eqref{e:kolmogorov} with initial condition $f_0(x)=p(x)$, where $p$ is a PMF,  $u_t(y,x)$ is $C([0,1])$ in time $t$ and satisfies the rate conditions \eqref{e:rate_conds}. Then $f_t(x)$ is a probability mass function (PMF) for all $t\in [0,1]$.
\end{lemma} 
\end{myframe}
\begin{proof}
Let $f_t(x)$, $t\in[0,1]$, be the solution to the Kolmogorov Equation, the existence and uniqueness of which is guarateed by \cref{thm:linear_system_ode_existence_and_uniqueness}. Now, $f_t(x)$ is a PMF if and only if it satisfies 
    \begin{equation}
        f_t(x)\geq 0, \text{ and } \sum_x f_t(x)=1.
    \end{equation}
The latter condition is shown to hold by summing both sides of the Kolmogorov Equation to get that the solution satisfies
\begin{equation*}
    \frac{\dd}{\dd t}\sum_x f_t(x) = \sum_x \sum_z u_t(x,z)p_t(z) = 0
\end{equation*}
where the second equality is due to $\sum_y u_t(y,x)=0$ in the rate conditions. Since $\sum_x f_0(x) = \sum_x p(x) = 1$ we hae that $\sum_x f_t(x) \equiv 1$ for all $t\in [0,1]$.

To prove that $f_t(x)\geq 0$ for all $x\in \gS$ we will use a result on convex invariant sets of dynamical systems. In particular, Theorem 7.3.4 in \cite{pruss2010gemeinliche} asserts that as long as $f_0=p$ satisfies this condition (which it does) and whenever $w(z)$ is on the boundary of this constraint, \ie, $w$ is a PMF and $w(z)=0$ for some $z\in \gS$, then a non-positive inner-product with the outer-normal to the constraint, \ie,  $\sum_{x,y} u_t(y,x) w(x) \delta(y,z) \geq 0 $ implies that the solution $f_t(x)\geq 0$ for all $t\in[0,1]$ and $x\in \gS$. Let us check this condition:
\begin{equation*}
    \sum_{x,y} u_t(y,x) w(x) \delta(y,z) = \sum_x u_t(z,x)w(x) = \sum_{x\ne z} u_t(z,x)w(x) \geq 0,
\end{equation*}
where in the second equality we use the fact that $w(z)=0$ and in the last inequality we used the rate condition \eqref{e:rate_conds} that $u_t(z,x)\geq 0$ for $z\ne x$ and $w(y)\geq 0$ for all $y$.

\end{proof}

\begin{myframe}
\begin{reptheorem}{thm:discrete_mass_conservation}[Discrete Mass Conservation]
Let $u_t(y,x)$ be in $C([0,1))$ and $p_t(x)$ a PMF in $C^1([0,1))$ in time $t$. Then, the following are equivalent:
    \begin{enumerate}
        \item $p_t,u_t$ satisfy the Kolmogorov Equation \eqref{e:kolmogorov} for $t\in[0,1)$, and $u_t$ satisfies the rate conditions \eqref{e:rate_conds}.
        \item $u_t$ generates $p_t$ in the sense of \ref{def:generates_ctmc} for $t\in[0,1)$.
    \end{enumerate}
\end{reptheorem}    
\end{myframe}
\begin{proof}
Let us start by assuming 2. In this case, the probability transition kernel $p_{t+h|t}(y|x)$ satisfies \eqref{e:ctmc_model},
\begin{equation}\label{ae:transition_kernel}
    p_{t+h|t}(y|x) = \delta(y,x) + hu_t(y,x) + o(h).
\end{equation}
By expressing the marginal $p_t(y)$ using the Law of Total Probability, we obtain
\begin{equation}\label{e:total_prob}
    p_{t+h}(y) = \sum_x p_{t+h|t}(y|x)\, p_t(x).
\end{equation}
By plugging \eqref{ae:transition_kernel} into \eqref{e:total_prob} and rearranging, we get 
\begin{align*}
    \frac{p_{t+h}(y) - p_t(y)}{h} &= \sum_x u_t(y,x)p_t(x) + o(1),
\end{align*}
where now $o(1)=o(h)/h\too 0$ as $h\too 0$, as per the definition of $o(h)$.
Taking the limit $h\too 0$, we get that the pair $(p_t,u_t)$ satisfies the Kolmogorov Equation \eqref{e:kolmogorov}.
Next, let us prove that $u_t$ satisfies the rate conditions \eqref{e:rate_conds}.
If $u_t(y,x)<0$ for some $y\ne x$, it follows from \eqref{ae:transition_kernel} that $p_{t+h|t}(y|x)<0$ for small $h>0$, and this contradicts $p_{t+h|t}$ being a probability kernel.
If $\sum_y u_t(y,x)=c\ne 0$, it follows from \ref{ae:transition_kernel} that $1=\sum_{x}p_{t+h|t}(y|x)=1 + hc + o(h)$, leading to a contradiction for small $h>0$.

Conversely, assume now condition 1.
That is, the pair $(u_t,p_t)$ satisfies the Kolmogorov Equation \eqref{e:kolmogorov} with initial condition $p_0=p$.
By \cref{thm:linear_system_ode_existence_and_uniqueness}, let $p_{s|t}(y|x)$ be the unique solution to the Kolomogorov Equation
\begin{equation}
    \frac{\dd}{\dd s}p_{s|t}(y|x) = \sum_z u_t(y,z) p_{s|t}(z|x),
\end{equation}
with initial condition $p_{t|t}(y,x)=\delta(y,x)$, where $0\leq t \leq s < 1$ and $t$ and $y$ are held constant.
By \cref{lem:pmf_solutions}, $p_{s|t}(\cdot|x)$ is a PMF.
The term $\sum_x p_{s|t}(y|x)p(x)$ also satisfies the Kolmogorov Equation, since
\begin{equation}\label{ea:semigroup}
    \frac{\dd}{\dd t}\sum_x p_{s|t}(y|x)p(x) = \sum_x \brac{\sum_z u_t(y,z)p_{s|t}(z|x)}p(x) = \sum_z u_t(y,z) \brac{\sum_x p_{s|t}(z|x) p(x)},
\end{equation}
now with initial conditions $\sum_x p_{t|t}(y|x)p(x) = p(y)$.
Due to the uniqueness of solutions to the Kolmogorov Equation (\cref{thm:linear_system_ode_existence_and_uniqueness}), it follows that $\sum_x p_{s|t}(y|x) p(x) = p_s(y)$, as required.
Lastly, the semi-group property of the transition kernel, $\sum_z p_{s|r}(y|z) p_{r|t}(z|x) = p_{s|t}(y|x)$ for $0\leq t\leq r\leq s < 1$, can be shown by repeating the argument in \ref{ea:semigroup} with $p_{r|t}$ as initial condition at time $r$. In conclusion, we found a transition kernel $p_{t+h|t}$ that generates $p_t$, as required.  
\end{proof}

\subsection{Manifold Marginalization Trick}
\label{a:manifold_marginalization_trick}
\begin{myframe}
\begin{reptheorem}{thm:rfm_main}[Manifold Marginalization Trick]
Under Assumption \ref{as:riemannian_p_t}, if $u_t(x|x_1)$ is conditionally integrable and generates the conditional probability path $p_t(\cdot|x_1)$ then the marginal velocity field $u_t(\cdot)$ generates the marginal probability path $p_t(\cdot)$.
\end{reptheorem} 
\end{myframe}
\begin{proof} 
To prove that $u_t(\cdot)$ generates $p_t(\cdot)$, we will again show they satisfy the conditions in the Mass Conservation Theorem. First, we check that $u_t(x)$ and $p_t(x)$ satisfy the Continuity Equation \eqref{e:riemannian_continuity}: 
\begin{align}\label{e:riemannian_marginal_trick_proof}
\frac{\dd}{\dd t}p_t(x) &\,\overset{({i})}{=} \int_\gM \frac{\dd}{\dd t}p_{t|1}(x|x_1)q(x_1)\dd \vol_{x_1} \\ &\overset{({ii})}{=} -\int_\gM \divv_g\brac{u_t(x|x_1) p_{t|1}(x|x_1)} q(x_1) \dd \vol_{x_1}\\
&\,\overset{({i})}{=} -\divv_g \int_\gM u_t(x|x_1) p_{t|1}(x|x_1)q(x_1) \dd \vol_{x_1}\\ 
&\overset{({iii})}{=} -\divv_g\brac{u_t(x)p_t(x)},
\end{align}
where in ({i}) we switched differentiation ($\frac{\dd}{\dd t}$ and $\divv_g$) and integration justified by Leibniz rule, and the fact that $p_{t|1}(x|x_1)$ and $u_t(x|x_1)$ are $C^1$ in $t,x$ and $q$ has bounded support or $\gM$ is compact. In ({ii}), we used the fact that $u_t(\cdot|x_1)$ generates $p_{t|1}(\cdot|x_1)$ and \cref{thm:riemannian_continuity}. In ({iii}), we multiplied and divided by $p_t(x)$ (which is strictly positive by assumption) and used the formula for $u_t$ in \eqref{e:riemanniam_u_t}. Lastly, $u_t$ is integrable and locally Lipschitz by employing the same arguments as in the proof of \cref{thm:fm_main}.
\end{proof}

\subsection{Regularity assumptions for KFE}
\label{appendix:regularity_assumptions_list}
We note that assumption 5 is true under relatively weak  assumptions and there is a diversity of mathematical literature on showing uniqueness of the solution of the KFE in $p_t$ for different settings. However, to the best of our knowledge, there is no known result that states regularity assumptions for general state spaces and Markov processes, which is why simply state it here as an assumption. For the machine learning practitioner, this assumption holds for any state space of interest. To illustrate this, we point the rich sources in the mathematical literature that show uniqueness that list the regularity assumptions  for specific spaces and classes of Markov processes:
\begin{enumerate}
\item Flows in $\mathbb{R}^d$ and manifolds: \citep[Mass conservation formula, page 15]{villani2009optimal}, \citep{diperna1989ordinary}, \citep{ambrosio2004transport}
\item Diffusion in $\mathbb{R}^d$ and manifolds: \citep[Diffusion theorem, page 16]{villani2009optimal}
\item General Ito-SDEs in $\mathbb{R}^d$: \citep[Theorem 1.3 and 1.4]{figalli2008existence}, \citep[Corollary 1.3]{kurtz2011equivalence}, \citep{bogachev2022fokker}
\item Discrete state spaces: Here, the KFE is a linear ODE, which has a unique solution under the assumption that the coefficients are continuous (see \cref{thm:discrete_mass_conservation}). 
\end{enumerate}

\end{document}

%% file: math_commands.tex
\newcommand{\fmlibrary}{\Verb{flow_matching} library}

\newcommand{\dd}{\mathrm{d}}
\newcommand{\tkernel}{p}
\newcommand{\action}[2]{\left \langle #1, #2\right \rangle }

\newcommand{\norm}[1]
{\left\Vert#1\right\Vert}

\newcommand{\abs}[1]{\left\vert#1\right\vert}

\newcommand{\set}[1]{\left\{#1\right\}}
\newcommand{\parr}[1]{\left (#1\right )}
\newcommand{\brac}[1]{\left [#1\right ]}
\newcommand{\ip}[1]{\left \langle #1 \right \rangle }
\newcommand{\Real}{\mathbb R}
\newcommand{\Nat}{\mathbb N}

\newcommand{\eps}{\varepsilon}
\newcommand{\too}{\rightarrow}

\newcommand{\dist}{d} %
\newcommand{\divv}{\mathrm{div}} %
\newcommand{\vol}{\mathrm{vol}} %

\newcommand{\trace}{\textrm{tr}} %
\newcommand{\defe}{\coloneqq}%

\definecolor{mygray}{gray}{0.95}
\newcommand{\CM}{\scriptscriptstyle \text{CM}}
\newcommand{\M}{\scriptscriptstyle \text{M}}
\newcommand{\CFM}{\scriptscriptstyle \text{CFM}}
\newcommand{\FM}{\scriptscriptstyle \text{FM}}
\newcommand{\RFM}{\scriptscriptstyle \text{RFM}}
\newcommand{\RCFM}{\scriptscriptstyle \text{RCFM}}
\newcommand{\DFM}{\scriptscriptstyle \text{DFM}}
\newcommand{\CDFM}{\scriptscriptstyle \text{CDFM}}
\newcommand{\GM}{\scriptscriptstyle \text{GM}}

\newcommand{\CGM}{\scriptscriptstyle \text{CGM}}

\newcommand{\mask}{\texttt{m}} %

\newcommand*{\eg}{{\it e.g.}\@\xspace}
\newcommand*{\ie}{{\it i.e.}\@\xspace}

\theoremstyle{plain}
\newtheorem{theorem}{Theorem}%
\newtheorem{proposition}{Proposition}
\newtheorem{lemma}{Lemma}
\newtheorem{corollary}{Corollary}
\theoremstyle{definition}

\newtheorem{assumption}{Assumption} 
\theoremstyle{remark}

\makeatletter
\newtheorem*{rep@theorem}{\rep@title}
\newcommand{\newreptheorem}[2]{%
\newenvironment{rep#1}[1]{%
 \def\rep@title{\textbf{#2} \ref{##1}}%
 \begin{rep@theorem}}%
 {\end{rep@theorem}}}
\makeatother

\newreptheorem{theorem}{Theorem}
\newreptheorem{proposition}{Proposition}
\newreptheorem{lemma}{Lemma}
\newreptheorem{corollary}{Corollary}

\def\eqref#1{equation~\ref{#1}}

\def\1{\bm{1}}

\def\eps{{\epsilon}}

\DeclareMathAlphabet{\mathsfit}{\encodingdefault}{\sfdefault}{m}{sl}
\SetMathAlphabet{\mathsfit}{bold}{\encodingdefault}{\sfdefault}{bx}{n}

\def\gA{{\mathcal{A}}}

\def\gD{{\mathcal{D}}}

\def\gL{{\mathcal{L}}}
\def\gM{{\mathcal{M}}}
\def\gN{{\mathcal{N}}}

\def\gP{{\mathcal{P}}}

\def\gS{{\mathcal{S}}}
\def\gT{{\mathcal{T}}}
\def\gU{{\mathcal{U}}}

\def\gY{{\mathcal{Y}}}
\def\gZ{{\mathcal{Z}}}

\def\sP{{\mathbb{P}}}

\newcommand{\E}{\mathbb{E}}

\newcommand{\R}{\mathbb{R}}

\newcommand{\Cov}{\mathrm{Cov}}

\DeclareMathOperator*{\argmin}{arg\,min}

\newcolumntype{C}[1]{>{\Centering}m{#1}}

\newcolumntype{Z}[1]{>{\Left}m{#1}}